\documentclass[sn-mathphys-num]{sn-jnl}%

\usepackage{graphicx}%
\usepackage{multirow}%
\usepackage{makecell}
\usepackage{tabularx}
\usepackage{amsmath,amssymb,amsfonts}%
\usepackage{amsthm}%
\usepackage{mathrsfs}%
\usepackage[title]{appendix}%
\usepackage[dvipsnames]{xcolor}
\usepackage{textcomp}%
\usepackage{manyfoot}%
\usepackage{booktabs}%
\usepackage{algpseudocode}%
\usepackage{listings}%
\usepackage{caption}
\usepackage{booktabs}
\usepackage{pifont}
\newcommand{\xmark}{\textcolor{red}{\ding{55}}}%
\newcommand{\cmark}{\textcolor{ForestGreen}{\ding{51}}}%
\usepackage{footnote}
\usepackage[bottom]{footmisc}

\usepackage{xspace}

\newcommand{\blue}[1]{#1}

\usepackage{tcolorbox}
\definecolor{mygreen}{HTML}{3cb44b}

\usepackage[linesnumbered,ruled,vlined]{algorithm2e}
\DontPrintSemicolon

\SetKwComment{Comment}{\color{green!50!black}\# }{}

\newcommand{\var}{\texttt}

\SetKwProg{Function}{def}{:}{}

\SetKwProg{For}{for}{:}{}
\SetKwProg{If}{if}{:}{}
\newcommand{\VarSty}[1]{\textnormal{\ttfamily\color{blue!90!black}#1}\unskip}

\theoremstyle{thmstyleone}%

\theoremstyle{thmstyletwo}%

\theoremstyle{thmstylethree}%

\begin{document}

\title[Article Title]{
\begin{center}
Insect-Foundation: A Foundation Model and \\ Large Multimodal Dataset for \\ Vision-Language Insect Understanding
\end{center}
}

\author[1]{\fnm{Thanh-Dat} \sur{Truong}}\email{\texttt{tt032@uark.edu}}
\equalcont{These authors contributed equally to this work.}

\author[1]{\fnm{Hoang-Quan} \sur{Nguyen}}\email{\texttt{hn016@uark.edu}}
\equalcont{These authors contributed equally to this work.}

\author[1]{\fnm{Xuan-Bac} \sur{Nguyen}}\email{\texttt{xnguyen@uark.edu}}

\author[2]{\fnm{Ashley} \sur{Dowling}}\email{\texttt{adowling@uark.edu}}

\author[3]{\fnm{Xin} \sur{Li}}\email{\texttt{xli48@albany.edu}}

\author[1]{\fnm{Khoa} \sur{Luu}}\email{\texttt{khoaluu@uark.edu}}

\affil[1]{\orgdiv{Department of Electrical Engineering and Computer Science}, \orgname{University of Arkansas}, \orgaddress{\state{AR}}}

\affil[2]{\orgdiv{Department of Entomology and Plant Pathology}, \orgname{University~of~Arkansas}, \orgaddress{\state{AR}}}

\affil[3]{\orgdiv{Department of Computer Science}, \orgname{SUNY Albany}, \orgaddress{\state{NY}} \vspace{3mm}}

\affil{\small{\texttt{\href{https://uark-cviu.github.io/projects/insect-foundation}{https://uark-cviu.github.io/projects/insect-foundation}}}}

\newcommand{\Quan}[1]{\textcolor{teal}{Quan: #1}}

\abstract{

Multimodal conversational generative AI has shown impressive capabilities in various vision and language understanding through learning massive text-image data.
However, current conversational models still lack knowledge about visual insects since they are often trained on the general knowledge of vision-language data.
Meanwhile, understanding insects is a fundamental problem in precision agriculture, helping to promote sustainable development in agriculture.
Therefore, this paper proposes a novel multimodal conversational model, \textbf{Insect-LLaVA}, to promote visual understanding in insect-domain knowledge.
In particular, we first introduce a new large-scale \textbf{Multimodal Insect {Dataset}} with \textbf{Visual Insect Instruction} Data that enables the capability of learning the multimodal foundation models. 
Our proposed dataset enables conversational models to comprehend the visual and semantic features of the insects.
Second, we propose a new \textbf{Insect-LLaVA} model, a new general Large Language and Vision Assistant in Visual Insect Understanding.
Then, to enhance the capability of learning insect features, 
we develop an \textbf{Insect Foundation Model} by introducing a new micro-feature self-supervised learning with a Patch-wise Relevant Attention mechanism to capture the subtle differences among insect images.
We also present Description Consistency loss to improve micro-feature learning via text descriptions. 
The experimental results evaluated on our new \textbf{Visual Insect Question Answering} benchmarks illustrate the effective performance of our proposed approach in visual insect understanding and achieve State-of-the-Art performance on standard benchmarks of insect-related tasks.

}


\maketitle

\section{Introduction}

\begin{figure}[!b]
\centering
\includegraphics[width=\linewidth]{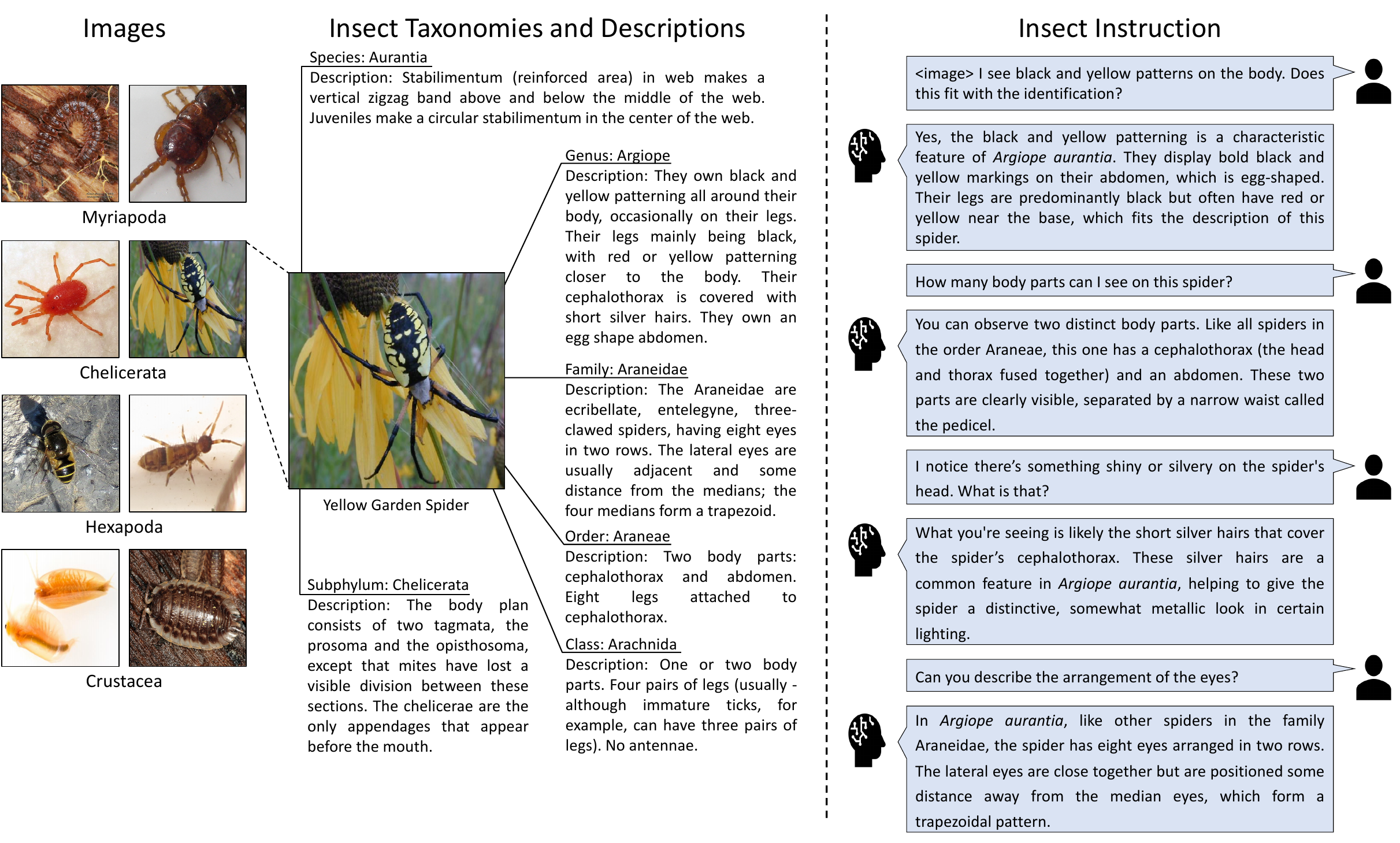}
\caption{\textbf{Our Proposed Multimodal Dataset and Visual Insect Instruction Data.} The left figure illustrates the samples of the four Subphylums, including Chelicerata, Crustacea, Hexapoda, and Myriapoda. The middle figure shows an example of hierarchical descriptions of the Aurantia Species. The right figure illustrates the corresponding insect instruction data.
} 
\label{fig:subphylums_with_description}
\end{figure}

Foundation models have become a widespread research interest that aims to build general-purpose assistants \cite{alayrac2022flamingo,yang2023dawn,li2024multimodal}, also known as conversational generative models, e.g., LLaVA \cite{liu2024visual,liu2024improved}, LLaVA-Med \cite{li2024llava}, etc.
These models \cite{liu2024visual,liu2024improved, radford2021learning, li2023blip} show strong capabilities in visual understanding, such as classification, detection, segmentation, and captioning.
While large multimodal models present their robustness, large-scale multimodal datasets play an important role in allowing the models to learn the alignments between different modalities.
Besides the general-purpose foundation models, the large multimodal models for specific fields, e.g., biomedicine \cite{li2024llava}, social analysis \cite{zhang2024somelvlm}, or agriculture \cite{gharaee2023step}, remain a significant research topic because of their fine-grain understanding requirements.
While the foundation models have been introduced for several tasks, e.g.,  biomedicine \cite{li2024llava} and social analysis \cite{zhang2024somelvlm}, there are limited studies in developing foundation models for precision agriculture. 
In the development of agriculture, the visual identification and understanding of insects play a significant role in establishing healthy crop growth and high-quality production.

The success of these models relies on two key factors, including the emergence of large-scale text-image data and the development of large pre-trained visual foundation models and large language models.
While the conversational generative models have been well developed for general purposes \cite{askell2021general,gan2022vision,li2022elevater}, to the best of our knowledge, there are limited studies on developing the generative models of visual insect understanding. 
Therefore, in this paper, levering the success of our developed large-scale insect dataset and the insect foundation model, we propose a new \textbf{Large Language and Vision Assistant in Visual Insect Understanding}, i.e., \textbf{Insect-LLaVA},  model. 
To develop our conversational generative model, i.e., Insect-LLaVA, we propose a new large-scale \textbf{Multimodal Insect Data} with \textbf{Visual Insect Instruction Data} (Figure \ref{fig:subphylums_with_description}). 
Table \ref{tab:data_comparison} compares our proposed Multimodal Insect data with prior insect datasets.
The success of our proposed Insect-LLaVA promises to empower entomology research and help entomologists in various tasks related to insect understanding.

\begin{table*}[!t]
\centering
\caption{Comparison with existing datasets related to insects. Our proposed dataset has hierarchical labels with six main hierarchical levels, i.e., Subphylum, Class, Order, Family, Genus, and Species, and large numbers of species and samples. Moreover, the proposed dataset contains hierarchical descriptions for each insect and auxiliary taxonomic level, and visual instruction data.  
}\label{tab:data_comparison}
\setlength{\tabcolsep}{2pt}
\resizebox{1.0\linewidth}{!}{
\begin{tabular}{lccccccccc}
\Xhline{2\arrayrulewidth}
\textbf{Dataset} & \textbf{Venue} & \textbf{Year} & \textbf{Species} & \makecell{\textbf{Hierarchical}\\ \textbf{Labels}} & \makecell{\textbf{Hierarchical}\\ \textbf{Levels}} & \makecell{\textbf{Insect} \\ \textbf{Description}} & \makecell{\textbf{Auxiliary} \\ \textbf{Taxonomic}} & \makecell{\textbf{Instruction} \\ \textbf{Data}} & \makecell{\textbf{Number of} \\ \textbf{Samples}} \\
\hline
Samanta et al. \cite{samanta2012tea}       & IJCES & 2012 & 8      & \xmark  & 1 & \xmark  & \xmark  & \xmark & 609 \\
Wang et al. \cite{wang2012new}             & KBS & 2012 & 221      & \cmark  & 3 & \xmark  & \xmark & \xmark &225 \\
Venugoban et al. \cite{venugoban2014image} & IJMLC & 2014 & 20     & \xmark  & 1 & \xmark  & \xmark  & \xmark &200 \\
Xie et al. \cite{xie2015automatic}         & CEA & 2015 & 24     & \xmark  & 1 & \xmark  & \xmark & \xmark &1,440 \\
Liu et al. \cite{liu2016localization}      & Sci. Rep. & 2016 & 12     & \xmark  & 1 & \xmark  & \xmark  & \xmark &5,136 \\
Xie et al. \cite{xie2018multi}             & CEA & 2018 & 40     & \xmark  & 1 & \xmark  & \xmark & \xmark &4,500 \\
Deng et al. \cite{deng2018research}        & BE & 2018 & 10     & \xmark  & 1 & \xmark  & \xmark & \xmark &563 \\
Alfarisy et al. \cite{alfarisy2018deep}    & ICMAI & 2018 & 13     & \xmark  & 1 & \xmark  & \xmark  & \xmark &4,511 \\
PestNet \cite{liu2019pestnet}              & IEEE Access & 2019 & 16  & \xmark & 1 & \xmark & \xmark & \xmark &88,670 \\
IP102 \cite{wu2019ip102}                   & CVPR & 2019 & 102    & \cmark & 3 & \xmark  & \xmark & \xmark &75,222 \\
AgriPest \cite{wang2021agripest}           & Sensors & 2021 & 14 & \cmark & 2 & \xmark & \xmark & \xmark &49,707 \\
INSECT \cite{badirli2021fine}              & NeurIPS & 2021 & 1,213 & \xmark & 1 & \xmark & \xmark & \xmark &21,212 \\
iNat-2021 \cite{van2021benchmarking}       & CVPR & 2021 & 2,752 & \cmark & 5 & \xmark & \xmark & \xmark &723,816 \\
\hline
\textbf{Insect-1M (Ours) \cite{nguyen2024insect}} & CVPR & \textbf{2024} & \textbf{34,212} & \textbf{\cmark} & \textbf{6} & \textbf{\cmark} & \textbf{\cmark} & \xmark & \textbf{1,017,036} \\

\textbf{Multimodal Insect (Ours)}                 & $-$ & \textbf{2024} & \textbf{34,212} & \textbf{\cmark} & \textbf{6} & \textbf{\cmark} & \textbf{\cmark} & \cmark & \textbf{1,017,036} \\
\Xhline{2\arrayrulewidth}
\end{tabular}
}
\end{table*}

One key factor in the success of the large multimodal foundation model is the vision encoder model, which extracts the meaningful feature representation of visual information.
Recent vision foundation models \cite{he2020momentum,chen2020improved,chen2020simple,xie2022simmim,he2022masked,caron2021emerging,oquab2023dinov2,radford2021learning,jia2021scaling,yu2022coca} show reliable performance on downstream tasks as they are designed to learn properties of images or videos from large-scale datasets that can perform well on unseen data.
These models show their capability with self-supervised and multi-modal learning on large-scale datasets \cite{deng2009imagenet,jia2021scaling,zhai2022scaling,schuhmann2022laion}.
However, the current insect datasets \cite{deng2018research,wu2019ip102,alves2020cotton,bollis2020weakly,samanta2012tea,wang2012new,venugoban2014image,xie2015automatic,liu2016localization,xie2018multi,alfarisy2018deep} are insufficient to establish the foundation model of insects due to their scale and diversity.
Indeed, the most recent work presents an insect recognition dataset containing over $75,000$ images of 102 species \cite{wu2019ip102}.
Although the dataset includes many species, compared to the species of insects in the natural environment with over 5.5 million species \cite{stork2018many, ratnasingham2007bold}, the current work needs to have the diversity of insects.
Furthermore, to our knowledge, the current insect dataset \cite{wu2019ip102} does not provide the corresponding insect descriptions, limiting the ability to learn the foundation models.

\begin{figure}[!t]
\centering
\includegraphics[width=1.0\linewidth]{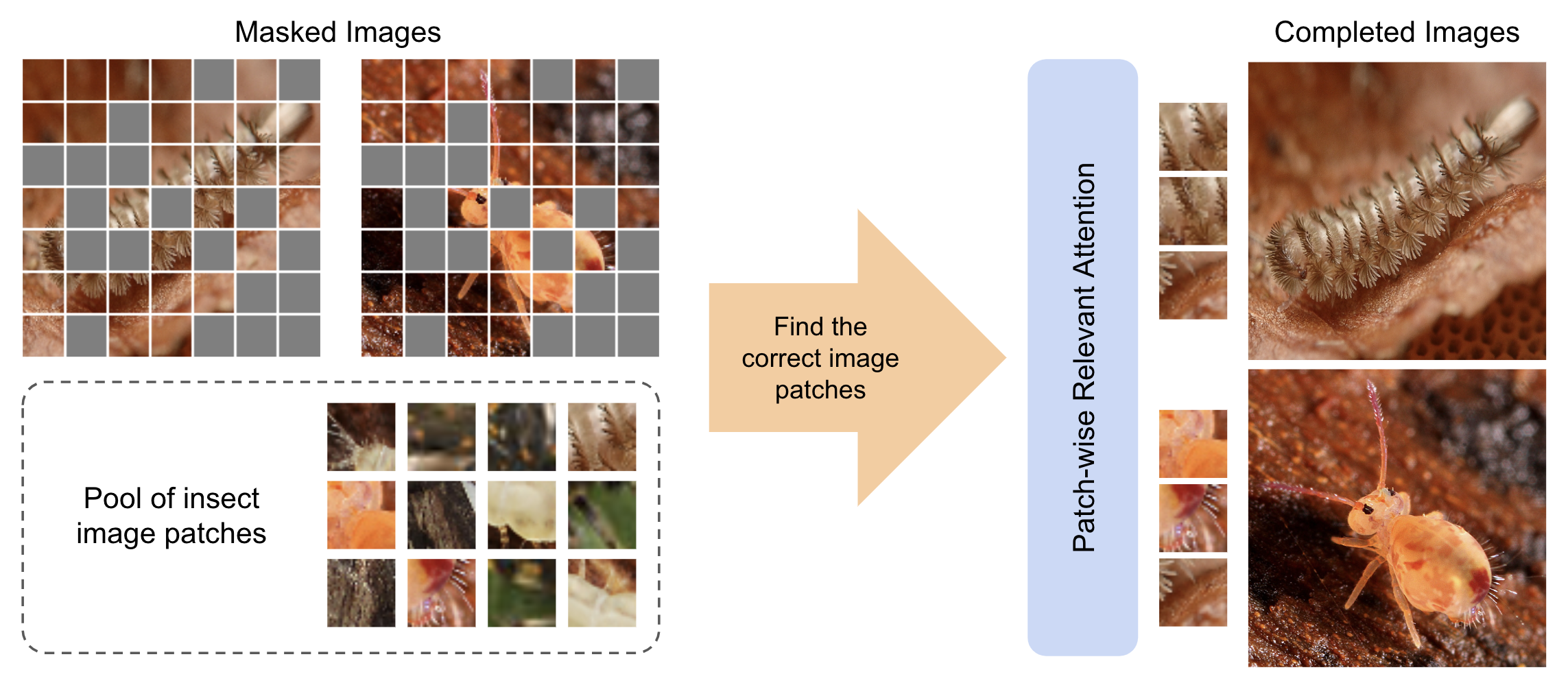}
\caption{
\textbf{Our Patch-wise Relevant Attention.}
Given masked insect images and separated image patches, our model can learn to distinguish patches with minor differences via relevant scores computed between masked images and image patches.
}
\vspace{-4mm}
\label{fig:idea}
\end{figure}

While the dataset plays a significant role in building an insect foundation model, the learning technique shows its significant factor in the performance of the foundation model.
Most of the approaches in foundation model learning are learning the alignment between multiple modalities, i.e., vision and language, to model the distribution of concepts \cite{radford2021learning,jia2021scaling,yu2022coca}.
Other learning approaches provide various pre-text tasks, i.e., self-supervised contrastive learning and distillation learning, for vision model learning and show their scaling ability and generalization with downstream tasks \cite{he2020momentum,chen2020improved,chen2021empirical,caron2021emerging,oquab2023dinov2,he2022masked}.
However, most prior approaches to the foundation models learn to extract the general information of natural images but lack specific knowledge.
\blue{
For example, self-supervised learning methods (e.g., MAE \cite{he2022masked}) focused on reconstructing masked image regions but struggled to capture the micro-features essential for distinguishing insect species. Jigsaw-based methods \cite{chen2023jigsaw, truong2022direcformer} learn structural relationships but fail to emphasize micro-features of insects. Similarly, Micron-BERT \cite{nguyen2023micron} highlights minor differences by swapping image regions, yet key species-defining traits often remain unchanged, limiting its ability to capture fine-grained distinctions. 
Joint vision-language pre-training models (e.g., CLIP \cite{radford2021learning}) aligned image-text embeddings but rely on high-level semantic annotations, which are insufficient for precise insect identification. These models were trained on datasets dominated by common objects rather than specialized entomological data, making them ineffective in capturing micro-level features of insects. Furthermore, the lack of detailed textual insect descriptions in large-scale datasets further limits their ability to guide fine-grained feature learning. Large vision-language assistant models (e.g., LLaVA \cite{liu2024visual, liu2024improved}) integrate advanced multimodal reasoning but primarily depend on language priors that often lack specialized insect knowledge or misinterpret species-specific details. Their attention mechanisms tend to prioritize global context over micro-features of features, leading to potential misidentification.
Therefore, capturing fine-grained discrimination (micro-features) between insect samples plays an important role in the insect foundation model because of the high diversity of species.
To develop a robust insect foundation model, the learning approach is required to represent the micro-features of insects.
}
Motivated by this intuition, we introduce a novel pre-text task to improve the learning capacity of the model between the micro features of the insect, as shown in Figure \ref{fig:idea}.

\noindent
{
\textbf{Contributions.} To advance precision agriculture research, we introduce \textbf{\textit{Insect-LLaVA}}, a novel multimodal foundation model designed for comprehensive insect understanding. At its core, we develop a new \textbf{\textit{Insect Foundation Model}} that captures fine-grained visual insect features and generalizes to various downstream applications, such as insect detection, classification, vision-language understanding, and visual question answering. To support this, we propose a large-scale dataset, \textbf{\textit{Multimodal Insect}}, which includes \textbf{\textit{Visual Insect Instruction Data}} to facilitate learning in generative and multimodal models.
}
{
Building on our preliminary work \cite{nguyen2024insect}, which introduced the Insect-1M dataset and achieved state-of-the-art results in insect understanding, we significantly expand our dataset and model to enhance multimodal capabilities. Our \textbf{\textit{Multimodal Insect Dataset}} comprises one million densely labeled insect images spanning the entire taxonomic hierarchy—from Class and Order to Genus and Species—each paired with a detailed textual description.
Notably, Insect-1M is $13\times$ larger than the previous IP102 dataset \cite{wu2019ip102}. 
}
{
To improve the vision encoder of \textbf{\textit{Insect-LLaVA}}, we introduce a self-supervised contrastive learning paradigm with a novel Patch-wise Relevant Attention mechanism to model intricate insect features. Additionally, we propose a Description Consistency loss to enhance learning from textual descriptions. To evaluate \textbf{\textit{Insect-LLaVA}}, we introduce a new \textbf{Visual Insect Question Answering (Insect-VQA)} benchmark, assessing the model’s understanding of insect-related visual tasks. Through extensive experiments on Insect Classification, Insect Detection, and Insect-VQA benchmarks, we demonstrate the superior performance of our approach compared to prior methods.
}

\section{Related Work} 

This section will first review the current available public insect datasets. Then, we present the current approaches to developing foundation models and large-vision assistant models.

\subsection{Insect Datasets}

Previous studies have presented insect datasets on a small scale for recognition tasks.
\cite{venugoban2014image} introduced a dataset containing $20$ species, with $10$ samples per species.
Later, \cite{xie2015automatic} presented a dataset with $1,440$ samples from $24$ species. 
More recently, larger datasets suitable for deep learning have been developed. \cite{xie2018multi} created a dataset of $4,500$ images spanning $40$ species for insect classification, and \cite{liu2016localization} introduced a dataset with over 5,000 samples focused on insect recognition and localization. 
PestNet \cite{liu2019pestnet}, and AgriPest \cite{wang2021agripest} were specifically developed for small pest detection tasks.
Additionally, \cite{wu2019ip102} introduced IP102, a large-scale dataset containing over $75,000$ insect samples from $102$ species for classification and detection tasks. 
Meanwhile, \cite{van2021benchmarking} presented a dataset with more than $723,000$ samples representing $2,752$ species from the Arthropoda phylum.
Although prior efforts promoted the development of vision and machine intelligence in precision agriculture, no dataset has a large volume of samples and diverse species for insect-related foundation model training. 
Therefore, this work introduces a novel dataset that not only contains a large number of samples, i.e. 1M images, but also has hierarchical labels from the high to the low taxonomy level, including class, order, family, genus, and species.
Table \ref{tab:data_comparison}  compares our proposed dataset with the prior ones.
In comparison with prior datasets, the number of images in our proposed Insect-1M dataset is $13\times$ higher than the prior IP102 dataset, and the number of species is $335\times$ higher than IP102 \cite{wu2019ip102}.
To preserve the rights of datasets and authors of images, instead of publishing images, we only provide labels and links to download images.

\subsection{Self-supervised Pre-training}

Self-supervised pre-training has gained significant research interest as a strategy for solving various visual recognition tasks, i.e., classification, localization, segmentation, video recognition, tracking, and many other tasks \cite{he2022masked, truong2022direcformer, nguyen2021clusformer, truong2021bimal, truong2024fairness, truong2023fredom, truong2023liaad,nguyen2023brainformer,nguyen2023fairness,sefl_supervised_medical,nguyen2023micron,nguyen2022two}.
SimCLR \cite{chen2020simple} learned visual representations through a contrastive learning framework applying various image augmentations.
MoCo \cite{he2020momentum} presented momentum updating to improve the encoder learning for image representations using contrastive learning. 
This framework was later refined to enhance the performance of SimCLR without needing a large batch size \cite{chen2020improved}.
The later method MoCo-V3 \cite{chen2021empirical} further improved by removing the memory queue, ensuring training stability for greater batch sizes.
DINO \cite{caron2021emerging} introduced a self-supervised learning method based on knowledge distillation, which was later extended to DINO-V2 \cite{oquab2023dinov2}, providing improved stability when scaling the size of models and data.
BEiT \cite{bao2021beit} presented a masked image modeling task where discrete visual tokens from the original image were used as prediction targets.
MAE \cite{he2022masked} and SimMIM \cite{xie2022simmim} used a decoder to directly reconstruct pixel values in masked image regions. 
Jigsaw-ViT \cite{chen2023jigsaw} proposed a pre-training task for transformer models by finding spatial positions from shuffled image patches.
This approach was also applied to temporal data to enhance video modeling robustness \cite{truong2022direcformer}.
Micron-BERT \cite{nguyen2023micron} explored subtle changes in facial videos by learning to detect minor differences in images where regions had been swapped between frames.

\subsection{Joint Vision-Language Pre-training}
Recent advantages of joint vision-language pre-training models have been introduced.
CLIP \cite{radford2021learning} and ALIGN \cite{jia2021scaling} demonstrated that dual-encoder models trained on image-text pairs with contrastive objectives can learn strong representations for cross-modal alignment and zero-shot image recognition tasks.
LiT \cite{zhai2022lit} and BASIC \cite{pham2023combined} introduced zero-shot transfer learning techniques by training the text model to learn from the pre-trained image model through contrastive losses on large-scale datasets.
SimVLM \cite{wang2021simvlm}, OFA \cite{wang2022ofa}, and BLIP \cite{li2022blip} employed an encoder-decoder architecture trained with language generative losses, achieving high performance on vision-language benchmarks.
CoCa \cite{yu2022coca} combined contrastive learning with generative image captioning to improve global representation learning and fine-grained image-text alignment.
Subsequent research \cite{zhai2023sigmoid} utilized sigmoid loss to calculate image-text similarity, enabling batch size scaling.
LexLIP \cite{luo2023lexlip} mapped images into a lexicon space to facilitate sparse image-text matching, while EQSIM \cite{wang2023equivariant} computed similarity via equivariant changes between images and text.

\subsection{Large Language-Vision Assistant Models}

The development of large-scale data processing and large language models (LLMs) has provided a new vehicle to solve complex problems, including multimodal data. Some of them can be accounted including large vision-language models \cite{liu2024improved, liu2024visual}, large video-language models \cite{weng2024longvlm, zhao2023learning}, or large audio-language models \cite{hussain2023m, ghosal2023text}.
By incorporating the power of LLMs, the large-scale multimodal models (LMMs) have revolutionized the research of large-scale multimodal.
In the design of LMMs, different input modalities, i.e., images/videos and languages, are connected to LLMs via the project modules \cite{liu2024visual}. Then, the alignment across modalities is performed via cross-attention \cite{liu2024improved}, Q-Formers \cite{li2023blip}, or MLP \cite{liu2024visual}.
The training procedure of LMMs typically has two major steps: pre-training and instruction-tuning.
While the first stage learns the alignment of features across modalities, the second stage enables reasoning about concepts in multimodal inputs and tasks.
Recently, Chen and Zhang \cite{chenfedmbridge} improved LMM learning via multimodal federated learning.
\blue{Other studies enhanced the performance of LMM by introducing high-resolution, Fine-grained, and Pixel-level Vision approaches \cite{ren2024pixellm, yuan2024osprey, zhang2024omgllava, fei2024vitron, wu24next}. However, these prior studies have not been specifically designed to address the challenges of modeling micro-features of insects.}
Several benchmarks were introduced to evaluate the LMM performance, e.g., MMMU \cite{yue2024mmmu}, and
MM-SpuBench \cite{ye2024mm}.

\begin{figure*}[!t]
\begin{center}
\includegraphics[width=\linewidth]{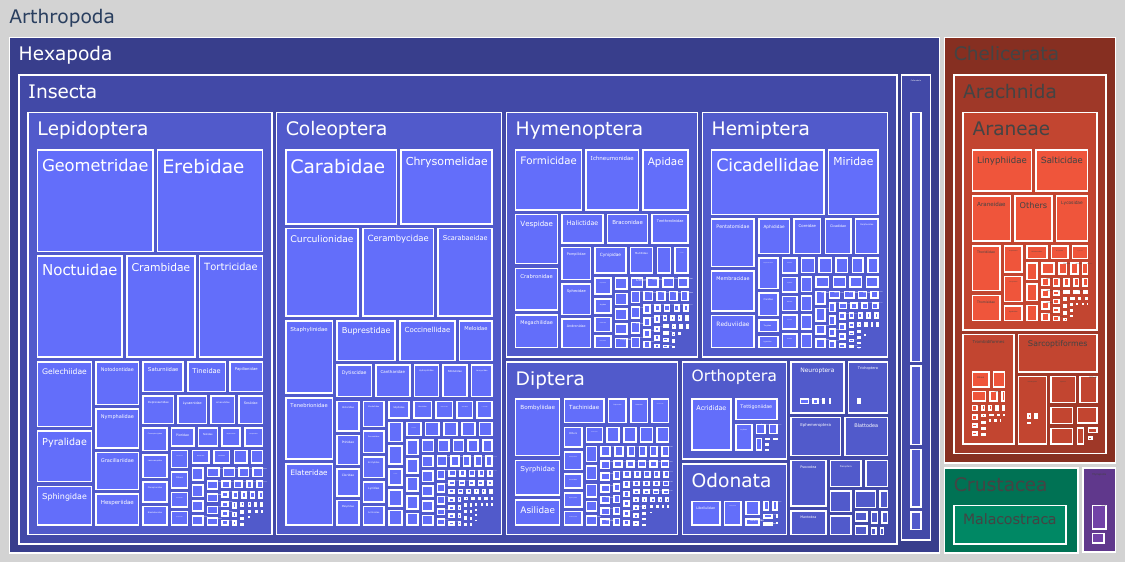}
\end{center}
\caption{\textbf{Treemap of the Multimodal Dataset.} Nested boxes represent classes, orders, and families. The size of the boxes represents the relative number of samples.}
\label{fig:treemap}
\vspace{-5mm}
\end{figure*}

\section{The Proposed Multimodal Insect with Visual Insect Instruction Data}

To establish the multimodal foundation and conversational generative models in visual insect understanding, the large-scale dataset of insects with diverse species is essential. Therefore, we collect a new insect dataset with dense labels of a hierarchical taxonomy. In particular, our Insect-1M dataset contains 1 million insect images with dense hierarchical labels with six main taxonomies, i.e., Subphylum, Class\footnote{In this paper, we use the term ``Class'' as a biological taxonomic level.}, Order, Family, Genus, and Species.
Figure \ref{fig:treemap} illustrates the treemap of our Multimodal Insect data.
The samples are in the Phylum Arthropoda and can be divided into 4 Subphylums, which are Chelicerata, Crustacea, Hexapoda, and Myriapoda as shown in Figure \ref{fig:subphylums_with_description}. 
Compared to prior datasets, our Insect-1M has more hierarchical levels with large numbers of species and samples as in Table \ref{tab:data_comparison}.

\begin{figure}[!b]
    \centering
    \includegraphics[width=1.0\linewidth]{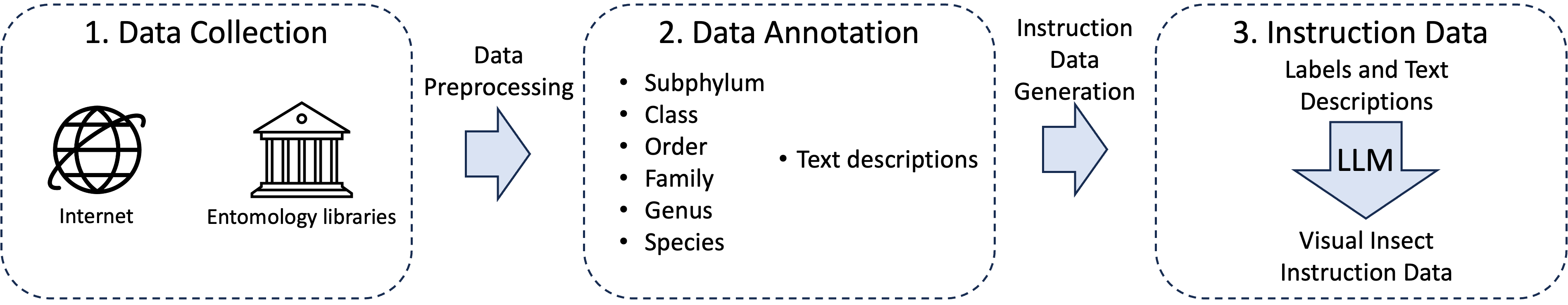}
    \caption{Our Data Collection Pipeline.}
    \vspace{-4mm}
    \label{fig:data-pipeline}
\end{figure}

\subsection{The Proposed Multimodal Insect Dataset}

\noindent
\textbf{Data Collection Protocol.} We utilize insect information containing insect data with images and taxonomies collected by naturalists and entomologists. Figure \ref{fig:data-pipeline} illustrates our data collection pipeline.
Each insect sample has a corresponding image and its taxonomic label.
From the taxonomic label, we crawl the identification description of the corresponding taxonomy.
Notice that the taxonomic labels are hierarchical. The description is written from high-level descriptions, e.g., Subphylum and Class, to low-level descriptions, e.g., Species. Figure \ref{fig:subphylums_with_description} shows an example of an insect description.

\vspace{2mm}

\noindent
\textbf{Data Preprocessing.} The raw data is stored in over 1 million HTML files with predefined HTML structures.
Then, we parse the data structures to collect the insect images and their labels.
More than 2 million raw images and their corresponding labels have been collected. 
However, the raw data collected consists of a lot of noise, e.g., incorrect identification of insects, corrupted images, and non-insect images. Therefore, to filter these outliers, our entomology experts must verify the images and their labels, i.e., insect identification. 
Finally, our collected Multimodal Insect (Insect 1M) dataset consists of $1,017,036$ clean images with dense labels of $34,212$ different insect species.

\subsection{Visual Insect Instruction Data}

While several prior insect datasets have been proposed, there is still a lack of visual instruction data. To address this problem, we further propose the \textbf{Visual Insect Instruction Data} based on our Multimodal Insect (Insect 1M) dataset. Following the standard instruction data of prior work \cite{liu2024visual}, our insect instruction data consists of two sets, i.e., the Pre-training Dataset and the Fine-tuning Dataset.

\vspace{2mm}

\noindent
\textbf{Visual Insect Feature Alignment Pre-training Dataset.} For an insect image $I$, we construct the pre-training dataset by sampling a question $\mathbf{X}_q$, which asks to describe the insect image. Then, the answer $\mathbf{X}_a$ will be corresponding to the insect description of insect image $I$. Formally, the single-turn instruction data of the pre-training insect dataset can be formed as $\texttt{Human:} \mathbf{X}_q, I \texttt{<STOP>\textbackslash n Assistant: } \mathbf{X}_a\texttt{<STOP>\textbackslash n}$. Then, question $\mathbf{X}_q$ are randomly sampled from the following list.

\begin{tcolorbox}
    \centering
    \small
\begin{itemize} %
\setlength{\itemsep}{2pt}
    \item ``Describe the following insect in detail"
    \item ``Provide a detailed description of the given insect image"
    \item ``Give an elaborate explanation of the insect you see"
    \item ``Share a comprehensive rundown of the presented insect"
    \item ``Offer a thorough analysis of the insect"
    \item ``Explain the various aspects of the insect before you"
    \item ``Clarify the contents of the displayed insect ge with great detail"
    \item ``Characterize the insect using a well-detailed description"
    \item ``Break down the elements of the insect in a detailed manner"
    \item ``Walk through the important details of the insect"
    \item ``Portray the insect with a rich, descriptive narrative"
    \item ``Narrate the contents of the insect with precision"
    \item ``Analyze the insect in a comprehensive and detailed manner"
    \item ``Illustrate the insect through a descriptive explanation"
    \item ``Examine the insect closely and share its details"
    \item ``Write an exhaustive depiction of the given insect"
\end{itemize}
\end{tcolorbox}

\noindent
\textbf{Insect Instruction Fine-tuning Dataset.}
To train the conversational generation model of insects aligned with diverse instructions, we present an approach to generate the insect multi-round conversations of insect images by prompting only language to the large language model. Due to the high cost of the commercial language model of GPT-4 \cite{achiam2023gpt}, we choose to use the open-source LLaMA-3 \cite{dubey2024llama} with a competitive performance for the optimal cost while we are still able to generate good quality instruction data.
In particular, given an insect description, we design instructions in a prompt that asks the language model to produce a multi-turn conversation (question and answer) in a tone as if the language model could see the insect image. 
{Figure \ref{fig:subphylums_with_description} illustrates the example of our insect instruction data.}
The prompt to create the instruction data can be formed as follows.

\begin{tcolorbox} 
\small
\VarSty{messages} = [
            \{\var{"role":"system", "content":} f``````You are an AI visual assistant specialized in entomology topics, and you are seeing a single insect image. What you see are provided with sentences, describing the same image you are looking at. Answer all questions as you are seeing the insect image. Design a conversation between you and a person asking about this insect photo. The answers should be in a tone that a visual AI assistant is seeing the insect image and answering the question. Ask diverse questions and give
            corresponding answers. Include questions about the image's visual content, including the insect phylum, subphylum, class, order, superfamily, family, subfamily, etc. Only include questions that have definite answers:
            
            (1) one can see the content in the insect image that the question asks about and can answer confidently;
            
            (2) one can determine confidently from the insect image that it is not in the image. Do not ask any questions that cannot be answered confidently.

    Also include complex questions that are relevant to the content in the insect image, for example, asking about background knowledge of the insects in the image, asking to discuss insects in
    the image, etc. Again, do not ask about uncertain details. Provide detailed answers when answering
    complex questions. For example, give detailed examples or reasoning steps to make the content more
    convincing and well-organized. You can include multiple paragraphs if necessary."""\}\\
        ]
        
    \For{ \VarSty{sample} in \VarSty{fewshot\_samples}}{
         \var{\VarSty{messages}.append(\{"role":"user", "content":\VarSty{sample[`context']}\})} \; \\
         \var{\VarSty{messages}.append(\{"role":"assistant", "content":\VarSty{sample[`response']}\} ) } \;
         }  
    \var{\VarSty{messages}.append(\{"role":"user", "content":`\textbackslash  n'.join(\VarSty{query})\})}

\end{tcolorbox}

\begin{figure}[!b]
\begin{center}
\includegraphics[width=\linewidth]{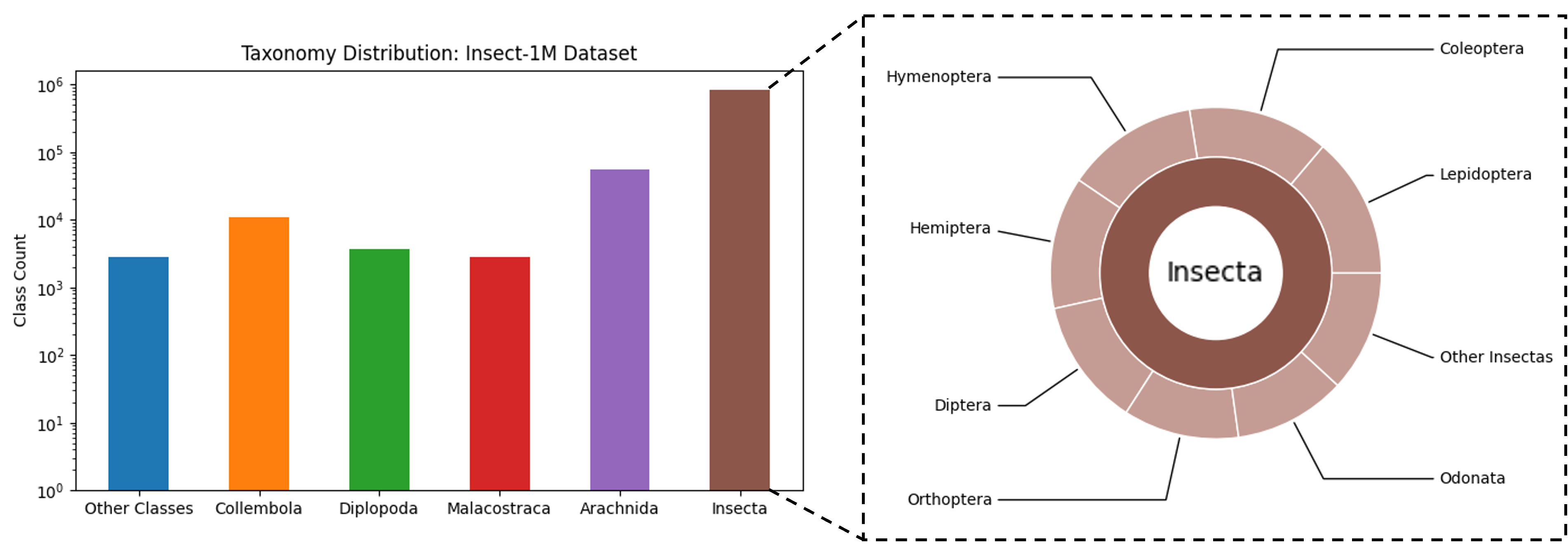}
\end{center}
\caption{
    The Distribution of Classes in Multimodal Insect Dataset (Left) and the Distribution of Insecta Orders (Right). 
}
\vspace{-6mm}
\label{fig:data_chart}
\end{figure}

\noindent
\blue{
\textbf{Instruction Data Filtering.}
Following the LLaVA protocol in data filtering, the text-based filtering approach has been applied to remove invalid or noisy answers, i.e., (1) The responses are incomplete, (2) LLaMA does not provide answers, (3) The responses include words that suggest the answer is not based on the insect image. Finally, we manually validate the responses of 15\% samples of the dataset to ensure the quality of the visual insect instruction data. 
}

\subsection{\blue{Data Statistic}}

\begin{figure}[!t]
    \centering
    \includegraphics[width=1.0\linewidth]{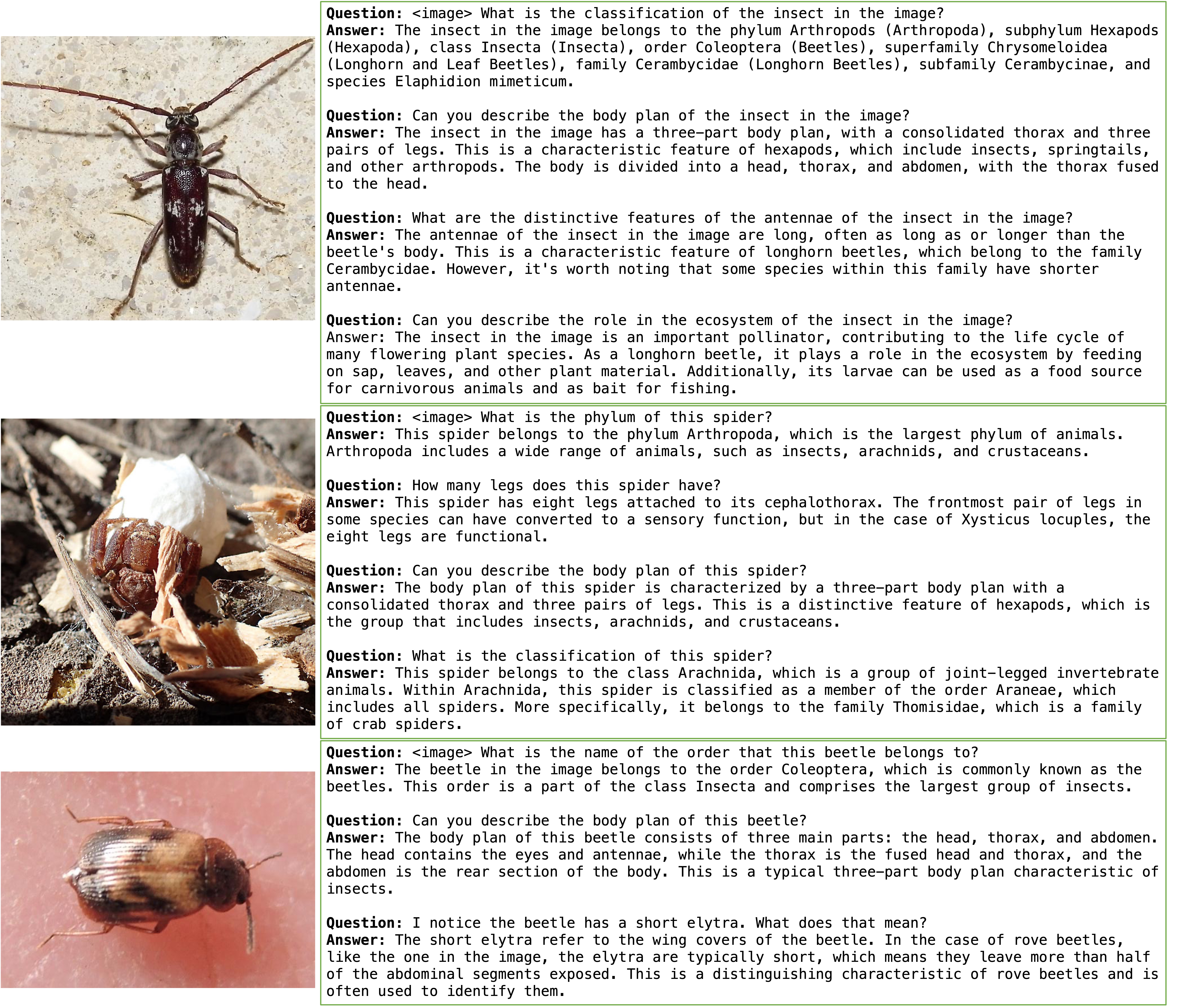}
    \caption{Examples of Visual Instruction Data in our Multimodal Insect Dataset.}
    \label{fig:insect-data-instruction}
\end{figure}

\blue{Our proposed Multimodal Insect Dataset comprises a total of $1,017,036$ images, each paired with instruction-response pairs, covering a diverse range of insect-related queries. These insect images span 34,212 different species, ensuring extensive taxonomic coverage. Each sample is systematically categorized into six main hierarchical taxonomy levels, specifically Subphylum, Class, Order, Family, Genus, and Species, allowing for precise classification and retrieval of taxonomic information.  Figure \ref{fig:data_chart}  shows the sample distributions of the Subphylums and their Classes.}
\blue{The instruction-response pairs are structured into multiple categories, addressing distinct aspects of insect-related inquiries. 
Figure \ref{fig:insect-data-instruction} illustrates an example of our visual insect instruction data.
The instruction categories in our proposed data include:}

\begin{itemize}
    \item\blue{ \textbf{Insect Identification} consists of queries related to recognizing and distinguishing different insect species.}

    \item\blue{ \textbf{Appearance} includes questions about visual characteristics such as size, color, and wing patterns.}
    
    \item\blue{ \textbf{Insect Characteristics} cover biological traits like behavior, lifecycle, feeding habits, and ecological roles.}
    
    \item\blue{ \textbf{Geographic Localization} provides insights into the natural habitat and geographical distribution of insect species.}
    
    \item\blue{  \textbf{Reasoning} involves answering complex queries that require logical deductions based on insect features.}
    
    \item\blue{ \textbf{Descriptive Queries.} cover general information, background, and unique facts about insects.}
\end{itemize}

\begin{figure}[!t]
    \centering
    \includegraphics[width=\linewidth]{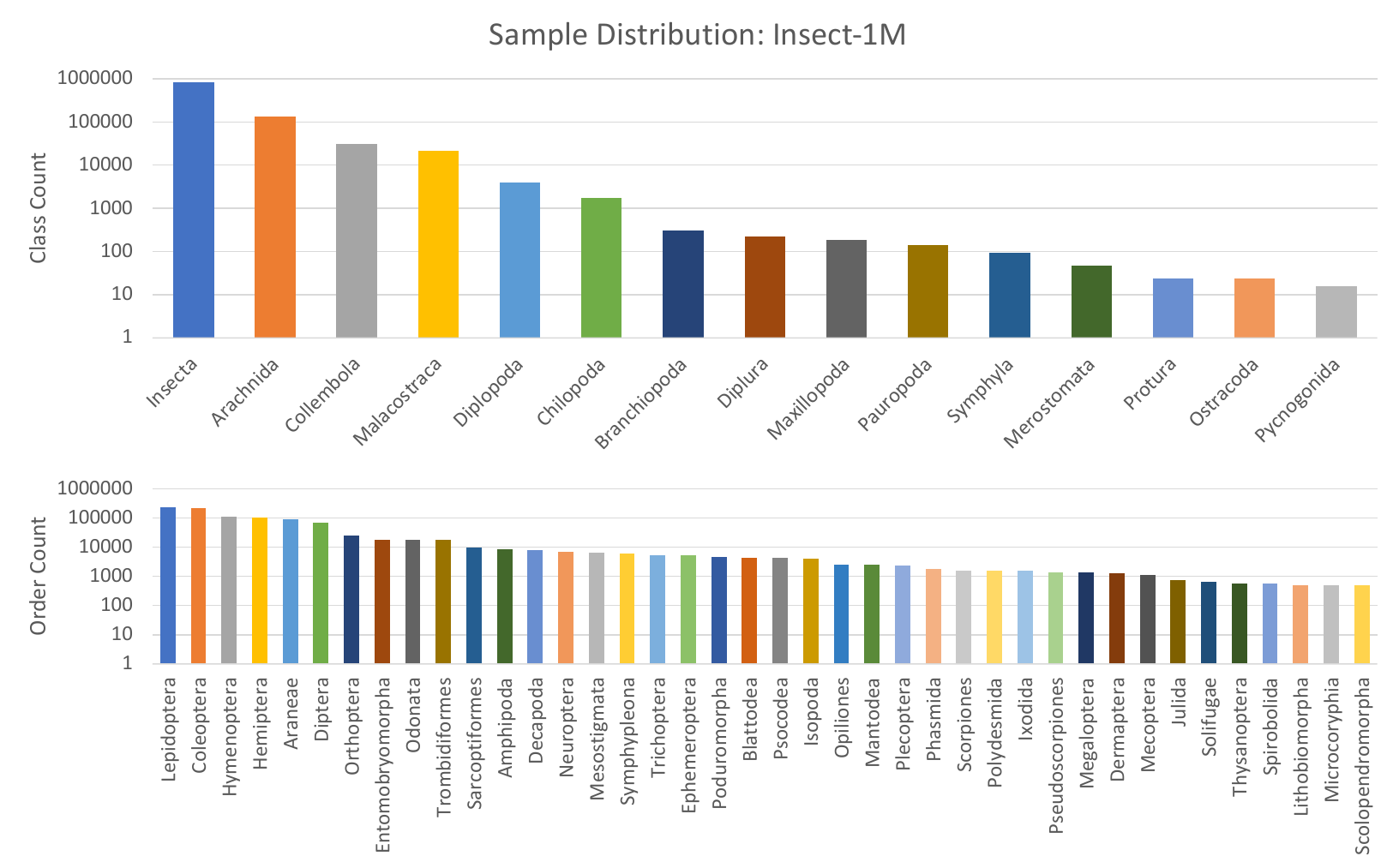}
    \caption{
    Sample distribution of the Classes (top) and the 40 most popular Orders (bottom) in the Insect-1M dataset.
    }
    \label{fig:class_distribution}
\end{figure}

\blue{The response length varies significantly depending on the complexity of the query, ranging from 4 to 341 words. On average, a single response contains 48 words, ensuring concise yet informative answers. Additionally, the number of instruction turns per interaction varies, ranging between 2 and 25 turns, with an average of 5 turns per instruction set. This dynamic range allows for both simple, direct responses and multi-turn, in-depth discussions, making the dataset versatile for training models that handle both basic and complex queries effectively.}
\blue{By incorporating a vast number of species, structured taxonomy, diverse instruction categories, and varied response lengths, our dataset is designed to enhance the understanding and recognition of insects. Figure \ref{fig:class_distribution} illustrates the sample distributions of Classes and Orders in the proposed dataset.}

\section{The Proposed Large Language and Vision Assistant in Visual Insect Understanding}

In this section, we first introduce our proposed Large Language and Vision Assistant in Visual Insect Understanding (Insect-LLaVA), followed by presenting the current limitations of foundation model training. Then, we will present our approach to developing a robust Insect Foundation Model for visual insect understanding tasks.

\subsection{The Proposed Insect-LLaVA}

\begin{figure}
    \centering
    \includegraphics[width=1.0\textwidth]{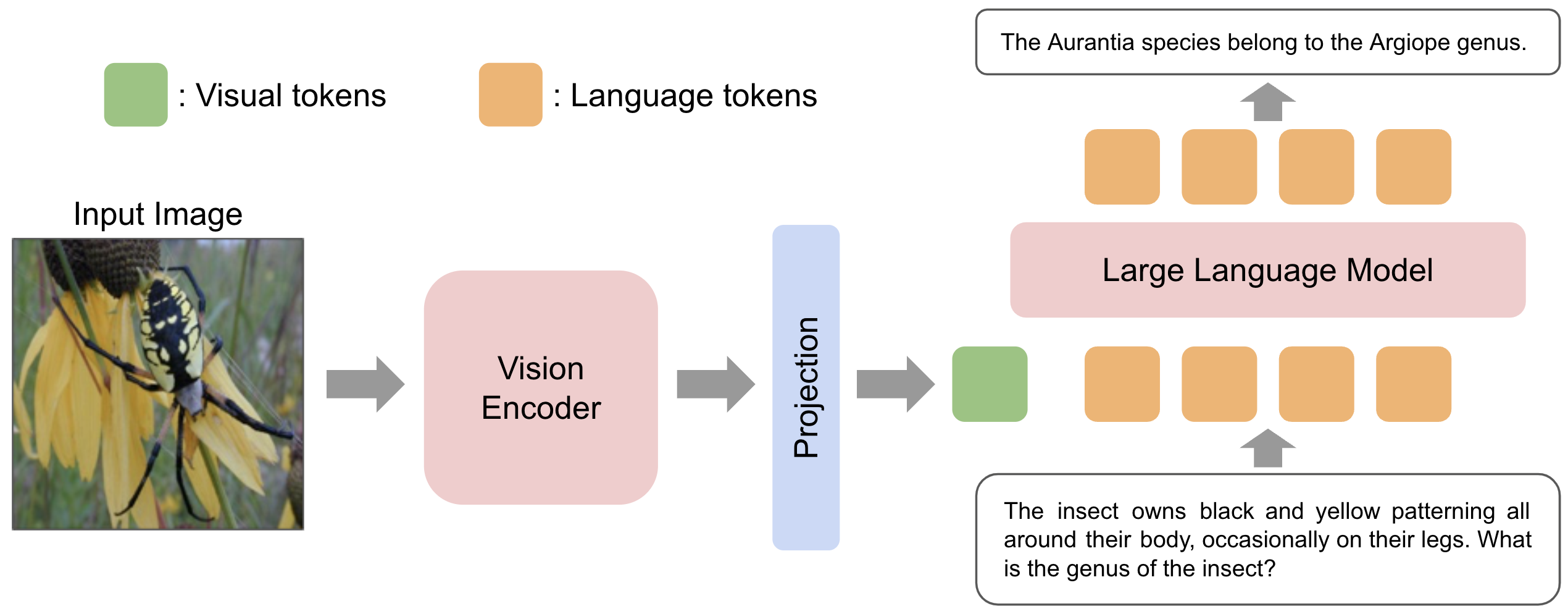}
    \caption{The Large Language and Vision Assistant Framework for Insect Understanding (Insect-LLaVA).}
    \label{fig:insect-llava}
\end{figure}

The conversational generative model relies on the power of both a pre-trained vision encoder and LLM. Inspired by LLaVA \cite{liu2024visual, liu2024improved}, in our approach, we develop our Insect-LLaVA (Figure \ref{fig:insect-llava}) by adopting the design of LLaVA. 
In particular, our Insect-LLaVA model consists of three major models, i.e., the vision encoder, the multi-layer perception connector (also known as the projection), and the large language model. 
For the LLM, we adopt the Vicuna model \cite{vicuna2023} as similar to LLaVA \cite{liu2024visual, liu2024improved}.
Meanwhile, the transformer-based vision encoder can produce meaningful features for insect understanding. 
Since the visual features of insects and language tokens lie on different feature spaces, the visual features will undergo a multi-layer perception connector to align with the textual features.

Formally, given the insect image $I$ and a multi-turn conversation data $\left(\mathbf{X}_q^1,\mathbf{X}_a^1, \mathbf{X}_q^2,\mathbf{X}_a^2, ..., \mathbf{X}_q^M,\mathbf{X}_a^M\right)$ where $M$ is the number of turns, following the standard protocol of \cite{liu2024visual, liu2024improved}, we construct the instruction data $\mathbf{X}^m_\text{instruct}$ in the $m^{th}$ turn as follows:
\begin{equation}\label{eqn:x_instruct_data}
    \mathbf{X}^m_\text{instruct} = \begin{cases}
        \text{Randomly Choose } [\mathbf{X}_q^1, I] \text{ or } [I, \mathbf{X}_q^1] & \text{If } m = 1 \\
        \mathbf{X}_q^m & \text{If } m > 1 
    \end{cases}
\end{equation}
Then, learning the conversational generative model can be formed as an auto-regressive training objective as follows:
\begin{equation}\label{eqn:insect_llava_learning}
\begin{split}
    \theta^* &= \arg\max_{\theta} \mathbb{E}_{\mathbf{X}_a, I, \mathbf{X}_\text{instruct}} \log p(\mathbf{X}_a | I, \mathbf{X}_\text{instruct}) \\
    &= \arg\max_{\theta} \mathbb{E}_{\mathbf{X}_a, I, \mathbf{X}_\text{instruct}} \sum_{i=1}^L \log p_{\theta}(x_i | I, \mathbf{X}_{\text{instruct} < i}, \mathbf{X}_{a < i})
\end{split}
\end{equation}
where $L$ is the sequence length of the answer $\mathbf{X}_a = [x_1, x_2, ..., x_L]$, $\theta$ is the parameters of the Insect-LLaVA model, and $\mathbf{X}_{\text{instruct} < i}$ and $\mathbf{X}_{a < i}$ are the tokens of instructions and answers in all turns before token $x_i$.
Similar to LLaVA \cite{liu2024visual, liu2024improved}, the Insect-LLaVA model follows two stages of the instruction-tuning procedure including pre-training and fine-tuning.

\begin{itemize}
    \item
    \textbf{Visual Insect Feature Alignment Pre-training.}
    To learn the alignment between visual feature features and language descriptions, we construct the pre-training dataset by considering each sample as single-turn instruction data. 
    In particular, for each insect sample, the question $\mathbf{X}_a$ will be randomly constructed, a sentence asking to briefly describe the image. Then, the corresponding answer $\mathbf{X}_a$ will be the insect description developed in our Insect-1M dataset.
    During the pre-training phase, the weights of the multi-layer perception connector are updated, while the weights of the vision encoder and the large language model are fixed. 

    \vspace{2mm}
    
    \item
    \textbf{Visual Insect Instruction Fine-tuning.}
    In the training phase, the entire Insect-LLaVA model is learned from our proposed insect instruction data. Both the large language model and the multi-layer perception connector are further updated. Meanwhile, since the Insect Foundation encoder has been well learned to represent the insect features, the weights of the vision encoder are frozen in this learning stage.
\end{itemize}

\begin{figure}[!b]
\centering
\includegraphics[width=1.0\linewidth]{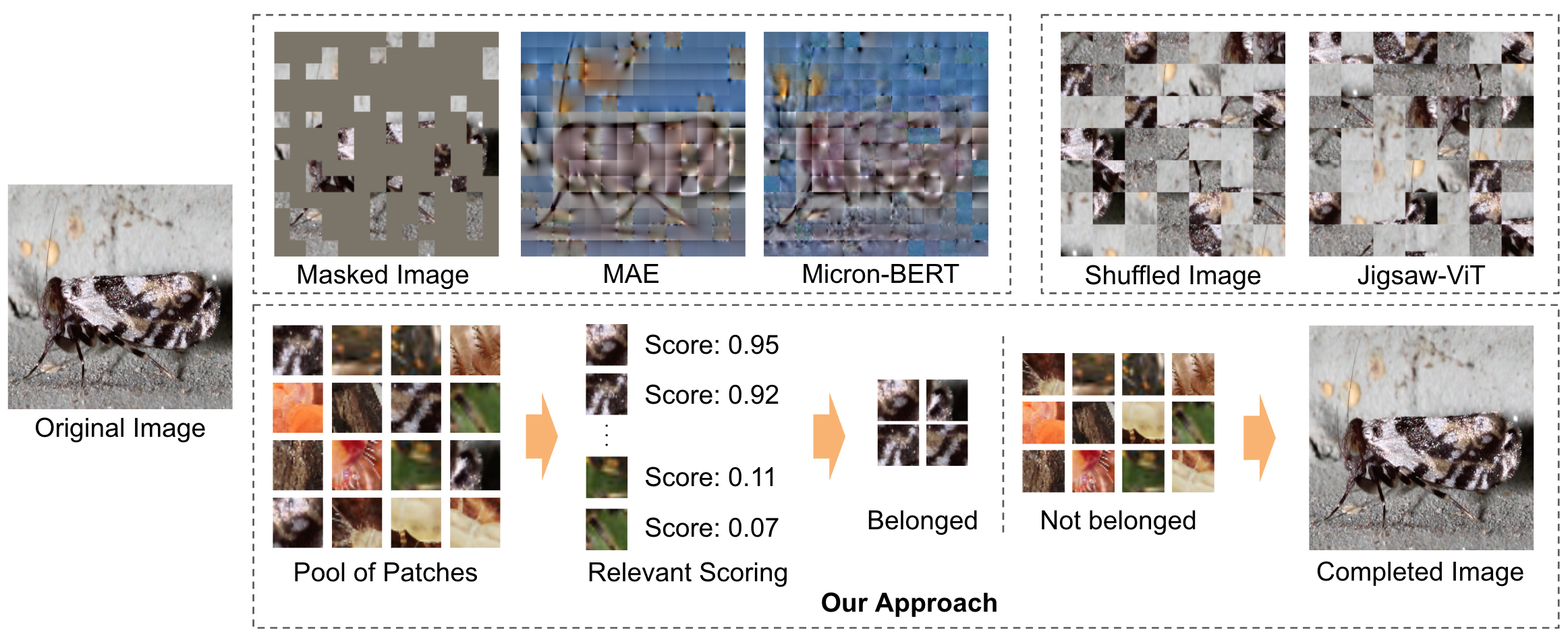}
\caption{\textbf{Comparisons of Different Self-supervised Foundation Model Training Methods.} MAE \cite{he2022masked} fails to reconstruct the details of the insect since it learns general information about the image. Micron-BERT \cite{nguyen2023micron} hardly distinguishes the insect and background. Jigsaw-ViT \cite{chen2023jigsaw} cannot correct shuffled patches due to confusion between the background and the object. Meanwhile, our approach can find separated patches belonging to the insect by scoring each patch.
\textbf{Best viewed in color.}
}
\label{fig:methods_comparison}
\end{figure}

\noindent
\textbf{Limitations of Current Vision Encoder in Insect-LLaVA.}
The success of large language and vision assistant models relies on the vision encoder, which produces a meaningful feature representation of visual knowledge. While the prior approaches \cite{liu2024visual, liu2024improved} adopt the CLIP model \cite{radford2021learning} for the vision encoder, this model remains inefficient in the visual insect understanding context. 
Indeed, the CLIP model is learned on the general knowledge dataset. Meanwhile, the vision encoder in Insect-LLAVA should be capable of specific knowledge of insects so that it can contain features that represent insect characteristics.
It is essential to develop an \textbf{Insect Foundation Model} that captures the insect features well and provides meaningful representations for visual insect understanding purposes.
Therefore, in the next section, we will present our approach to developing an insect foundation model that can be used as a powerful vision encoder in our Insect-LLaVA framework.

\subsection{Insect Foundation Model}

In this section, we first present the current limitations of foundation model learning approaches in developing insect foundation models. Then, we present our proposed learning approaches to develop a robust insect foundation model that can model the micro features of insects.

\subsubsection{Limitations of Prior Foundation Training Approaches}

In visual insect understanding, the visual representation and discrimination of the small and undistinguished features of the insects are the main challenges when building foundation models.
A popular self-supervised learning method MAE \cite{he2022masked} focuses on the context of an image individually while learning to reconstruct an image from the masked image. This learning strategy is hard to represent the small details to discriminate between insects.
On the other hand, self-supervised learning methods using Jigsaw solving \cite{noroozi2016unsupervised,chen2023jigsaw} correct the position of shuffled image patches to learn the image structure. This strategy requires more processes to focus on the undistinguished details of the image.
Meanwhile, Micron-BERT \cite{nguyen2023micron} represents the minor differences in images by swapping the regions between two images with similar contexts.
However, the minor differences in the insect image still maintain the key features representing the insect, which makes the model fail to recognize the small features of the insects.
To overcome these limitations, we propose a novel approach that trains a model to extract the minor features in insect images.
Our method distinguishes these features from the background by learning the micro differences between image patches.
Figure \ref{fig:methods_comparison} shows a comparison of previous self-supervised techniques \cite{he2022masked,nguyen2023micron,chen2023jigsaw} with our method.

\begin{figure}[!t]
\begin{center}
\includegraphics[width=\linewidth]{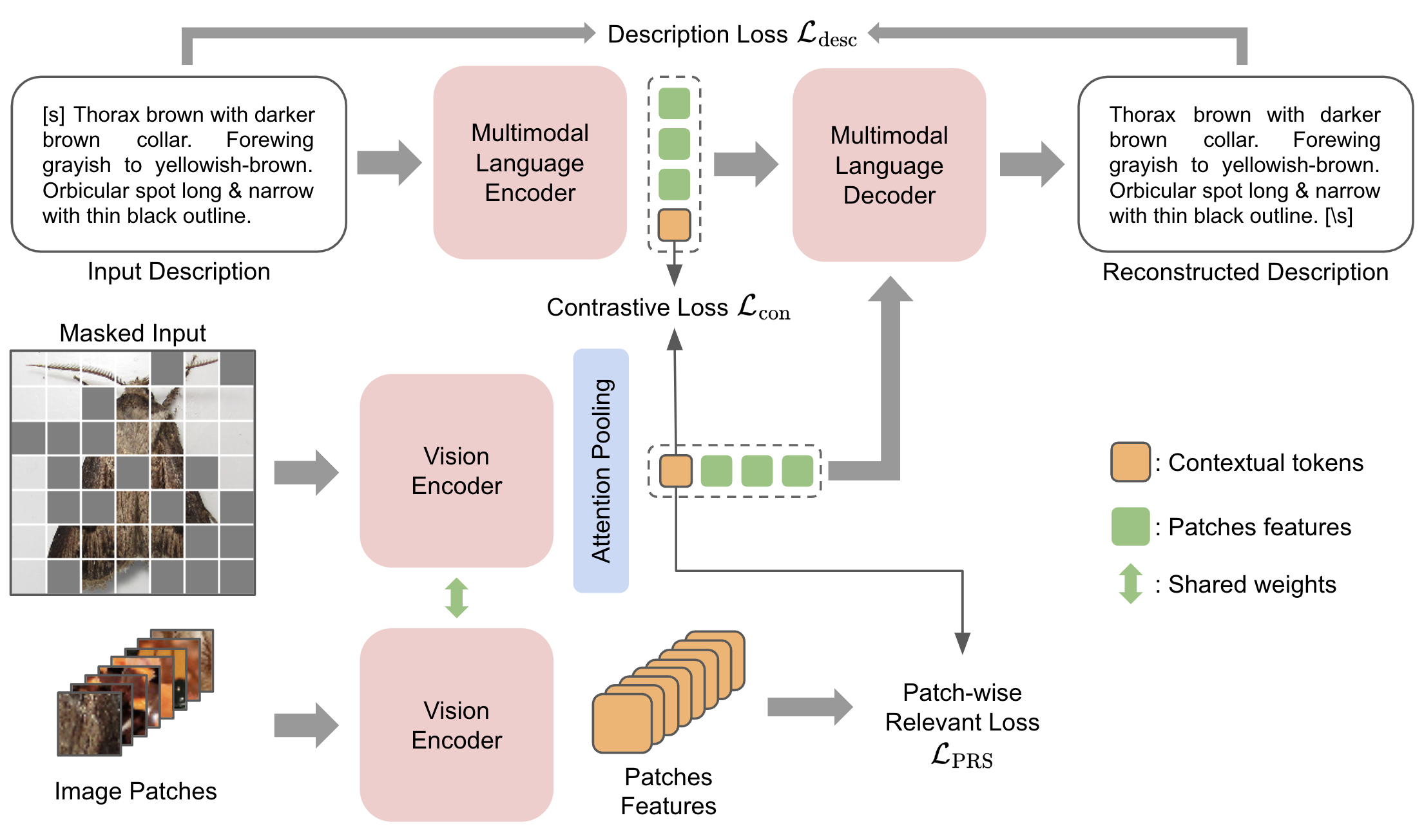}
\end{center}
\caption{
The Overview Framework of Our Proposed Approach to Insect Foundation Model. 
}
\label{fig:overview_framework}
\end{figure}

Figure \ref{fig:overview_framework} presents our insect foundation model.
The model is built to capture minor differences in insect features, including textures, limbs, and other parts, through a novel self-supervised pre-text task. 
Additionally, the model is pre-trained to learn the fine-grained alignment between insect descriptions in text and their visual information.
In detail, given an input image $I$, we divide $I$ into non-overlapping patches $P$.
Then, a subset of patches $P_s$ is sampled randomly from $P$, while the remaining patches are placed into a pool of image patches $P_{\text{pool}}$.
An image encoder is applied to extract $P_s$ into latent vectors.
For the corresponding insect description $T$, a text encoder is used to extract relevant information.
Then, a text decoder and a joint image-text contrastive learning module are employed to map the description to the image.
Finally, a novel Patch-wise Relevant Attention module is introduced for self-supervised learning to enhance the discrimination capability of the model.

\subsubsection{Insect Foundation Model}

\noindent
\textbf{Input Processing.}
Let $P = \{p_s^i\}_{i=1}^{N_P}$ be non-overlapping patches divided from the input image $I \in \mathbb{R}^{H \times W \times 3}$ where $N_P$ is the total number of patches and $H$ and $W$ represent the height and width of the input image.
The patches $P$ are then randomly sampled to create a subset of patches $P_s \subset P$, while the other patches are put into a global pool of image patches $P_{\text{pool}}$. 
Note that $P_{\text{pool}}$ includes patches from multiple images in the training set.

\vspace{2mm}

\noindent
\textbf{Vision Encoder.}
Each patch $p_s^i \in P_s$ is projected into a latent vector $\mathbf{x}_s^i \in \mathbb{R}^d$ where $d$ is defined as the dimension of the latent vectors.
A subset patches $P_s$ can be expressed as follows:
\begin{equation}
    \mathbf{X}_s = \text{concat}(\{\mathbf{x}_s^i\}_{i=1}^{N_{P_s}}) \in \mathbb{R}^{N_{P_s} \times d}, \quad
    \mathbf{x}_s^i = \alpha_p(p_s^i) + \mathbf{e}_p(i)
\end{equation}
where $\alpha_p$ represents the projection embedding, and $\mathbf{e}_p$ denotes the position embedding.
Then, 
let an image encoder $\mathcal{E}_{\text{image}}(\mathbf{X}_s)$ consist of $L_e$ transformer blocks where each block includes multi-head self-attention $\mathcal{F}_\text{MSA}$ and multi-layer perceptron $\mathcal{F}_\text{MLP}$. The Insect Foundation Model can be formed as follows:
\begin{equation}
\begin{split}
    \mathbf{X}^\prime_l &= \mathbf{X}_{l-1} + \mathcal{F}_\text{MSA}(\mathbf{X}_{l-1}) \\
    \mathbf{X}_l &= \mathbf{X}^\prime_l + \mathcal{F}_\text{MLP}(\mathbf{X}^\prime_l) \\
    \mathbf{X}_0 &= \mathbf{X}_s, \: 1 \leq l \leq L_e
\end{split}
\end{equation}
Then, the output latent vector $\mathbf{Z}_s$ is computed from $\mathbf{X}_s$ as follows:
\begin{equation}
    \mathbf{Z}_s = \mathcal{E}_{\text{image}}(\mathbf{X}_s), \quad \mathbf{Z}_s \in \mathbb{R}^{N_{P_s} \times d}
\end{equation}

\subsubsection{Insect Micro-feature Self-supervised Learning}

\begin{figure}[!t]
\begin{center}
\includegraphics[width=\linewidth]{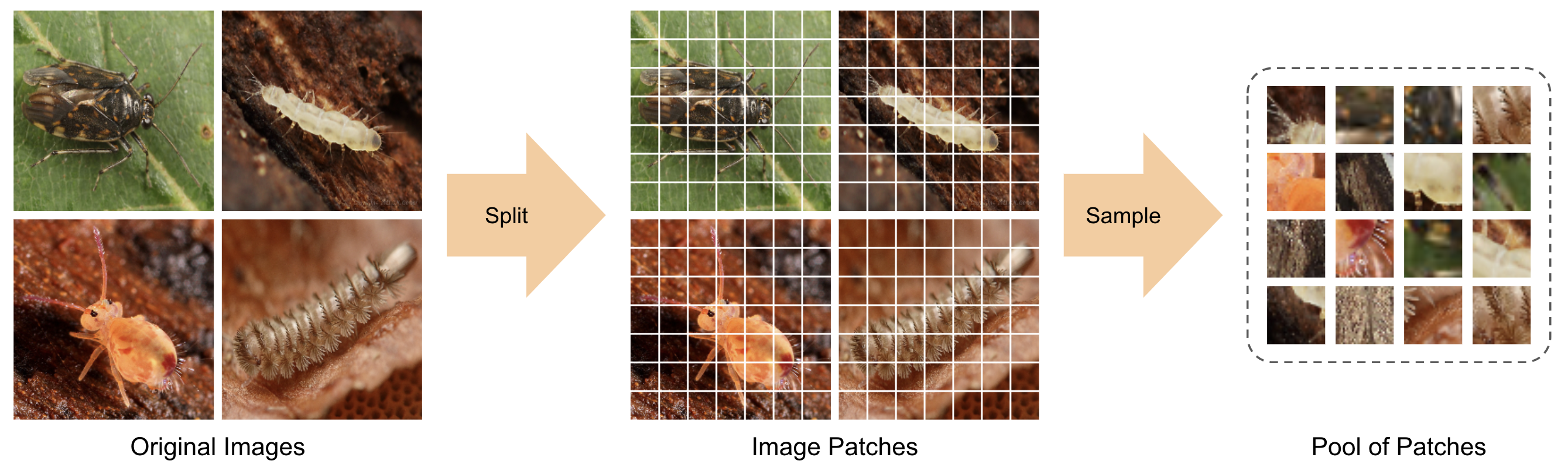}
\end{center}
\caption{\textbf{Pool of Image Patches.} Original images are split into image patches. Then, these patches are randomly sampled and placed into a pool of patches for the self-supervised pre-text task.}
\label{fig:pool_of_patches}
\end{figure}

Insect recognition depends on small features like texture, eyes, or limbs, which are challenging to detect.
To improve the robustness of the model to these tiny features in insect images, we propose a novel self-supervised learning strategy that identifies minor differences in the images to capture these features.
Insects can be distinguished by detecting and recognizing key features across their various parts.
To strengthen this capability, a pre-text task is introduced.
In detail, after extracting global information from a masked insect image, the model learns to identify the missing patches by comparing image segments from different insect species.
Through this learning mechanism, the model effectively recognizes the key features of each insect and distinguishes the small differences between species.
As shown in Figure \ref{fig:pool_of_patches}, given a subset of patches $P_s$ from the image $I$ and a pool of image patches $P_{\text{pool}}$, we train the model to match the patches $p_t \in P_{\text{pool}}$ to the belonging image $I$.
A patch-wise relevance score (PRS) is then computed between the latent vectors $\mathbf{Z}_s$ of $P_s$ and each patch $p \in P_{\text{pool}}$.
The score can be defined as $f_{\text{PRS}}(\mathbf{Z}_s, p) \in [0, 1]$.
The higher the score is, the more possibility that $p \in P$.

\vspace{2mm}

\begin{figure}[!t]
\begin{center}
\includegraphics[width=0.85\linewidth]{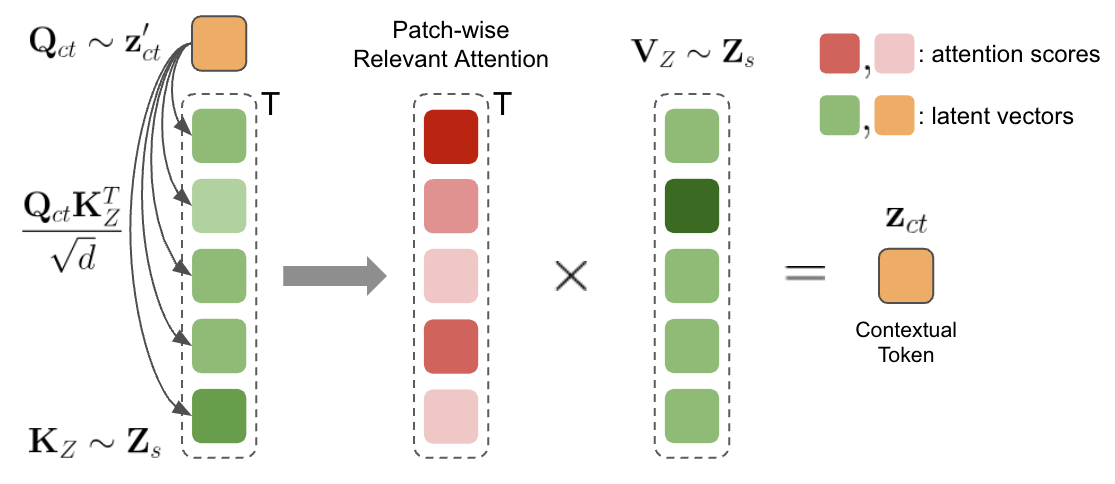}
\end{center}
\caption{\textbf{Attention Pooling Module.} The contextual token $\mathbf{z}_{ct}$ represents the global information of the image $I$.}
\label{fig:attention_pooling}
\end{figure}

\noindent
\textbf{Attention Pooling.}
To measure the relevance between latent vectors $\mathbf{Z}_s$ of the image $I$ and the patch $p \in P_{\text{pool}}$, the latent vectors $\mathbf{Z}_s$ need to be aggregated to capture the overall information of $I$.
Inspired by \cite{yu2022coca}, we utilize attention pooling to compute the global representation of $I$.
Using a placeholder contextual token $\mathbf{z}_{ct}^\prime$ as the query $\mathbf{Q}_{ct}$ and the latent vectors $\mathbf{Z}_s$ as the key $\mathbf{K}_{Z}$ and value $\mathbf{V}_{Z}$, we calculate an attention map between $\mathbf{Q}_{ct}$ and $\mathbf{K}_{Z}$.
The attention map is then used to be combined with the value $\mathbf{V}_{Z}$ to compute a contextual token $\mathbf{z}_{ct}$ representing the global information of $I$. 
The attention pooling illustrated as Figure \ref{fig:attention_pooling} can be described as Eqn. \eqref{eq:attn_pool}.
\begin{equation}
\begin{split}
    \mathbf{Q}_{ct} &= \text{Linear}(\mathbf{z}_{ct}^\prime) \quad \mathbf{K}_{Z} = \text{Linear}(\mathbf{Z}_s) \quad \mathbf{V}_{Z} = \text{Linear}(\mathbf{Z}_s) \\
    \mathbf{z}_{ct} &= \text{softmax}\left(\frac{\mathbf{Q}_{ct} \mathbf{K}_{Z}^T}{\sqrt{d}}\right) \mathbf{V}_{Z}
\end{split}
\label{eq:attn_pool}
\end{equation}

\noindent
\textbf{Patch-wise Relevant Attention.}
With $\mathbf{z}_{ct}$ as a contextual token extracted from the information of $I$, we calculate the relevance between $\mathbf{z}_{ct}$ and $p \in P_\text{pool}$.
We expand the attention score function $f_{\text{PRS}}$ as described in Eqn. \eqref{eq:prs}.
\begin{equation}
    f_{\text{PRS}}(\mathbf{Z}_s, p) = \mathcal{H}(\mathbf{z}_{ct}, \mathbf{z}_p)
\label{eq:prs}
\end{equation}
where $\mathbf{z}_p = \mathcal{E}_\text{image}(\alpha_p(p))$ is a latent vector representing the patch $p$ and $\mathcal{H}$ is a similarity function between two latent vectors.
Based on Eqn. \eqref{eq:prs}, we extent the score function into a self-supervised loss function $\mathcal{L}_{\text{PRS}}$ as follow:
\begin{equation}
    \mathcal{L}_{\text{PRS}} = - y \log(\mathcal{H}(\mathbf{z}_{ct}, \mathbf{z}_p)) - (1 - y) \log(1 - \mathcal{H}(\mathbf{z}_{ct}, \mathbf{z}_p))
\end{equation}
where $y = 1$ if $p \in P$ and $y = 0$ otherwise.
{In this work, the cosine similarity is applied for the similarity function $\mathcal{H}$.}

\subsubsection{Fine-grained Insect Image-Text Alignment}

Each species has a unique definition and description that can be mapped to different parts of the insect image.
We use a text decoder to generate species descriptions based on the insect images.
Additionally, to capture the overall characteristics of each species, we apply contrastive learning between the global features of the insect images and their descriptions.
This enables the model to learn specific details from the insect images through the corresponding descriptions.

Formally, an insect description is tokenized into $T = \{t_i\}_{i=1}^{N_T}$ while $N_T$ represents the number of tokens of the description.
Each token $t_i \in T$ is then embedded as a latent vector $\mathbf{w}_i \in \mathbb{R}^d$.
The description can be represented as:
\begin{equation}
    \mathbf{W} = \text{concat}(\{\mathbf{w}_i\}_{i=1}^{N_T}) \in \mathbb{R}^{N_T \times d}, \quad
    \mathbf{w}_i = \alpha_w + \mathbf{e}_w(i)
\label{eq:text_input_model}
\end{equation}
where $\alpha_w$ and $\mathbf{e}_w$ are the projection embedding and position embedding.

Similar to the image encoder, the text encoder $\mathcal{E}_\text{text}(\mathbf{W})$ consists of $L^\prime_e$ transformer blocks, each containing multi-head self-attention and multi-layer perceptron.
The output latent vector $\mathbf{Z}^\prime$ of the description is computed as
\begin{equation}
    \mathbf{W}^\prime = \mathcal{E}_\text{text}(\mathbf{W}), \quad \mathbf{Z}^\prime \in \mathbb{R}^{N_T \times d}
\end{equation}
Then, we utilize the latent vector $\mathbf{Z}_s$ from the insect image and $\mathbf{W}^\prime$ from the description text for image-text contrastive learning and multi-modal image description decoding.

\vspace{2mm}

\noindent
\textbf{Image-text Contrastive Learning.}
Inspired by the previous language model approaches \cite{devlin2018bert,liu2019roberta,lewis2019bart,raffel2020exploring}, a contextual token $\mathbf{w}_{ct}$ representing the semantic information of the description is added to the beginning of $\mathbf{W}$, as shown in Eqn. \eqref{eq:text_input_model}.
The two encoders $\mathcal{E}_\text{image}$ and $\mathcal{E}_\text{text}$ are then jointly optimized via contrastive learning as follow:
\begin{equation}
\begin{split}
    \mathcal{L}_\text{con} &= \frac{-1}{N} \sum_{i=1}^N \left[\log 
    \frac{\exp( \mathbf{z}_i^T \mathbf{w}_i )}{\sum_{j=1}^N \exp( \mathbf{z}_i^T \mathbf{w}_j )} + \log 
    \frac{\exp( \mathbf{w}_i^T \mathbf{z}_i )}{\sum_{j=1}^N \exp( \mathbf{w}_i^T \mathbf{z}_j )}\right]
\end{split}
\label{eq:contrastive_text_image}
\end{equation}
where $\mathbf{z}_i$ and $\mathbf{w}_i$ are the contextual token of the $i$-th insect image and description, respectively.

\vspace{2mm}

\noindent
\textbf{Multi-modal Image Description Decoding.}
While image-text contrastive learning extracts the global semantic information between the image and description, the multi-modal image description decoding focuses on the fine-grained details by predicting the tokenized texts of $T$ in an autoregressive manner, as described in Eqn. \eqref{eq:description_loss}.
\begin{equation}
    \mathcal{L}_\text{desc} = 
    - \sum_{t=1}^{N_T} \log \mathcal{D}_\text{multi} (\mathbf{w}_t | \mathbf{W}_{0:t-1}, \mathbf{Z}_s)
\label{eq:description_loss}
\end{equation}
where $\mathcal{D}_\text{multi}$ is an autoregressive multi-modal text decoder.

\section{Experiments}

In this section, we first present our implementation and evaluation benchmarks used in our experiments.
Then, we present the experimental results of our proposed Insect-LLaVA and Insect Foundation Model by comparing our results with prior state-of-the-art models. 
Finally, our ablative experiments will analyze the effectiveness of different aspects of our proposed approach.

\subsection{Evaluation Benchmarks and Implementation}

\begin{figure}[!t]
    \centering
    \includegraphics[width=1.0\linewidth]{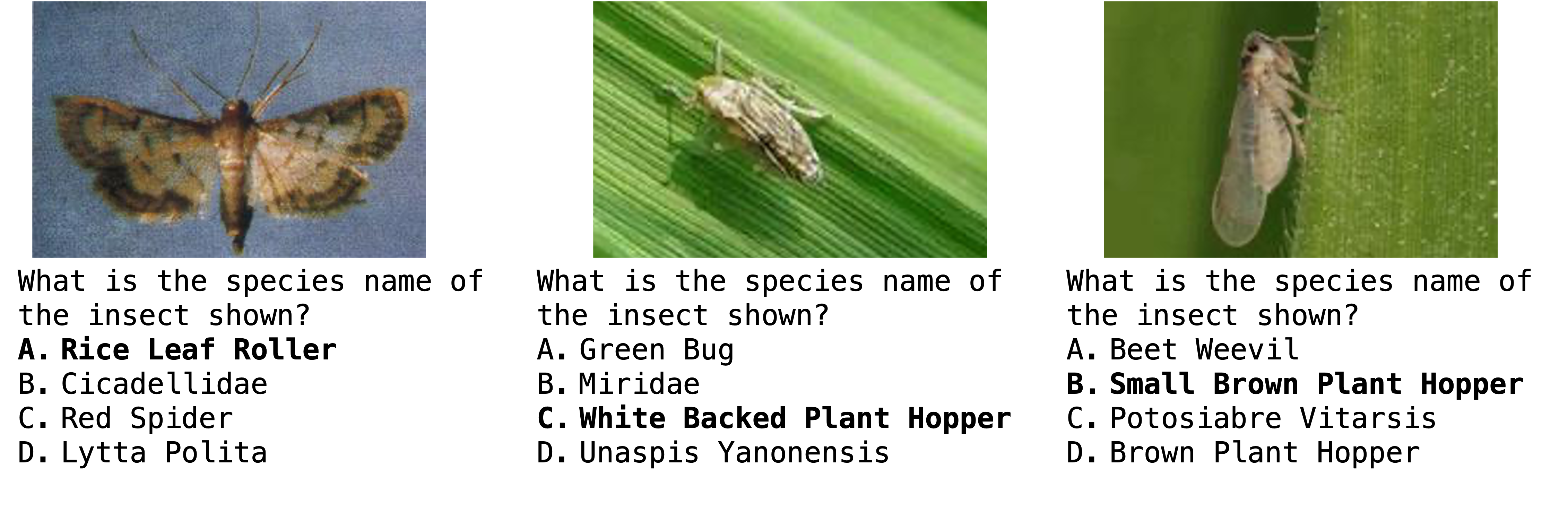}
    \caption{The Examples of Our Visual Insect Question Answering Benchmark.}
    \label{fig:ip-vqa-sample}
\end{figure}

\begin{table}[!b]
\centering
\caption{Accuracy (\%) on Visual Insect Question Answering Benchmark based on IP102.}
\label{tab:insect-llava}
\begin{tabular}{lllcc}
\toprule
Method & Vision Encoder      & Language Model & Training Dataset       & Accuracy \\
\midrule
LLaVA      & CLIP              & Vicuna 7B      & LLaVA V1.5    & 38.14   \\
LLaVA      & CLIP              & Vicuna 13B     & LLaVA V1.5    &   40.20\\
\midrule
LLaVA      & CLIP              & Vicuna 7B      & Multimodal Insect &   42.09 \\
LLaVA      & CLIP              & Vicuna 13B     & Multimodal Insect &   45.10 \\
\midrule
\textbf{Insect-LLAVA}      & Insect-Foundation & Vicuna 7B      & Multimodal Insect &   \textbf{46.08} \\
\textbf{Insect-LLAVA}      & Insect-Foundation & Vicuna 13B     & Multimodal Insect &  \textbf{48.53} \\
\bottomrule
\end{tabular}
\end{table}

\begin{table}[!t]
\centering
\caption{Accuracy (\%) on Visual Insect Question Answering Benchmark based on Multimodal Insect.}
\label{tab:insect-llava-v2}
\begin{tabular}{lllcc}
\toprule
Method       & Vision Encoder    & Language Model & Training Dataset  & Accuracy \\
\hline
LLaVA        & CLIP              & Vicuna 7B      & LLaVA v1.5        & 41.8     \\
LLaVA        & CLIP              & Vicuna 13B     & LLaVA v1.5        & 42.6     \\
\hline
LLaVA        & CLIP              & Vicuna 7B      & Multimodal Insect & 54.4     \\
LLaVA        & CLIP              & Vicuna 13B     & Multimodal Insect & 56.5     \\
\hline
\textbf{Insect-LLaVA} & Insect-Foundation & Vicuna 7B      & Multimodal Insect & \textbf{55.7}     \\
\textbf{Insect-LLaVA} & Insect-Foundation & Vicuna 13B     & Multimodal Insect & \textbf{57.2}    \\
\bottomrule
\end{tabular}
\end{table}

\noindent
\blue{
\textbf{Visual Insect Question Answering Benchmark.} To efficiently evaluate the LLaVA model, we develop the two sets of Visual Insect Question Answering benchmarks based on the IP-102 dataset and our Multimodal Insect dataset.
For the first set, inspired by the Science-VQA \cite{lu2022learn} benchmark, we adopt the insect images and ground truths in the testing set of the IP-102 dataset to develop the multiple-choice questions.
For each insect image, the question will be ``What is the species name of the insect shown?" four choices will be given, with only one being the answer.
In this benchmark, the performance of the model is evaluated using accuracy calculated based on the predictions of conversational generative models. Figure \ref{fig:ip-vqa-sample} illustrates examples of our visual insect question-answering benchmark.
For the second set, we sample 10,000 images from our Multimodal Insect dataset. Then, we develop a set of multi-choice questions similar to the first set based on our ground truths.
}

\noindent
\textbf{IP102 Classification Benchmark.} The IP102 dataset \cite{wu2019ip102} contains 102 insect species and includes 45,095 training samples, 7,508 validation samples, and 22,619 testing samples.
Each species image might depict a single insect, multiple insects, or even a diseased crop caused by the species.
Insects are presented in various life stages for each class, including egg, larva, pupa, and adult.
The classification performance is evaluated using Top 1 accuracy (Acc@1) and Top 5 accuracy (Acc@5).

\vspace{2mm}

\noindent
\textbf{iNat-2021 Insect Classification Benchmark.} This data is a subset of iNat-2021 \cite{van2021benchmarking} which includes 723K images of 2752 different species. Similar to the IP102 Classification benchmark, we adopt the 
metrics of  Top 1 accuracy (Acc@1) and Top 5 accuracy (Acc@5) to evaluate our proposed approach.

\vspace{2mm}

\noindent
\textbf{IP102 Detection Benchmark.} The IP102 detection data \cite{wu2019ip102} provides 15,178 training images and 3,798 testing images representing 102 different species.
Following the COCO benchmark \cite{lin2014microsoft}, the insect detection performance is assessed by Average Precision (AP) and Average Precision at IoU thresholds of 0.5 (AP$^{.50}$) and 0.75 (AP$^{.75}$).

\noindent
{
\textbf{Fine-Grained Zero-Shot Learning on INSECT Benchmark.}
The INSECT Benchmark \cite{badirli2021fine} includes $21,212$ samples with $1,213$ classes.
These classes are split into 1,080 seen classes and 121 unseen classes for evaluation.
In addition, the dataset utilizes DNA sequences as auxiliary information.
Following \cite{badirli2021fine}, the accuracy metric is applied for the seen and unseen classes.
Moreover, the harmonic mean between the accuracies is utilized for comparison.
}

\vspace{2mm}

\noindent
\textbf{Implementation Details.}
In our experiments, our Insect-LLaVA model utilizes the Vicuna 7B and Vicuna 13B for our language model. We adopt the learning hyper-parameter of LLaVA v1.5 \cite{liu2024improved} for our pre-training and fine-tuning phases.
For the vision encoder, our Insect Foundation model uses ViT-Base (ViT-B/16) \cite{dosovitskiy2020image} as the backbone model.
The images are resized and randomly cropped to a resolution of $224 \times 224$.
Each image is then split into $16 \times 16$ patches, resulting in $N_P = 196$ patches.
A patch sampling ratio of $50\%$ is applied, with the remaining patches placed into the pool of image patches.
Each patch is projected into a latent space of $d = 768$ dimensions.
The text encoder and multi-modal text decoder are based on the pre-trained BERT model \cite{devlin2018bert}.
The model is implemented using PyTorch \cite{paszke2019pytorch} and trained on $16 \times \text{A100}$ GPUs.
The initial learning rate is set to $1.5 \times 10^{-4}$, following the cosine learning rate scheduler \cite{loshchilov2016sgdr}.
The model optimization uses AdamW \cite{loshchilov2017decoupled} for 200 epochs, with a batch size of 64 per GPU.

\subsection{Visual Insect Question Answering}

\begin{figure}[!b]
    \centering
    \includegraphics[width=1.0\textwidth]{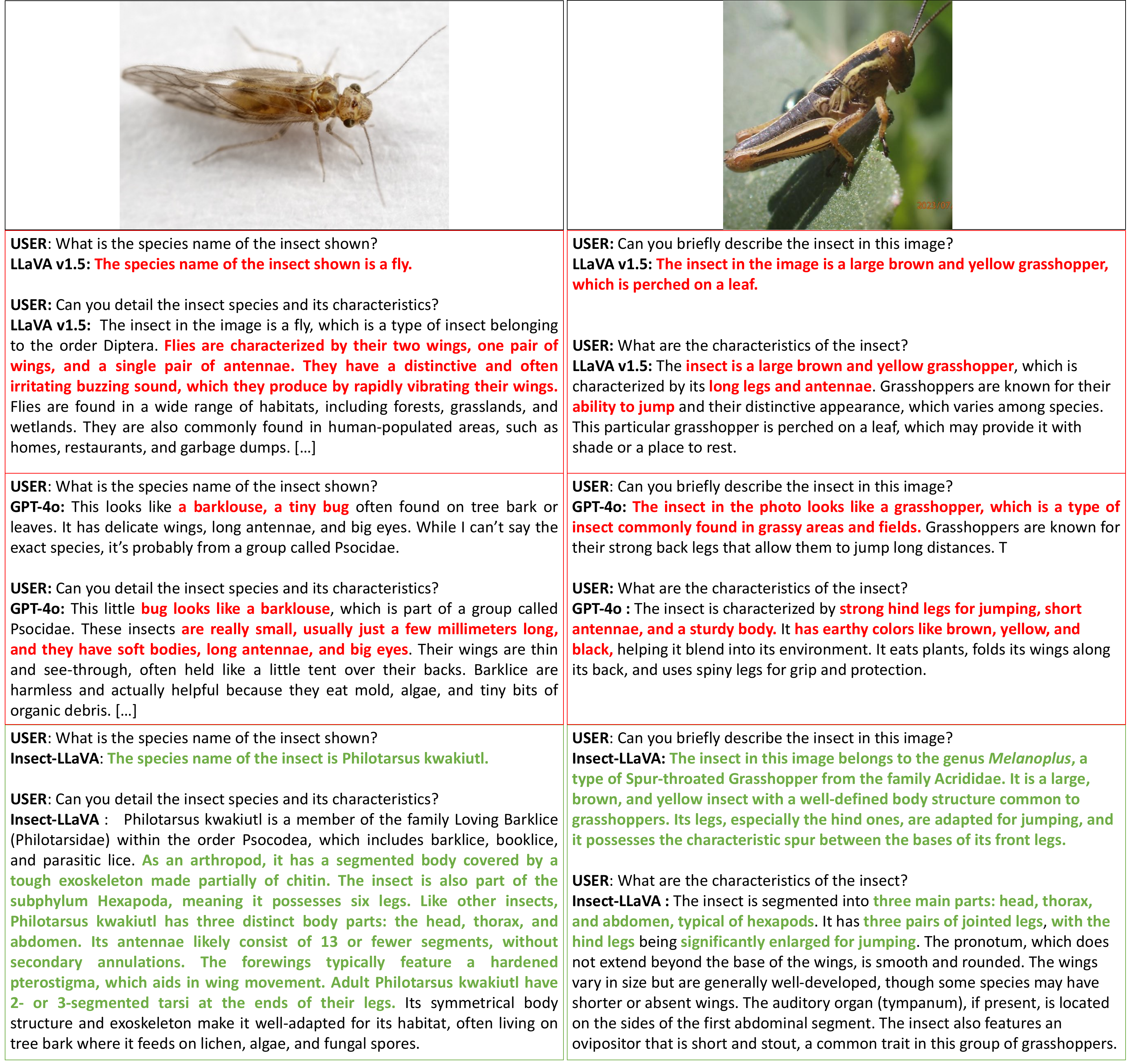}
    \caption{The Comparison of Conversation of Insect Understanding Between LLaVA v1.5, GPT-4o, and \cite{liu2024improved} and Insect LLaVA. As shown in our results, our model can produce more accurate information about the insects with better details and reasoning of insect characteristics. Meanwhile, LLaVA v1.5 is only able to provide very general information about the insects.}
    \label{fig:insect-chat}
\end{figure}

\noindent
\textbf{Visual Insect Question Answering.}
\blue{
We evaluate the performance of our Insect-LLaVA model using our developed Visual Insect Question Answering (Insect VQA) benchmarks using different language models, i.e., Vicuna 7B and Vicuna 13B.
We compare our approach with LLaVA v1.5 \cite{liu2024improved} trained on the LLaVA v1.5 instruction data and our insect instruction data.
Table \ref{tab:insect-llava} illustrates our results on the Insect VQA benchmark developed based on IP102. By using our insect instruction data, the performance of the LLaVA model is further improved since our data provide better and more detailed knowledge of visual insect understanding compared to the general knowledge of the LLaVA v1.5 instruction data.
Meanwhile, if we adopt the Insect Foundation model as the vision encoder, the performance of Insect-LLaVA models is significantly improved since the Insect Foundation model trained on Insect-1M offers better visual feature representations of insects and well captures micro-features of insects.
Similarly, as shown in Table \ref{tab:insect-llava-v2}, our proposed approach and dataset also achieved state-of-the-art performance on the Insect VQA benchmark developed based on our multimodal insect dataset.
}

\vspace{2mm}

\noindent
\textbf{Multimodal Chatbox.} Figure \ref{fig:insect-chat} illustrates examples of the conversations produced by our Insect-LLaVA model compared to the LLaVA model. As shown in our results, while  LLaVA v1.5 only provides a general knowledge of insects, our Insect-LLaVA offers a better response with comprehensive knowledge of insects. 
The qualitative experimental results in Figure \ref{fig:insect-chat} further confirm the effectiveness of our proposed approach and insect instruction data in both the quality of insect knowledge and the ability of insect modeling of the foundation model.

\begin{table*}[!t]
\centering
\caption{
\textbf{Classification results on IP102 Classification benchmark.} Both proposed models, pre-trained with and without the insect descriptions, outperform prior methods by a large margin.
$^*$Since the vision encoder is used for the inference only. We only compute the model size and FLOPS of the vision encoder.
}\label{tab:ip102_classification}
\resizebox{1.0\linewidth}{!}{
\begin{tabular}{lcccccc}
\Xhline{2\arrayrulewidth}
\textbf{Method}  & \makecell{\textbf{Model} \\  \textbf{Size}} & \textbf{FLOPS} & \textbf{Description} & \makecell{\textbf{Pre-train} \\  \textbf{Data}}  & \makecell{\textbf{Acc@1} \\ \textbf{(\%)}} & \makecell{\textbf{Acc@5} \\ \textbf{(\%)}}  \\
\hline
ResNet \cite{wu2019ip102} & 26M & 8.7G & \xmark & ImageNet1K & 49.4 & - \\
EfficientNet \cite{bollis2020weakly} & 30M & 9.9G & \xmark & ImageNet1K & 60.7 & - \\
DenseNet \cite{nanni2020insect} & 33M & 11.5G & \xmark & ImageNet1K & 61.9 & - \\
GAEnsemble \cite{ayan2020crop} & - & - & \xmark & ImageNet1K & 67.1 & - \\
WS-SAM \cite{he2024weakly} & 26M & 8.7G & \xmark & ImageNet1K & 71.1 & 87.6 \\
ViT \cite{dosovitskiy2020image} & 87M & 17.6G & \xmark & ImageNet1K & 71.6 & 87.7 \\
\hline
SINet \cite{fan2020camouflaged} & 26M & 8.7G & \xmark & COD10K \cite{fan2020camouflaged} & 71.2 & 89.8 \\
FEDER \cite{he2023camouflaged} & 26M & 8.7G & \xmark & COD10K \cite{fan2020camouflaged} & 72.5 & 90.8 \\
\hline
MoCo \cite{he2020momentum} & 87M & 17.6G & \xmark & Insect-1M & 70.6 & 88.4 \\
DINO \cite{caron2021emerging} & 87M & 17.6G & \xmark & Insect-1M & 71.5 & 91.4 \\
MAE \cite{he2022masked} & 87M & 17.6G & \xmark & Insect-1M & 72.0 & 91.5 \\
\textbf{Insect-Foundation} & 87M & 17.6G & \xmark &  Insect-1M & \textbf{73.3} & \textbf{91.6} \\
\hline
CoCa \cite{yu2022coca} & 87M$^*$ & 17.6G$^*$ & \cmark & Insect-1M & 72.8 & 91.1 \\
\textbf{Insect-Foundation} & 87M$^*$ & 17.6G$^*$ & \cmark &  Insect-1M & \textbf{75.8} & \textbf{92.1} \\
\Xhline{2\arrayrulewidth}
\end{tabular}
}
\end{table*}

\subsection{The Impact of Insect Foundation Model}

To further illustrate the superior performance of our proposed Insect Foundation model in our Insect-LLaVA, we conduct experiments to evaluate it and compare it with prior methods on classification, detection, and zero-shot classification benchmarks.

\vspace{2mm}
\noindent
\textbf{IP102 Insect Classification Task.}
We fine-tune the linear layer of our pre-trained model on the IP102 dataset \cite{wu2019ip102} for the insect classification task.
As presented in Table \ref{tab:ip102_classification}, our model outperforms prior deep learning models \cite{krizhevsky2012imagenet,szegedy2015going,simonyan2014very,he2016deep,dosovitskiy2020image} pre-trained on ImageNet \cite{deng2009imagenet} by a large margin.
Compared to other methods \cite{he2020momentum,caron2021emerging, he2022masked,yu2022coca} pre-trained on the proposed Insect-1M dataset, our model shows better performance with accuracy scores of $73.3\%$ without insect descriptions and $75.8\%$ when using insect descriptions.
These results indicate that our proposed approach provides a better visual representation of insect images than previous pre-training methods on the same dataset.
\blue{
In addition, we compared our results with prior approaches tailored for camouflaged objects \cite{he2024weakly, fan2020camouflaged, he2023camouflaged}. As shown in the results, our approach outperforms existing camouflaged methods. This demonstrates the capability of our model in effectively capturing the micro-features of insects, offering superior performance compared to prior methods. 
}

\vspace{2mm}

\noindent
\textbf{iNat-2021 Insect Classification Task.} To further illustrate the effectiveness of our proposed approach, we further conduct experiments to evaluate our model on the Insect subset of iNat-2021 \cite{van2021benchmarking} for the classification task on the full training dataset.
As shown in Table \ref{tab:inat21_classification}, compared to prior foundation model training methods, i.e., MAE and CoCa, our proposed approach outperforms these methods by a large margin, i.e., the accuracy of our Insect Foundation model has been increased from $87.52\%$ to $89.23\%$ without taxonomic descriptions and from $88.22\%$ to $90.40\%$ with descriptions.

\begin{figure}[!t]
    \centering
    \includegraphics[width=0.9\linewidth]{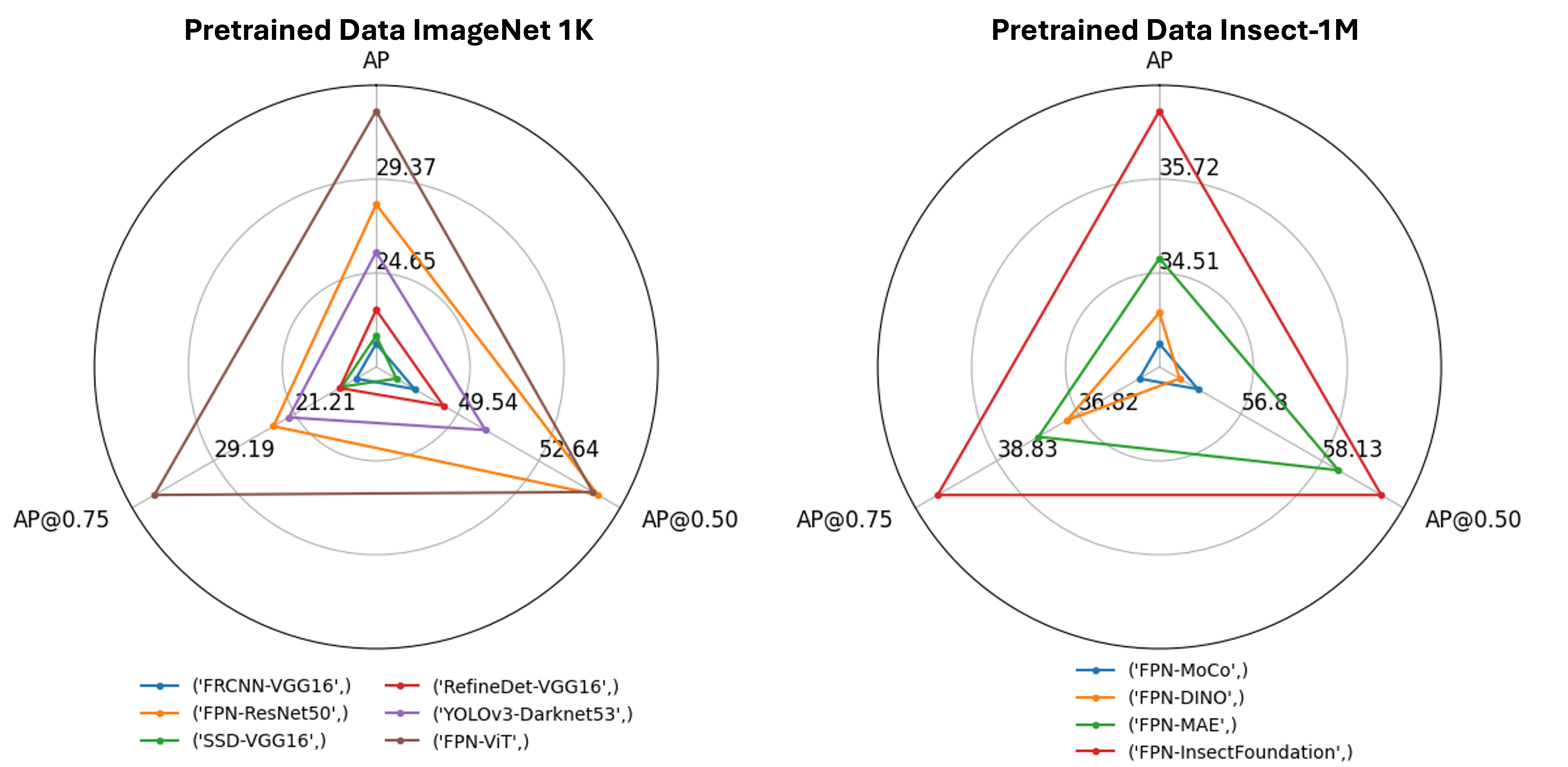}
    \caption{\textbf{Detection results on IP102 Detection benchmark.} 
    Our proposed model outperforms prior detection methods, e.g., FRCNN \cite{ren2015faster}, FPN \cite{lin2017feature}, SSD300 \cite{liu2016ssd}, RefineDet \cite{zhang2018single}, YOLOv3 \cite{redmon2018yolov3}, with different backbones VGG-16 \cite{simonyan2014very}, ResNet-50 \cite{he2016deep}, DarkNet-53 \cite{redmon2018yolov3}, ViT \cite{dosovitskiy2020image}, MoCo \cite{he2020momentum}, DINO \cite{caron2021emerging}, MAE \cite{he2022masked}.}\label{fig:ip102_detection}
\end{figure}

\begin{table}[!b]
\vspace{-4mm}
\centering
\caption{\textbf{Classification results on iNat-2021 Insect Benchmark \cite{van2021benchmarking}.} Both proposed models, pre-trained with and without the insect descriptions, outperform prior methods by a large margin.}\label{tab:inat21_classification}

    \begin{tabular}{lcccccc}

    \toprule
    \textbf{Method}  & \makecell{\textbf{Model}\\\textbf{Size}}& \textbf{GFLOPS} &\textbf{Description} & \makecell{\textbf{Pre-train} \\  \textbf{Data}}  & \makecell{\textbf{Acc@1} \\ \textbf{(\%)}} & \makecell{\textbf{Acc@5} \\ \textbf{(\%)}}  \\
    
    \hline
    Vit-B/16 \cite{dosovitskiy2020image} & 87M & 17.6G & \xmark & ImageNet1K & 87.00 & 96.21 \\
    
    \hline
    
    MAE \cite{he2022masked} & 87M & 17.6G & \xmark & Insect-1M & 87.52 & 96.42 \\
    
    \textbf{Insect-Foundation} & 87M & 17.6G & \xmark & {Insect-1M} & \textbf{89.23} & \textbf{96.88} \\
    
    \hline
    
    CoCa \cite{yu2022coca} & 87M & 17.6G & \cmark & Insect-1M & 88.22 & 96.70 \\
    
    \textbf{Insect-Foundation} & 87M & 17.6G & \cmark & {Insect-1M} & \textbf{90.40} & \textbf{97.36} \\
    
    \bottomrule
    \end{tabular}
\end{table}

\vspace{2mm}

\noindent
\textbf{IP102 Insect Detection Task.}
As presented in Figure \ref{fig:ip102_detection}, we train a Faster R-CNN model \cite{ren2015faster} on the IP102 Detection dataset with the pre-trained ViT backbone adapted for FPN \cite{lin2017feature}.
Our model achieves state-of-the-art results, with an average precision of $36.6\%$ and $\text{AP}^{.50}$ of $59.1\%$, surpassing the same backbone pre-trained on ImageNet \cite{deng2009imagenet}, which had an AP of $32.8\%$ and $\text{AP}^{.50}$ of $54.7\%$.
Compared to other self-supervised methods \cite{he2020momentum,caron2021emerging,he2022masked}, our model demonstrates higher precision, indicating better focus on insect features than previous approaches.

\vspace{2mm}

\begin{wraptable}[12]{r}{0.5\textwidth}%
\centering
\caption{\textbf{Zero-shot classification results on IP102 Classification benchmark.} The proposed model outperforms prior vision-language pre-training methods.}
\setlength{\tabcolsep}{2pt}
\resizebox{1.0\linewidth}{!}{
\begin{tabular}{lccccc}
\Xhline{2\arrayrulewidth}
Method & \makecell{Model\\Size} & GFLOPS & \makecell{Pre-train\\Data} & Acc@1 (\%)  & {Acc@5} (\%) \\
\hline
CLIP \cite{radford2021learning}  & 87M & 17.6G  & Insect-1M     & 41.1 & {65.2}  \\
LiT  \cite{zhai2022lit}  & 87M & 17.6G & Insect-1M & 43.6 & {68.7}  \\
CoCa \cite{yu2022coca}  & 87M & 17.6G & Insect-1M & 45.3 & {72.1}  \\
\hline
\textbf{Ours}   &    87M & 17.6G & Insect-1M & \textbf{49.9} & {\textbf{75.4}}  \\
\Xhline{2\arrayrulewidth}
\end{tabular}
}
\label{tab:zero_shot_classification}
\end{wraptable}
\noindent
\textbf{Zero-shot Insect Classification Task.}
We evaluate the performance of our model on the IP102 Classification dataset \cite{wu2019ip102} using a zero-shot approach.
For each species, a description is provided to the text encoder, which extracts semantic information. 
The image encoder then extracts global features from each insect image and compares them to the description features to predict the species.
Table \ref{tab:zero_shot_classification} shows that our model achieves an accuracy of $49.9\%$, outperforming previous image-text pre-training methods \cite{radford2021learning,zhai2022lit,yu2022coca}. This demonstrates a strong alignment between insect images and their descriptions.

\vspace{2mm}

\begin{table}[t]
    \centering
    \resizebox{0.9\linewidth}{!}{
    \begin{tabular}{lccc}
        \Xhline{2\arrayrulewidth}
        \textbf{Backbone} & \textbf{Unseen Acc (\%)} & \textbf{Seen Acc (\%)} & \textbf{Harmonic Mean (\%)} \\
        \hline
        ResNet101 \cite{he2016deep} & 20.8 & 38.3 & 27.0 \\
        ViT-Base \cite{dosovitskiy2020image} & 21.6 & 44.3 & 29.1 \\
        \hline
        \textbf{Ours} & \textbf{24.1} & \textbf{50.2} & \textbf{32.6} \\
        \Xhline{2\arrayrulewidth}
    \end{tabular}
    }
    \caption{\textbf{Fine-Grained Zero-Shot Learning on INSECT Benchmark \cite{badirli2021fine}.} The proposed model outperforms prior visual backbones.}
    \label{tab:fine_grained_zsl}
\end{table}

\noindent
{
\textbf{Fine-Grained Zero-Shot Learning on INSECT Benchmark.}
To evaluate the robustness of the proposed model, we conduct an experiment for a fine-grain zero-shot learning problem with DNA sequences as side information \cite{badirli2021fine}.
As shown in Table \ref{tab:fine_grained_zsl}, our model improves the accuracies from $21.6\%$ to $24.1\%$ for the unseen classes and from $44.3\%$ to $50.2\%$ for the seen classes.
}

\begin{figure}[!t]
\begin{center}
\includegraphics[width=\linewidth]{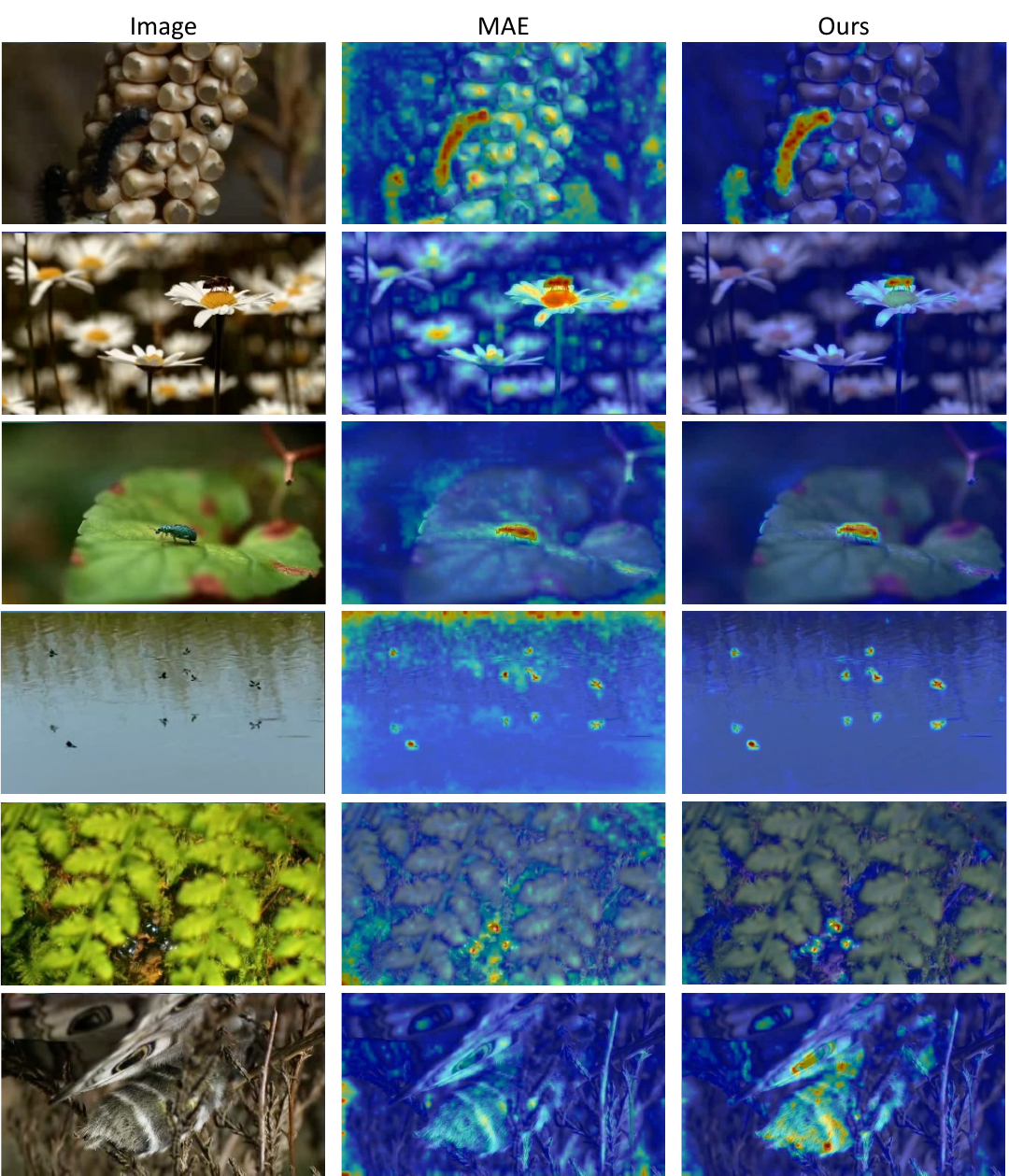}
\end{center}
\caption{\textbf{Attention Visualization.} Compared to MAE \cite{he2022masked}, our model is robust to small details of insect images. The model can focus on the small textures of the insect, even if the texture is hard to see. \textbf{Best viewed in color.}}
\label{fig:attention_visualization}
\end{figure}

\noindent
\textbf{Visualization Results.} 
Figure \ref{fig:attention_visualization} presents attention map visualizations of our model compared to MAE \cite{he2022masked} pre-trained on the proposed dataset.
Since the textures of the insects are hard to see in the background, MAE struggles to focus on the small details of the insects.
In contrast, our model can effectively detect the key features of the insects.

\noindent
{
\textbf{Computational Analysis.}
As shown in Table \ref{tab:ip102_classification}, the proposed model has a computational cost and model size similar to conventional Vision Transformers \cite{dosovitskiy2020image}.
In detail, the vision encoder of the Insect-Foundation model contains about 87M parameters with 17.6G floating-point operations per second.
While our model maintains a similar computation to the original Vision Transformer, our Insect Foundation Models consistently achieve state-of-the-art performance across the network backbones and benchmarks.
}

\subsection{Ablation Studies}

We conducted ablation experiments to evaluate the effectiveness of our proposed model and hyper-parameters on the IP102 Classification Benchmark, as presented in Table \ref{tab:patches_sampling_ratio} and Table \ref{tab:abl_studies}.

\vspace{2mm}

\noindent
\textbf{How does Patch Sampling Ratio Affect the Performance?}
The experimental results in Table \ref{tab:patches_sampling_ratio} illustrate the effectiveness of the patch sampling ratio to the model performance.
The evaluation shows that the sampling ratio of $50\%$ is the best ratio when the lower ratio of $25\%$ prevents the model from having sufficient information for pre-training.
Meanwhile, higher ratios, i.e., $75\%$ and $90\%$, weaken the learning ability of the model.

\vspace{2mm}

\begin{wraptable}[10]{r}{0.5\textwidth}
\centering
\vspace{-10mm}
\caption{\textbf{Effectiveness of Patch Sampling Ratio.} We evaluate the impact of the sampling ratio on IP102 Classification \cite{wu2019ip102} with four sampling ratio, i.e. $25\%$, $50\%$, $75\%$, and $90\%$.}
\label{tab:patches_sampling_ratio}
\resizebox{0.5\textwidth}{!}{
\begin{tabular}{ccccc}
\toprule
\makecell{\textbf{Sampling}\\\textbf{Ratio}} & 25\% & \textbf{50\%} & 75\% & 90\% \\
\midrule
\textbf{Acc@1 (\%)} & 65.2 & \textbf{73.3} & 72.1 & 69.8 \\
\textbf{Acc@5 (\%)} & 82.8 & \textbf{91.6} & 90.9 & 88.3 \\
\bottomrule
\end{tabular}
}
\end{wraptable}
\noindent
\textbf{Do Network Backbones Improve Performance?}
Table \ref{tab:abl_studies} examines the impact of different Vision Transformer backbone sizes, including ViT-small/16, ViT-base/16, and ViT-large/16.
Our results demonstrate that more robust backbones lead to more significant improvements.
Specifically, when increasing the Transformer backbone from small to base, the accuracy rises significantly by $4.3\%$, while switching to a large backbone further enhances accuracy by $1.1\%$.

\begin{table*}[!b]
\centering
\caption{\textbf{Effectiveness of our method on the IP102 Classification.} We evaluate approach with three different vision transformer backbones, i.e., ViT-small/16, ViT-base/16, and ViT-large/16, without or with Attention Pooling (Attn Pool), and three different losses, i.e. Patch-wise Relevant Loss ($\mathcal{L}_\text{rel}$), Image-Text Contrastive Loss ($\mathcal{L}_\text{con}$), and Description Loss ($\mathcal{L}_\text{desc}$).}
\resizebox{1.0\linewidth}{!}{
\begin{tabular}{ccc|cccc|cc}
\toprule
\textbf{Backbone} & \makecell{\textbf{Model} \\ \textbf{Size}} & \textbf{GFLOPS} &
  $\mathcal{L}_\text{rel}$ &
  \makecell{\textbf{Attn} \\ \textbf{Pool}} &
  $\mathcal{L}_\text{con}$ &
  $\mathcal{L}_\text{desc}$ &
  \makecell{\textbf{Acc@1} \\ \textbf{(\%)}} &
  \makecell{\textbf{Acc@5} \\ \textbf{(\%)}} \\
\midrule
\multirow{4}{*}{ViT-small/16} & \multirow{4}{*}{22M} & \multirow{4}{*}{4.6G} & \cmark &            &            &            & 68.9 & 88.8 \\
                              & & & \cmark & \cmark &            &            & 69.5 & 89.7 \\
                              & & & \cmark & \cmark & \cmark &            & 70.7 & \textbf{89.9} \\
                              & & & \cmark & \cmark & \cmark & \cmark & \textbf{71.5} & 87.7 \\
\midrule
\multirow{4}{*}{ViT-base/16}  & \multirow{4}{*}{87M} & \multirow{4}{*}{17.6G} & \cmark &            &            &            & 72.4 & 91.0 \\
                              & & & \cmark & \cmark &            &            & 73.3 & 91.6 \\
                              & & & \cmark & \cmark & \cmark &            & 74.2 & 91.9 \\
                              & & & \cmark & \cmark & \cmark & \cmark & \textbf{75.8} & \textbf{92.1} \\
\midrule
\multirow{4}{*}{ViT-large/16} & \multirow{4}{*}{304M} & \multirow{4}{*}{61.6G} & \cmark &            &            &            & 73.8 & 90.9 \\
                              & & & \cmark & \cmark &            &            & 74.6 & 91.6 \\
                              & & & \cmark & \cmark & \cmark &            & 75.9 & 91.4 \\
                              & & & \cmark & \cmark & \cmark & \cmark & \textbf{76.9} & \textbf{92.7} \\
\bottomrule
\end{tabular}
}
\label{tab:abl_studies}
\end{table*}

\vspace{2mm}

\noindent
\textbf{Does Attention Pooling Improve Micro-feature Modeling?}
We assess the effect of attention pooling on the visual representation of insect images. 
As shown in Table \ref{tab:abl_studies}, the Attention Pooling provides better representation than the standard classification token computed through transformer layers.
The top-1 accuracy for the three backbones, i.e., small, base, and large backbones, increases from $68.9\%$ to $69.5\%$, $72.4\%$ to $73.3\%$, and $73.8\%$ to $74.6\%$, respectively.

\vspace{2mm}

\noindent
\textbf{Does Image-Text Contrastive Loss Matter?}
As shown in Table \ref{tab:abl_studies}, the model can extract the information of the insect images better when the model learns to match with their descriptions.
Specifically, the accuracy increases by $0.8\%$, $0.9\%$, and $1.3\%$ for the small, base, and large backbones, respectively, with the application of Image-Text Contrastive Loss.

\vspace{2mm}

\noindent
\textbf{Does Description Loss Promote Performance?}
The complete configuration results shown in Table \ref{tab:abl_studies} highlight the performance of our model using the Description Loss.
This loss function helps the model better align image information with descriptive details, allowing it to capture fine-grained insect features more effectively.
The accuracy improves from $70.7\%$ to $71.5\%$, $74.2\%$ to $75.8\%$, and $75.9\%$ to $76.9\%$ for ViT-small/16, ViT-base/16, and ViT-large/16, respectively.

\vspace{2mm}

\noindent
\blue{
\textbf{Does Multi-Turn Visual Instruction Data Matter?} 
We conducted an ablation study to examine the impact of multi-turn instruction data on our model performance. Specifically, we derived a single-turn dataset from our existing multi-turn instruction data by isolating individual interactions. We then trained the model using this single-turn dataset and compared its performance against the model trained on the original multi-turn dataset. As shown in Table \ref{tab:abl-turn-data}, incorporating multi-turn data leads to a significant improvement in model performance. This enhancement can be attributed to the model’s ability to better capture contextual dependencies and develop advanced reasoning capabilities when exposed to multi-turn instructions. Meanwhile, using the single-turn instruction data will result in lower performance since it lacks continuity and depth in instruction-following.
}

\begin{figure}[!b]
    \centering
    \includegraphics[width=0.9\linewidth]{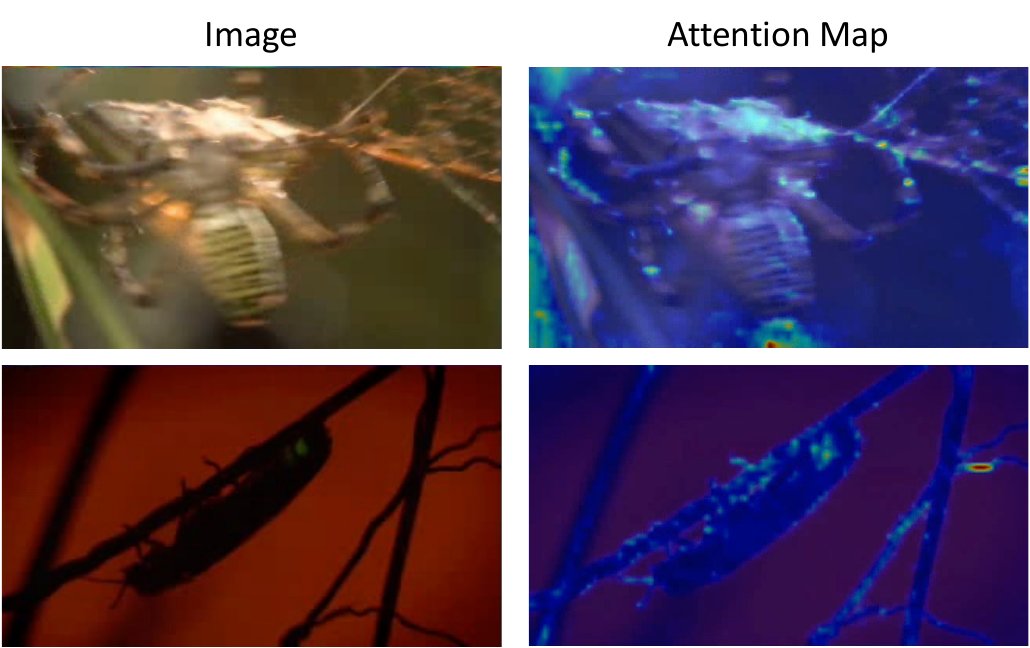}
    \caption{Failure cases analysis under extreme conditions.}
    \label{fig:failure_cases}
\end{figure}

\noindent
{
\textbf{Failure Cases Analysis. }
Figure \ref{fig:failure_cases} illustrates failure cases where the model cannot focus on the subject.
In extreme cases, e.g., blurring or low-light conditions, the model struggles to accurately capture the important patterns and distinguishing features of the insect samples.
Blurring can obscure critical micro-features of insects, such as wing venation, body segmentation, or fine textures, which are essential for species differentiation. 
Low-light conditions further degrade the visibility of important details, making it difficult for the model to extract meaningful visual features. 
}

\begin{table}[!t]
\centering
\caption{Effectiveness of Multi-Turn Instruction Data on Visual Insect Question Answering Benchmark.}
\label{tab:abl-turn-data}
\begin{tabular}{lllcc}
\toprule
Method       & Vision Encoder    & Language Model & Training Dataset & Accuracy \\
\hline
LLaVA        & CLIP              & Vicuna 7B      & Single-Turn      & 40.29    \\
LLaVA        & CLIP              & Vicuna 7B      & Multi-Turn       & 42.09    \\
\hline
LLaVA        & CLIP              & Vicuna 13B     & Single-Turn      & 41.61    \\
LLaVA        & CLIP              & Vicuna 13B     & Multi-Turn       & 45.10    \\
\hline
Insect-LLaVA & Insect-Foundation & Vicuna 7B      & Single Turn      & 43.63    \\
Insect-LLaVA & Insect-Foundation & Vicuna 7B      & Multi-Turn       & 46.08    \\
\hline
Insect-LLaVA & Insect-Foundation & Vicuna 13B     & Single Turn      & 45.57    \\
Insect-LLaVA & Insect-Foundation & Vicuna 13B     & Multi-Turn       & 48.53   \\
\bottomrule
\end{tabular}
\end{table}

\section{Conclusions and Limitations}

\textbf{Conclusions.}
This paper has presented a new large-scale Multimodal Insect dataset with Visual Insect Instruction Data aimed at developing large foundation models for precision agriculture.
Our proposed dataset contains a wide range of insect species annotated with multi-level taxonomy labels.
Importantly, our Multimodal Insect data offers detailed descriptions and visual instruction data that enable the vision-language training for large-scale multimodal foundation models.
In addition to our proposed data, we have introduced an Insect-LLaVA model, a new conversational generative model for visual insect understanding.
To enhance the capability of insect feature modeling in Insect-LLaVA, we have proposed an Insect Foundation Model with the Patch-wise Relevant Attention mechanism to better capture the micro features of insects.
Then, we presented a new Description of Consistency loss to further improve our ability to model fine-grained features.
The experimental results have shown that our proposed Insect-LLaVA achieved State-of-the-Art performance on our proposed Visual Insect Question Answering benchmarks.
Our results have also illustrated the importance of our Multimodal Insect dataset in developing vision methods for visual insect understanding tasks.

\vspace{2mm}

\noindent
\textbf{Limitations.} 
This study employed a specific network design and learning parameters to validate our hypothesis.
However, our approach has certain limitations, particularly in the design of the Patch-wise Relevant Attention mechanism, which treats background and foreground patches equally. 
This may hinder the ability to learn the difference between insect features.
Addressing this limitation will inspire future research to refine the Insect Foundation Model and improve Micro-feature Modeling.
In addition, due to the high cost of GPT-4 \cite{achiam2023gpt}, our insect instruction data are generated via the open-source LLaMA-3 \cite{dubey2024llama} for cost efficiency. While the quality of conversation data generated by LLaMA-3 \cite{dubey2024llama} is acceptable due to the competitive performance on our evaluation, we believe that using the commercial language model (e.g., GPT-4 \cite{achiam2023gpt}) could yield better quality and realistic conversations.

\section{Data Availability Statement}

The proposed datasets used in this study are available in the following public domain resources:
\begin{itemize}
    \item \textbf{Insect-1M} \cite{nguyen2024insect}: \url{https://uark-cviu.github.io/projects/insect-foundation/}
    \vspace{2mm}
    \item \textbf{Insect VQA}: \url{https://uark-cviu.github.io/projects/insect-foundation/}
    \vspace{2mm}
    \item \textbf{IP102} \cite{wu2019ip102}: \url{https://github.com/xpwu95/IP102}
    \vspace{2mm}
    \item \textbf{iNat-2021} \cite{van2021benchmarking}: \url{https://github.com/visipedia/inat_comp}
\end{itemize}

The Insect-LLaVA models and Multimodal Insect Dataset with Visual Insect Instruction Data proposed in this paper are available for academic research from the corresponding author upon reasonable request.

\bibliography{sn-bibliography}%


\begin{thebibliography}{105}
\ifx \bisbn   \undefined \def \bisbn  #1{ISBN #1}\fi
\ifx \binits  \undefined \def \binits#1{#1}\fi
\ifx \bauthor  \undefined \def \bauthor#1{#1}\fi
\ifx \batitle  \undefined \def \batitle#1{#1}\fi
\ifx \bjtitle  \undefined \def \bjtitle#1{#1}\fi
\ifx \bvolume  \undefined \def \bvolume#1{\textbf{#1}}\fi
\ifx \byear  \undefined \def \byear#1{#1}\fi
\ifx \bissue  \undefined \def \bissue#1{#1}\fi
\ifx \bfpage  \undefined \def \bfpage#1{#1}\fi
\ifx \blpage  \undefined \def \blpage #1{#1}\fi
\ifx \burl  \undefined \def \burl#1{\textsf{#1}}\fi
\ifx \doiurl  \undefined \def \doiurl#1{\url{https://doi.org/#1}}\fi
\ifx \betal  \undefined \def \betal{\textit{et al.}}\fi
\ifx \binstitute  \undefined \def \binstitute#1{#1}\fi
\ifx \binstitutionaled  \undefined \def \binstitutionaled#1{#1}\fi
\ifx \bctitle  \undefined \def \bctitle#1{#1}\fi
\ifx \beditor  \undefined \def \beditor#1{#1}\fi
\ifx \bpublisher  \undefined \def \bpublisher#1{#1}\fi
\ifx \bbtitle  \undefined \def \bbtitle#1{#1}\fi
\ifx \bedition  \undefined \def \bedition#1{#1}\fi
\ifx \bseriesno  \undefined \def \bseriesno#1{#1}\fi
\ifx \blocation  \undefined \def \blocation#1{#1}\fi
\ifx \bsertitle  \undefined \def \bsertitle#1{#1}\fi
\ifx \bsnm \undefined \def \bsnm#1{#1}\fi
\ifx \bsuffix \undefined \def \bsuffix#1{#1}\fi
\ifx \bparticle \undefined \def \bparticle#1{#1}\fi
\ifx \barticle \undefined \def \barticle#1{#1}\fi
\bibcommenthead
\ifx \bconfdate \undefined \def \bconfdate #1{#1}\fi
\ifx \botherref \undefined \def \botherref #1{#1}\fi
\ifx \url \undefined \def \url#1{\textsf{#1}}\fi
\ifx \bchapter \undefined \def \bchapter#1{#1}\fi
\ifx \bbook \undefined \def \bbook#1{#1}\fi
\ifx \bcomment \undefined \def \bcomment#1{#1}\fi
\ifx \oauthor \undefined \def \oauthor#1{#1}\fi
\ifx \citeauthoryear \undefined \def \citeauthoryear#1{#1}\fi
\ifx \endbibitem  \undefined \def \endbibitem {}\fi
\ifx \bconflocation  \undefined \def \bconflocation#1{#1}\fi
\ifx \arxivurl  \undefined \def \arxivurl#1{\textsf{#1}}\fi
\csname PreBibitemsHook\endcsname

\bibitem[\protect\citeauthoryear{Alayrac et~al.}{2022}]{alayrac2022flamingo}
\begin{barticle}
\bauthor{\bsnm{Alayrac}, \binits{J.-B.}},
\bauthor{\bsnm{Donahue}, \binits{J.}},
\bauthor{\bsnm{Luc}, \binits{P.}},
\bauthor{\bsnm{Miech}, \binits{A.}},
\bauthor{\bsnm{Barr}, \binits{I.}},
\bauthor{\bsnm{Hasson}, \binits{Y.}},
\bauthor{\bsnm{Lenc}, \binits{K.}},
\bauthor{\bsnm{Mensch}, \binits{A.}},
\bauthor{\bsnm{Millican}, \binits{K.}},
\bauthor{\bsnm{Reynolds}, \binits{M.}}, \betal:
\batitle{Flamingo: a visual language model for few-shot learning}.
\bjtitle{Advances in neural information processing systems}
\bvolume{35},
\bfpage{23716}--\blpage{23736}
(\byear{2022})
\end{barticle}
\endbibitem

\bibitem[\protect\citeauthoryear{Yang et~al.}{2023}]{yang2023dawn}
\begin{barticle}
\bauthor{\bsnm{Yang}, \binits{Z.}},
\bauthor{\bsnm{Li}, \binits{L.}},
\bauthor{\bsnm{Lin}, \binits{K.}},
\bauthor{\bsnm{Wang}, \binits{J.}},
\bauthor{\bsnm{Lin}, \binits{C.-C.}},
\bauthor{\bsnm{Liu}, \binits{Z.}},
\bauthor{\bsnm{Wang}, \binits{L.}}:
\batitle{The dawn of lmms: Preliminary explorations with gpt-4v (ision)}.
\bjtitle{arXiv preprint arXiv:2309.17421}
\bvolume{9}(\bissue{1}),
\bfpage{1}
(\byear{2023})
\end{barticle}
\endbibitem

\bibitem[\protect\citeauthoryear{Li et~al.}{2024}]{li2024multimodal}
\begin{barticle}
\bauthor{\bsnm{Li}, \binits{C.}},
\bauthor{\bsnm{Gan}, \binits{Z.}},
\bauthor{\bsnm{Yang}, \binits{Z.}},
\bauthor{\bsnm{Yang}, \binits{J.}},
\bauthor{\bsnm{Li}, \binits{L.}},
\bauthor{\bsnm{Wang}, \binits{L.}},
\bauthor{\bsnm{Gao}, \binits{J.}}, \betal:
\batitle{Multimodal foundation models: From specialists to general-purpose assistants}.
\bjtitle{Foundations and Trends{\textregistered} in Computer Graphics and Vision}
\bvolume{16}(\bissue{1-2}),
\bfpage{1}--\blpage{214}
(\byear{2024})
\end{barticle}
\endbibitem

\bibitem[\protect\citeauthoryear{Liu et~al.}{2024a}]{liu2024visual}
\begin{botherref}
\oauthor{\bsnm{Liu}, \binits{H.}},
\oauthor{\bsnm{Li}, \binits{C.}},
\oauthor{\bsnm{Wu}, \binits{Q.}},
\oauthor{\bsnm{Lee}, \binits{Y.J.}}:
Visual instruction tuning.
Advances in neural information processing systems
\textbf{36}
(2024)
\end{botherref}
\endbibitem

\bibitem[\protect\citeauthoryear{Liu et~al.}{2024b}]{liu2024improved}
\begin{bchapter}
\bauthor{\bsnm{Liu}, \binits{H.}},
\bauthor{\bsnm{Li}, \binits{C.}},
\bauthor{\bsnm{Li}, \binits{Y.}},
\bauthor{\bsnm{Lee}, \binits{Y.J.}}:
\bctitle{Improved baselines with visual instruction tuning}.
In: \bbtitle{Proceedings of the IEEE/CVF Conference on Computer Vision and Pattern Recognition},
pp. \bfpage{26296}--\blpage{26306}
(\byear{2024})
\end{bchapter}
\endbibitem

\bibitem[\protect\citeauthoryear{Li et~al.}{2024}]{li2024llava}
\begin{botherref}
\oauthor{\bsnm{Li}, \binits{C.}},
\oauthor{\bsnm{Wong}, \binits{C.}},
\oauthor{\bsnm{Zhang}, \binits{S.}},
\oauthor{\bsnm{Usuyama}, \binits{N.}},
\oauthor{\bsnm{Liu}, \binits{H.}},
\oauthor{\bsnm{Yang}, \binits{J.}},
\oauthor{\bsnm{Naumann}, \binits{T.}},
\oauthor{\bsnm{Poon}, \binits{H.}},
\oauthor{\bsnm{Gao}, \binits{J.}}:
Llava-med: Training a large language-and-vision assistant for biomedicine in one day.
Advances in Neural Information Processing Systems
\textbf{36}
(2024)
\end{botherref}
\endbibitem

\bibitem[\protect\citeauthoryear{Radford et~al.}{2021}]{radford2021learning}
\begin{bchapter}
\bauthor{\bsnm{Radford}, \binits{A.}},
\bauthor{\bsnm{Kim}, \binits{J.W.}},
\bauthor{\bsnm{Hallacy}, \binits{C.}},
\bauthor{\bsnm{Ramesh}, \binits{A.}},
\bauthor{\bsnm{Goh}, \binits{G.}},
\bauthor{\bsnm{Agarwal}, \binits{S.}},
\bauthor{\bsnm{Sastry}, \binits{G.}},
\bauthor{\bsnm{Askell}, \binits{A.}},
\bauthor{\bsnm{Mishkin}, \binits{P.}},
\bauthor{\bsnm{Clark}, \binits{J.}}, \betal:
\bctitle{Learning transferable visual models from natural language supervision}.
In: \bbtitle{International Conference on Machine Learning},
pp. \bfpage{8748}--\blpage{8763}
(\byear{2021}).
\bcomment{PMLR}
\end{bchapter}
\endbibitem

\bibitem[\protect\citeauthoryear{Li et~al.}{2023}]{li2023blip}
\begin{bchapter}
\bauthor{\bsnm{Li}, \binits{J.}},
\bauthor{\bsnm{Li}, \binits{D.}},
\bauthor{\bsnm{Savarese}, \binits{S.}},
\bauthor{\bsnm{Hoi}, \binits{S.}}:
\bctitle{Blip-2: Bootstrapping language-image pre-training with frozen image encoders and large language models}.
In: \bbtitle{International Conference on Machine Learning},
pp. \bfpage{19730}--\blpage{19742}
(\byear{2023}).
\bcomment{PMLR}
\end{bchapter}
\endbibitem

\bibitem[\protect\citeauthoryear{Zhang et~al.}{2024}]{zhang2024somelvlm}
\begin{botherref}
\oauthor{\bsnm{Zhang}, \binits{X.}},
\oauthor{\bsnm{Kuang}, \binits{H.}},
\oauthor{\bsnm{Mou}, \binits{X.}},
\oauthor{\bsnm{Lyu}, \binits{H.}},
\oauthor{\bsnm{Wu}, \binits{K.}},
\oauthor{\bsnm{Chen}, \binits{S.}},
\oauthor{\bsnm{Luo}, \binits{J.}},
\oauthor{\bsnm{Huang}, \binits{X.}},
\oauthor{\bsnm{Wei}, \binits{Z.}}:
Somelvlm: A large vision language model for social media processing.
arXiv preprint arXiv:2402.13022
(2024)
\end{botherref}
\endbibitem

\bibitem[\protect\citeauthoryear{Gharaee et~al.}{2023}]{gharaee2023step}
\begin{bchapter}
\bauthor{\bsnm{Gharaee}, \binits{Z.}},
\bauthor{\bsnm{Gong}, \binits{Z.}},
\bauthor{\bsnm{Pellegrino}, \binits{N.}},
\bauthor{\bsnm{Zarubiieva}, \binits{I.}},
\bauthor{\bsnm{Haurum}, \binits{J.B.}},
\bauthor{\bsnm{Lowe}, \binits{S.C.}},
\bauthor{\bsnm{McKeown}, \binits{J.T.A.}},
\bauthor{\bsnm{Ho}, \binits{C.Y.}},
\bauthor{\bsnm{McLeod}, \binits{J.}},
\bauthor{\bsnm{Wei}, \binits{Y.C.}},
\bauthor{\bsnm{Agda}, \binits{J.}},
\bauthor{\bsnm{Ratnasingham}, \binits{S.}},
\bauthor{\bsnm{Steinke}, \binits{D.}},
\bauthor{\bsnm{Chang}, \binits{A.X.}},
\bauthor{\bsnm{Taylor}, \binits{G.W.}},
\bauthor{\bsnm{Fieguth}, \binits{P.}}:
\bctitle{A step towards worldwide biodiversity assessment: The {BIOSCAN-1M} insect dataset}.
In: \beditor{\bsnm{Oh}, \binits{A.}},
\beditor{\bsnm{Neumann}, \binits{T.}},
\beditor{\bsnm{Globerson}, \binits{A.}},
\beditor{\bsnm{Saenko}, \binits{K.}},
\beditor{\bsnm{Hardt}, \binits{M.}},
\beditor{\bsnm{Levine}, \binits{S.}} (eds.)
\bbtitle{Advances in Neural Information Processing Systems},
vol. \bseriesno{36},
pp. \bfpage{43593}--\blpage{43619}.
\bpublisher{Curran Associates, Inc.}, \blocation{???}
(\byear{2023}).
\burl{https://proceedings.neurips.cc/paper_files/paper/2023/file/87dbbdc3a685a97ad28489a1d57c45c1-Paper-Datasets_and_Benchmarks.pdf}
\end{bchapter}
\endbibitem

\bibitem[\protect\citeauthoryear{Askell et~al.}{2021}]{askell2021general}
\begin{botherref}
\oauthor{\bsnm{Askell}, \binits{A.}},
\oauthor{\bsnm{Bai}, \binits{Y.}},
\oauthor{\bsnm{Chen}, \binits{A.}},
\oauthor{\bsnm{Drain}, \binits{D.}},
\oauthor{\bsnm{Ganguli}, \binits{D.}},
\oauthor{\bsnm{Henighan}, \binits{T.}},
\oauthor{\bsnm{Jones}, \binits{A.}},
\oauthor{\bsnm{Joseph}, \binits{N.}},
\oauthor{\bsnm{Mann}, \binits{B.}},
\oauthor{\bsnm{DasSarma}, \binits{N.}}, et al.:
A general language assistant as a laboratory for alignment.
arXiv preprint arXiv:2112.00861
(2021)
\end{botherref}
\endbibitem

\bibitem[\protect\citeauthoryear{Gan et~al.}{2022}]{gan2022vision}
\begin{barticle}
\bauthor{\bsnm{Gan}, \binits{Z.}},
\bauthor{\bsnm{Li}, \binits{L.}},
\bauthor{\bsnm{Li}, \binits{C.}},
\bauthor{\bsnm{Wang}, \binits{L.}},
\bauthor{\bsnm{Liu}, \binits{Z.}},
\bauthor{\bsnm{Gao}, \binits{J.}}, \betal:
\batitle{Vision-language pre-training: Basics, recent advances, and future trends}.
\bjtitle{Foundations and Trends{\textregistered} in Computer Graphics and Vision}
\bvolume{14}(\bissue{3--4}),
\bfpage{163}--\blpage{352}
(\byear{2022})
\end{barticle}
\endbibitem

\bibitem[\protect\citeauthoryear{Li et~al.}{2022}]{li2022elevater}
\begin{barticle}
\bauthor{\bsnm{Li}, \binits{C.}},
\bauthor{\bsnm{Liu}, \binits{H.}},
\bauthor{\bsnm{Li}, \binits{L.}},
\bauthor{\bsnm{Zhang}, \binits{P.}},
\bauthor{\bsnm{Aneja}, \binits{J.}},
\bauthor{\bsnm{Yang}, \binits{J.}},
\bauthor{\bsnm{Jin}, \binits{P.}},
\bauthor{\bsnm{Hu}, \binits{H.}},
\bauthor{\bsnm{Liu}, \binits{Z.}},
\bauthor{\bsnm{Lee}, \binits{Y.J.}}, \betal:
\batitle{Elevater: A benchmark and toolkit for evaluating language-augmented visual models}.
\bjtitle{Advances in Neural Information Processing Systems}
\bvolume{35},
\bfpage{9287}--\blpage{9301}
(\byear{2022})
\end{barticle}
\endbibitem

\bibitem[\protect\citeauthoryear{Samanta and Ghosh}{2012}]{samanta2012tea}
\begin{barticle}
\bauthor{\bsnm{Samanta}, \binits{R.}},
\bauthor{\bsnm{Ghosh}, \binits{I.}}:
\batitle{Tea insect pests classification based on artificial neural networks}.
\bjtitle{International Journal of Computer Engineering Science (IJCES)}
\bvolume{2}(\bissue{6}),
\bfpage{1}--\blpage{13}
(\byear{2012})
\end{barticle}
\endbibitem

\bibitem[\protect\citeauthoryear{Wang et~al.}{2012}]{wang2012new}
\begin{barticle}
\bauthor{\bsnm{Wang}, \binits{J.}},
\bauthor{\bsnm{Lin}, \binits{C.}},
\bauthor{\bsnm{Ji}, \binits{L.}},
\bauthor{\bsnm{Liang}, \binits{A.}}:
\batitle{A new automatic identification system of insect images at the order level}.
\bjtitle{Knowledge-Based Systems}
\bvolume{33},
\bfpage{102}--\blpage{110}
(\byear{2012})
\end{barticle}
\endbibitem

\bibitem[\protect\citeauthoryear{Venugoban and Ramanan}{2014}]{venugoban2014image}
\begin{barticle}
\bauthor{\bsnm{Venugoban}, \binits{K.}},
\bauthor{\bsnm{Ramanan}, \binits{A.}}:
\batitle{Image classification of paddy field insect pests using gradient-based features}.
\bjtitle{International Journal of Machine Learning and Computing}
\bvolume{4}(\bissue{1}),
\bfpage{1}
(\byear{2014})
\end{barticle}
\endbibitem

\bibitem[\protect\citeauthoryear{Xie et~al.}{2015}]{xie2015automatic}
\begin{barticle}
\bauthor{\bsnm{Xie}, \binits{C.}},
\bauthor{\bsnm{Zhang}, \binits{J.}},
\bauthor{\bsnm{Li}, \binits{R.}},
\bauthor{\bsnm{Li}, \binits{J.}},
\bauthor{\bsnm{Hong}, \binits{P.}},
\bauthor{\bsnm{Xia}, \binits{J.}},
\bauthor{\bsnm{Chen}, \binits{P.}}:
\batitle{Automatic classification for field crop insects via multiple-task sparse representation and multiple-kernel learning}.
\bjtitle{Computers and Electronics in Agriculture}
\bvolume{119},
\bfpage{123}--\blpage{132}
(\byear{2015})
\end{barticle}
\endbibitem

\bibitem[\protect\citeauthoryear{Liu et~al.}{2016}]{liu2016localization}
\begin{barticle}
\bauthor{\bsnm{Liu}, \binits{Z.}},
\bauthor{\bsnm{Gao}, \binits{J.}},
\bauthor{\bsnm{Yang}, \binits{G.}},
\bauthor{\bsnm{Zhang}, \binits{H.}},
\bauthor{\bsnm{He}, \binits{Y.}}:
\batitle{Localization and classification of paddy field pests using a saliency map and deep convolutional neural network}.
\bjtitle{Scientific reports}
\bvolume{6}(\bissue{1}),
\bfpage{20410}
(\byear{2016})
\end{barticle}
\endbibitem

\bibitem[\protect\citeauthoryear{Xie et~al.}{2018}]{xie2018multi}
\begin{barticle}
\bauthor{\bsnm{Xie}, \binits{C.}},
\bauthor{\bsnm{Wang}, \binits{R.}},
\bauthor{\bsnm{Zhang}, \binits{J.}},
\bauthor{\bsnm{Chen}, \binits{P.}},
\bauthor{\bsnm{Dong}, \binits{W.}},
\bauthor{\bsnm{Li}, \binits{R.}},
\bauthor{\bsnm{Chen}, \binits{T.}},
\bauthor{\bsnm{Chen}, \binits{H.}}:
\batitle{Multi-level learning features for automatic classification of field crop pests}.
\bjtitle{Computers and Electronics in Agriculture}
\bvolume{152},
\bfpage{233}--\blpage{241}
(\byear{2018})
\end{barticle}
\endbibitem

\bibitem[\protect\citeauthoryear{Deng et~al.}{2018}]{deng2018research}
\begin{barticle}
\bauthor{\bsnm{Deng}, \binits{L.}},
\bauthor{\bsnm{Wang}, \binits{Y.}},
\bauthor{\bsnm{Han}, \binits{Z.}},
\bauthor{\bsnm{Yu}, \binits{R.}}:
\batitle{Research on insect pest image detection and recognition based on bio-inspired methods}.
\bjtitle{Biosystems Engineering}
\bvolume{169},
\bfpage{139}--\blpage{148}
(\byear{2018})
\end{barticle}
\endbibitem

\bibitem[\protect\citeauthoryear{Alfarisy et~al.}{2018}]{alfarisy2018deep}
\begin{bchapter}
\bauthor{\bsnm{Alfarisy}, \binits{A.A.}},
\bauthor{\bsnm{Chen}, \binits{Q.}},
\bauthor{\bsnm{Guo}, \binits{M.}}:
\bctitle{Deep learning based classification for paddy pests \& diseases recognition}.
In: \bbtitle{Proceedings of 2018 International Conference on Mathematics and Artificial Intelligence},
pp. \bfpage{21}--\blpage{25}
(\byear{2018})
\end{bchapter}
\endbibitem

\bibitem[\protect\citeauthoryear{Liu et~al.}{2019}]{liu2019pestnet}
\begin{barticle}
\bauthor{\bsnm{Liu}, \binits{L.}},
\bauthor{\bsnm{Wang}, \binits{R.}},
\bauthor{\bsnm{Xie}, \binits{C.}},
\bauthor{\bsnm{Yang}, \binits{P.}},
\bauthor{\bsnm{Wang}, \binits{F.}},
\bauthor{\bsnm{Sudirman}, \binits{S.}},
\bauthor{\bsnm{Liu}, \binits{W.}}:
\batitle{Pestnet: An end-to-end deep learning approach for large-scale multi-class pest detection and classification}.
\bjtitle{Ieee Access}
\bvolume{7},
\bfpage{45301}--\blpage{45312}
(\byear{2019})
\end{barticle}
\endbibitem

\bibitem[\protect\citeauthoryear{Wu et~al.}{2019}]{wu2019ip102}
\begin{bchapter}
\bauthor{\bsnm{Wu}, \binits{X.}},
\bauthor{\bsnm{Zhan}, \binits{C.}},
\bauthor{\bsnm{Lai}, \binits{Y.-K.}},
\bauthor{\bsnm{Cheng}, \binits{M.-M.}},
\bauthor{\bsnm{Yang}, \binits{J.}}:
\bctitle{Ip102: A large-scale benchmark dataset for insect pest recognition}.
In: \bbtitle{Proceedings of the IEEE/CVF Conference on Computer Vision and Pattern Recognition},
pp. \bfpage{8787}--\blpage{8796}
(\byear{2019})
\end{bchapter}
\endbibitem

\bibitem[\protect\citeauthoryear{Wang et~al.}{2021}]{wang2021agripest}
\begin{barticle}
\bauthor{\bsnm{Wang}, \binits{R.}},
\bauthor{\bsnm{Liu}, \binits{L.}},
\bauthor{\bsnm{Xie}, \binits{C.}},
\bauthor{\bsnm{Yang}, \binits{P.}},
\bauthor{\bsnm{Li}, \binits{R.}},
\bauthor{\bsnm{Zhou}, \binits{M.}}:
\batitle{Agripest: A large-scale domain-specific benchmark dataset for practical agricultural pest detection in the wild}.
\bjtitle{Sensors}
\bvolume{21}(\bissue{5}),
\bfpage{1601}
(\byear{2021})
\end{barticle}
\endbibitem

\bibitem[\protect\citeauthoryear{Badirli et~al.}{2021}]{badirli2021fine}
\begin{barticle}
\bauthor{\bsnm{Badirli}, \binits{S.}},
\bauthor{\bsnm{Akata}, \binits{Z.}},
\bauthor{\bsnm{Mohler}, \binits{G.}},
\bauthor{\bsnm{Picard}, \binits{C.}},
\bauthor{\bsnm{Dundar}, \binits{M.M.}}:
\batitle{Fine-grained zero-shot learning with dna as side information}.
\bjtitle{Advances in Neural Information Processing Systems}
\bvolume{34},
\bfpage{19352}--\blpage{19362}
(\byear{2021})
\end{barticle}
\endbibitem

\bibitem[\protect\citeauthoryear{Van~Horn et~al.}{2021}]{van2021benchmarking}
\begin{bchapter}
\bauthor{\bsnm{Van~Horn}, \binits{G.}},
\bauthor{\bsnm{Cole}, \binits{E.}},
\bauthor{\bsnm{Beery}, \binits{S.}},
\bauthor{\bsnm{Wilber}, \binits{K.}},
\bauthor{\bsnm{Belongie}, \binits{S.}},
\bauthor{\bsnm{Mac~Aodha}, \binits{O.}}:
\bctitle{Benchmarking representation learning for natural world image collections}.
In: \bbtitle{Proceedings of the IEEE/CVF Conference on Computer Vision and Pattern Recognition},
pp. \bfpage{12884}--\blpage{12893}
(\byear{2021})
\end{bchapter}
\endbibitem

\bibitem[\protect\citeauthoryear{Nguyen et~al.}{2024}]{nguyen2024insect}
\begin{bchapter}
\bauthor{\bsnm{Nguyen}, \binits{H.-Q.}},
\bauthor{\bsnm{Truong}, \binits{T.-D.}},
\bauthor{\bsnm{Nguyen}, \binits{X.B.}},
\bauthor{\bsnm{Dowling}, \binits{A.}},
\bauthor{\bsnm{Li}, \binits{X.}},
\bauthor{\bsnm{Luu}, \binits{K.}}:
\bctitle{Insect-foundation: A foundation model and large-scale 1m dataset for visual insect understanding}.
In: \bbtitle{Proceedings of the IEEE/CVF Conference on Computer Vision and Pattern Recognition},
pp. \bfpage{21945}--\blpage{21955}
(\byear{2024})
\end{bchapter}
\endbibitem

\bibitem[\protect\citeauthoryear{He et~al.}{2020}]{he2020momentum}
\begin{bchapter}
\bauthor{\bsnm{He}, \binits{K.}},
\bauthor{\bsnm{Fan}, \binits{H.}},
\bauthor{\bsnm{Wu}, \binits{Y.}},
\bauthor{\bsnm{Xie}, \binits{S.}},
\bauthor{\bsnm{Girshick}, \binits{R.}}:
\bctitle{Momentum contrast for unsupervised visual representation learning}.
In: \bbtitle{Proceedings of the IEEE/CVF Conference on Computer Vision and Pattern Recognition},
pp. \bfpage{9729}--\blpage{9738}
(\byear{2020})
\end{bchapter}
\endbibitem

\bibitem[\protect\citeauthoryear{Chen et~al.}{2020a}]{chen2020improved}
\begin{botherref}
\oauthor{\bsnm{Chen}, \binits{X.}},
\oauthor{\bsnm{Fan}, \binits{H.}},
\oauthor{\bsnm{Girshick}, \binits{R.}},
\oauthor{\bsnm{He}, \binits{K.}}:
Improved baselines with momentum contrastive learning.
arXiv preprint arXiv:2003.04297
(2020)
\end{botherref}
\endbibitem

\bibitem[\protect\citeauthoryear{Chen et~al.}{2020b}]{chen2020simple}
\begin{bchapter}
\bauthor{\bsnm{Chen}, \binits{T.}},
\bauthor{\bsnm{Kornblith}, \binits{S.}},
\bauthor{\bsnm{Norouzi}, \binits{M.}},
\bauthor{\bsnm{Hinton}, \binits{G.}}:
\bctitle{A simple framework for contrastive learning of visual representations}.
In: \bbtitle{International Conference on Machine Learning},
pp. \bfpage{1597}--\blpage{1607}
(\byear{2020}).
\bcomment{PMLR}
\end{bchapter}
\endbibitem

\bibitem[\protect\citeauthoryear{Xie et~al.}{2022}]{xie2022simmim}
\begin{bchapter}
\bauthor{\bsnm{Xie}, \binits{Z.}},
\bauthor{\bsnm{Zhang}, \binits{Z.}},
\bauthor{\bsnm{Cao}, \binits{Y.}},
\bauthor{\bsnm{Lin}, \binits{Y.}},
\bauthor{\bsnm{Bao}, \binits{J.}},
\bauthor{\bsnm{Yao}, \binits{Z.}},
\bauthor{\bsnm{Dai}, \binits{Q.}},
\bauthor{\bsnm{Hu}, \binits{H.}}:
\bctitle{Simmim: A simple framework for masked image modeling}.
In: \bbtitle{Proceedings of the IEEE/CVF Conference on Computer Vision and Pattern Recognition},
pp. \bfpage{9653}--\blpage{9663}
(\byear{2022})
\end{bchapter}
\endbibitem

\bibitem[\protect\citeauthoryear{He et~al.}{2022}]{he2022masked}
\begin{bchapter}
\bauthor{\bsnm{He}, \binits{K.}},
\bauthor{\bsnm{Chen}, \binits{X.}},
\bauthor{\bsnm{Xie}, \binits{S.}},
\bauthor{\bsnm{Li}, \binits{Y.}},
\bauthor{\bsnm{Doll{\'a}r}, \binits{P.}},
\bauthor{\bsnm{Girshick}, \binits{R.}}:
\bctitle{Masked autoencoders are scalable vision learners}.
In: \bbtitle{Proceedings of the IEEE/CVF Conference on Computer Vision and Pattern Recognition},
pp. \bfpage{16000}--\blpage{16009}
(\byear{2022})
\end{bchapter}
\endbibitem

\bibitem[\protect\citeauthoryear{Caron et~al.}{2021}]{caron2021emerging}
\begin{bchapter}
\bauthor{\bsnm{Caron}, \binits{M.}},
\bauthor{\bsnm{Touvron}, \binits{H.}},
\bauthor{\bsnm{Misra}, \binits{I.}},
\bauthor{\bsnm{J{\'e}gou}, \binits{H.}},
\bauthor{\bsnm{Mairal}, \binits{J.}},
\bauthor{\bsnm{Bojanowski}, \binits{P.}},
\bauthor{\bsnm{Joulin}, \binits{A.}}:
\bctitle{Emerging properties in self-supervised vision transformers}.
In: \bbtitle{Proceedings of the IEEE/CVF International Conference on Computer Vision},
pp. \bfpage{9650}--\blpage{9660}
(\byear{2021})
\end{bchapter}
\endbibitem

\bibitem[\protect\citeauthoryear{Oquab et~al.}{2023}]{oquab2023dinov2}
\begin{botherref}
\oauthor{\bsnm{Oquab}, \binits{M.}},
\oauthor{\bsnm{Darcet}, \binits{T.}},
\oauthor{\bsnm{Moutakanni}, \binits{T.}},
\oauthor{\bsnm{Vo}, \binits{H.}},
\oauthor{\bsnm{Szafraniec}, \binits{M.}},
\oauthor{\bsnm{Khalidov}, \binits{V.}},
\oauthor{\bsnm{Fernandez}, \binits{P.}},
\oauthor{\bsnm{Haziza}, \binits{D.}},
\oauthor{\bsnm{Massa}, \binits{F.}},
\oauthor{\bsnm{El-Nouby}, \binits{A.}}, et al.:
Dinov2: Learning robust visual features without supervision.
arXiv preprint arXiv:2304.07193
(2023)
\end{botherref}
\endbibitem

\bibitem[\protect\citeauthoryear{Jia et~al.}{2021}]{jia2021scaling}
\begin{bchapter}
\bauthor{\bsnm{Jia}, \binits{C.}},
\bauthor{\bsnm{Yang}, \binits{Y.}},
\bauthor{\bsnm{Xia}, \binits{Y.}},
\bauthor{\bsnm{Chen}, \binits{Y.-T.}},
\bauthor{\bsnm{Parekh}, \binits{Z.}},
\bauthor{\bsnm{Pham}, \binits{H.}},
\bauthor{\bsnm{Le}, \binits{Q.}},
\bauthor{\bsnm{Sung}, \binits{Y.-H.}},
\bauthor{\bsnm{Li}, \binits{Z.}},
\bauthor{\bsnm{Duerig}, \binits{T.}}:
\bctitle{Scaling up visual and vision-language representation learning with noisy text supervision}.
In: \bbtitle{International Conference on Machine Learning},
pp. \bfpage{4904}--\blpage{4916}
(\byear{2021}).
\bcomment{PMLR}
\end{bchapter}
\endbibitem

\bibitem[\protect\citeauthoryear{Yu et~al.}{2022}]{yu2022coca}
\begin{botherref}
\oauthor{\bsnm{Yu}, \binits{J.}},
\oauthor{\bsnm{Wang}, \binits{Z.}},
\oauthor{\bsnm{Vasudevan}, \binits{V.}},
\oauthor{\bsnm{Yeung}, \binits{L.}},
\oauthor{\bsnm{Seyedhosseini}, \binits{M.}},
\oauthor{\bsnm{Wu}, \binits{Y.}}:
Coca: Contrastive captioners are image-text foundation models.
arXiv preprint arXiv:2205.01917
(2022)
\end{botherref}
\endbibitem

\bibitem[\protect\citeauthoryear{Deng et~al.}{2009}]{deng2009imagenet}
\begin{bchapter}
\bauthor{\bsnm{Deng}, \binits{J.}},
\bauthor{\bsnm{Dong}, \binits{W.}},
\bauthor{\bsnm{Socher}, \binits{R.}},
\bauthor{\bsnm{Li}, \binits{L.-J.}},
\bauthor{\bsnm{Li}, \binits{K.}},
\bauthor{\bsnm{Fei-Fei}, \binits{L.}}:
\bctitle{Imagenet: A large-scale hierarchical image database}.
In: \bbtitle{2009 IEEE Conference on Computer Vision and Pattern Recognition},
pp. \bfpage{248}--\blpage{255}
(\byear{2009}).
\bcomment{Ieee}
\end{bchapter}
\endbibitem

\bibitem[\protect\citeauthoryear{Zhai et~al.}{2022}]{zhai2022scaling}
\begin{bchapter}
\bauthor{\bsnm{Zhai}, \binits{X.}},
\bauthor{\bsnm{Kolesnikov}, \binits{A.}},
\bauthor{\bsnm{Houlsby}, \binits{N.}},
\bauthor{\bsnm{Beyer}, \binits{L.}}:
\bctitle{Scaling vision transformers}.
In: \bbtitle{Proceedings of the IEEE/CVF Conference on Computer Vision and Pattern Recognition},
pp. \bfpage{12104}--\blpage{12113}
(\byear{2022})
\end{bchapter}
\endbibitem

\bibitem[\protect\citeauthoryear{Schuhmann et~al.}{2022}]{schuhmann2022laion}
\begin{barticle}
\bauthor{\bsnm{Schuhmann}, \binits{C.}},
\bauthor{\bsnm{Beaumont}, \binits{R.}},
\bauthor{\bsnm{Vencu}, \binits{R.}},
\bauthor{\bsnm{Gordon}, \binits{C.}},
\bauthor{\bsnm{Wightman}, \binits{R.}},
\bauthor{\bsnm{Cherti}, \binits{M.}},
\bauthor{\bsnm{Coombes}, \binits{T.}},
\bauthor{\bsnm{Katta}, \binits{A.}},
\bauthor{\bsnm{Mullis}, \binits{C.}},
\bauthor{\bsnm{Wortsman}, \binits{M.}}, \betal:
\batitle{Laion-5b: An open large-scale dataset for training next generation image-text models}.
\bjtitle{Advances in Neural Information Processing Systems}
\bvolume{35},
\bfpage{25278}--\blpage{25294}
(\byear{2022})
\end{barticle}
\endbibitem

\bibitem[\protect\citeauthoryear{Alves et~al.}{2020}]{alves2020cotton}
\begin{barticle}
\bauthor{\bsnm{Alves}, \binits{A.N.}},
\bauthor{\bsnm{Souza}, \binits{W.S.}},
\bauthor{\bsnm{Borges}, \binits{D.L.}}:
\batitle{Cotton pests classification in field-based images using deep residual networks}.
\bjtitle{Computers and Electronics in Agriculture}
\bvolume{174},
\bfpage{105488}
(\byear{2020})
\end{barticle}
\endbibitem

\bibitem[\protect\citeauthoryear{Bollis et~al.}{2020}]{bollis2020weakly}
\begin{bchapter}
\bauthor{\bsnm{Bollis}, \binits{E.}},
\bauthor{\bsnm{Pedrini}, \binits{H.}},
\bauthor{\bsnm{Avila}, \binits{S.}}:
\bctitle{Weakly supervised learning guided by activation mapping applied to a novel citrus pest benchmark}.
In: \bbtitle{Proceedings of the IEEE/CVF Conference on Computer Vision and Pattern Recognition Workshops},
pp. \bfpage{70}--\blpage{71}
(\byear{2020})
\end{bchapter}
\endbibitem

\bibitem[\protect\citeauthoryear{Stork}{2018}]{stork2018many}
\begin{barticle}
\bauthor{\bsnm{Stork}, \binits{N.E.}}:
\batitle{How many species of insects and other terrestrial arthropods are there on earth?}
\bjtitle{Annual review of entomology}
\bvolume{63},
\bfpage{31}--\blpage{45}
(\byear{2018})
\end{barticle}
\endbibitem

\bibitem[\protect\citeauthoryear{Ratnasingham and Hebert}{2007}]{ratnasingham2007bold}
\begin{barticle}
\bauthor{\bsnm{Ratnasingham}, \binits{S.}},
\bauthor{\bsnm{Hebert}, \binits{P.D.}}:
\batitle{Bold: The barcode of life data system (http://www. barcodinglife. org)}.
\bjtitle{Molecular ecology notes}
\bvolume{7}(\bissue{3}),
\bfpage{355}--\blpage{364}
(\byear{2007})
\end{barticle}
\endbibitem

\bibitem[\protect\citeauthoryear{Chen et~al.}{2021}]{chen2021empirical}
\begin{bchapter}
\bauthor{\bsnm{Chen}, \binits{X.}},
\bauthor{\bsnm{Xie}, \binits{S.}},
\bauthor{\bsnm{He}, \binits{K.}}:
\bctitle{An empirical study of training self-supervised vision transformers. in 2021 ieee}.
In: \bbtitle{CVF International Conference on Computer Vision (ICCV)},
pp. \bfpage{9620}--\blpage{9629}
(\byear{2021})
\end{bchapter}
\endbibitem

\bibitem[\protect\citeauthoryear{Chen et~al.}{2023}]{chen2023jigsaw}
\begin{barticle}
\bauthor{\bsnm{Chen}, \binits{Y.}},
\bauthor{\bsnm{Shen}, \binits{X.}},
\bauthor{\bsnm{Liu}, \binits{Y.}},
\bauthor{\bsnm{Tao}, \binits{Q.}},
\bauthor{\bsnm{Suykens}, \binits{J.A.}}:
\batitle{Jigsaw-vit: Learning jigsaw puzzles in vision transformer}.
\bjtitle{Pattern Recognition Letters}
\bvolume{166},
\bfpage{53}--\blpage{60}
(\byear{2023})
\end{barticle}
\endbibitem

\bibitem[\protect\citeauthoryear{Truong et~al.}{2022}]{truong2022direcformer}
\begin{bchapter}
\bauthor{\bsnm{Truong}, \binits{T.-D.}},
\bauthor{\bsnm{Bui}, \binits{Q.-H.}},
\bauthor{\bsnm{Duong}, \binits{C.N.}},
\bauthor{\bsnm{Seo}, \binits{H.-S.}},
\bauthor{\bsnm{Phung}, \binits{S.L.}},
\bauthor{\bsnm{Li}, \binits{X.}},
\bauthor{\bsnm{Luu}, \binits{K.}}:
\bctitle{Direcformer: A directed attention in transformer approach to robust action recognition}.
In: \bbtitle{Proceedings of the IEEE/CVF Conference on Computer Vision and Pattern Recognition},
pp. \bfpage{20030}--\blpage{20040}
(\byear{2022})
\end{bchapter}
\endbibitem

\bibitem[\protect\citeauthoryear{Nguyen et~al.}{2023}]{nguyen2023micron}
\begin{bchapter}
\bauthor{\bsnm{Nguyen}, \binits{X.-B.}},
\bauthor{\bsnm{Duong}, \binits{C.N.}},
\bauthor{\bsnm{Li}, \binits{X.}},
\bauthor{\bsnm{Gauch}, \binits{S.}},
\bauthor{\bsnm{Seo}, \binits{H.-S.}},
\bauthor{\bsnm{Luu}, \binits{K.}}:
\bctitle{Micron-bert: Bert-based facial micro-expression recognition}.
In: \bbtitle{Proceedings of the IEEE/CVF Conference on Computer Vision and Pattern Recognition},
pp. \bfpage{1482}--\blpage{1492}
(\byear{2023})
\end{bchapter}
\endbibitem

\bibitem[\protect\citeauthoryear{Nguyen et~al.}{2021}]{nguyen2021clusformer}
\begin{bchapter}
\bauthor{\bsnm{Nguyen}, \binits{X.-B.}},
\bauthor{\bsnm{Bui}, \binits{D.T.}},
\bauthor{\bsnm{Duong}, \binits{C.N.}},
\bauthor{\bsnm{Bui}, \binits{T.D.}},
\bauthor{\bsnm{Luu}, \binits{K.}}:
\bctitle{Clusformer: A transformer based clustering approach to unsupervised large-scale face and visual landmark recognition}.
In: \bbtitle{Proceedings of the IEEE/CVF Conference on Computer Vision and Pattern Recognition},
pp. \bfpage{10847}--\blpage{10856}
(\byear{2021})
\end{bchapter}
\endbibitem

\bibitem[\protect\citeauthoryear{Truong et~al.}{2021}]{truong2021bimal}
\begin{bchapter}
\bauthor{\bsnm{Truong}, \binits{T.-D.}},
\bauthor{\bsnm{Duong}, \binits{C.N.}},
\bauthor{\bsnm{Le}, \binits{N.}},
\bauthor{\bsnm{Phung}, \binits{S.L.}},
\bauthor{\bsnm{Rainwater}, \binits{C.}},
\bauthor{\bsnm{Luu}, \binits{K.}}:
\bctitle{Bimal: Bijective maximum likelihood approach to domain adaptation in semantic scene segmentation}.
In: \bbtitle{Proceedings of the IEEE/CVF International Conference on Computer Vision},
pp. \bfpage{8548}--\blpage{8557}
(\byear{2021})
\end{bchapter}
\endbibitem

\bibitem[\protect\citeauthoryear{Truong et~al.}{2024}]{truong2024fairness}
\begin{botherref}
\oauthor{\bsnm{Truong}, \binits{T.-D.}},
\oauthor{\bsnm{Nguyen}, \binits{H.-Q.}},
\oauthor{\bsnm{Raj}, \binits{B.}},
\oauthor{\bsnm{Luu}, \binits{K.}}:
Fairness continual learning approach to semantic scene understanding in open-world environments.
Advances in Neural Information Processing Systems
\textbf{36}
(2024)
\end{botherref}
\endbibitem

\bibitem[\protect\citeauthoryear{Truong et~al.}{2023a}]{truong2023fredom}
\begin{bchapter}
\bauthor{\bsnm{Truong}, \binits{T.-D.}},
\bauthor{\bsnm{Le}, \binits{N.}},
\bauthor{\bsnm{Raj}, \binits{B.}},
\bauthor{\bsnm{Cothren}, \binits{J.}},
\bauthor{\bsnm{Luu}, \binits{K.}}:
\bctitle{Fredom: Fairness domain adaptation approach to semantic scene understanding}.
In: \bbtitle{IEEE/CVF Computer Vision and Pattern Recognition (CVPR)}
(\byear{2023})
\end{bchapter}
\endbibitem

\bibitem[\protect\citeauthoryear{Truong et~al.}{2023b}]{truong2023liaad}
\begin{barticle}
\bauthor{\bsnm{Truong}, \binits{T.-D.}},
\bauthor{\bsnm{Duong}, \binits{C.N.}},
\bauthor{\bsnm{Quach}, \binits{K.G.}},
\bauthor{\bsnm{Le}, \binits{N.}},
\bauthor{\bsnm{Bui}, \binits{T.D.}},
\bauthor{\bsnm{Luu}, \binits{K.}}:
\batitle{Liaad: Lightweight attentive angular distillation for large-scale age-invariant face recognition}.
\bjtitle{Neurocomputing}
\bvolume{543},
\bfpage{126198}
(\byear{2023})
\end{barticle}
\endbibitem

\bibitem[\protect\citeauthoryear{Nguyen et~al.}{2023a}]{nguyen2023brainformer}
\begin{botherref}
\oauthor{\bsnm{Nguyen}, \binits{X.-B.}},
\oauthor{\bsnm{Li}, \binits{X.}},
\oauthor{\bsnm{Khan}, \binits{S.U.}},
\oauthor{\bsnm{Luu}, \binits{K.}}:
Brainformer: Modeling mri brain functions to machine vision.
arXiv preprint arXiv:2312.00236
(2023)
\end{botherref}
\endbibitem

\bibitem[\protect\citeauthoryear{Nguyen et~al.}{2023b}]{nguyen2023fairness}
\begin{botherref}
\oauthor{\bsnm{Nguyen}, \binits{X.-B.}},
\oauthor{\bsnm{Duong}, \binits{C.N.}},
\oauthor{\bsnm{Savvides}, \binits{M.}},
\oauthor{\bsnm{Roy}, \binits{K.}},
\oauthor{\bsnm{Luu}, \binits{K.}}:
Fairness in visual clustering: A novel transformer clustering approach.
arXiv preprint arXiv:2304.07408
(2023)
\end{botherref}
\endbibitem

\bibitem[\protect\citeauthoryear{Nguyen et~al.}{2020}]{sefl_supervised_medical}
\begin{barticle}
\bauthor{\bsnm{Nguyen}, \binits{X.-B.}},
\bauthor{\bsnm{Lee}, \binits{G.S.}},
\bauthor{\bsnm{Kim}, \binits{S.H.}},
\bauthor{\bsnm{Yang}, \binits{H.J.}}:
\batitle{Self-supervised learning based on spatial awareness for medical image analysis}.
\bjtitle{IEEE Access}
\bvolume{8},
\bfpage{162973}--\blpage{162981}
(\byear{2020})
\doiurl{10.1109/ACCESS.2020.3021469}
\end{barticle}
\endbibitem

\bibitem[\protect\citeauthoryear{Nguyen et~al.}{2022}]{nguyen2022two}
\begin{botherref}
\oauthor{\bsnm{Nguyen}, \binits{X.B.}},
\oauthor{\bsnm{Bisht}, \binits{A.}},
\oauthor{\bsnm{Churchill}, \binits{H.}},
\oauthor{\bsnm{Luu}, \binits{K.}}:
Two-dimensional quantum material identification via self-attention and soft-labeling in deep learning.
arXiv preprint arXiv:2205.15948
(2022)
\end{botherref}
\endbibitem

\bibitem[\protect\citeauthoryear{Bao et~al.}{2021}]{bao2021beit}
\begin{botherref}
\oauthor{\bsnm{Bao}, \binits{H.}},
\oauthor{\bsnm{Dong}, \binits{L.}},
\oauthor{\bsnm{Piao}, \binits{S.}},
\oauthor{\bsnm{Wei}, \binits{F.}}:
Beit: Bert pre-training of image transformers.
arXiv preprint arXiv:2106.08254
(2021)
\end{botherref}
\endbibitem

\bibitem[\protect\citeauthoryear{Zhai et~al.}{2022}]{zhai2022lit}
\begin{bchapter}
\bauthor{\bsnm{Zhai}, \binits{X.}},
\bauthor{\bsnm{Wang}, \binits{X.}},
\bauthor{\bsnm{Mustafa}, \binits{B.}},
\bauthor{\bsnm{Steiner}, \binits{A.}},
\bauthor{\bsnm{Keysers}, \binits{D.}},
\bauthor{\bsnm{Kolesnikov}, \binits{A.}},
\bauthor{\bsnm{Beyer}, \binits{L.}}:
\bctitle{Lit: Zero-shot transfer with locked-image text tuning}.
In: \bbtitle{Proceedings of the IEEE/CVF Conference on Computer Vision and Pattern Recognition},
pp. \bfpage{18123}--\blpage{18133}
(\byear{2022})
\end{bchapter}
\endbibitem

\bibitem[\protect\citeauthoryear{Pham et~al.}{2023}]{pham2023combined}
\begin{barticle}
\bauthor{\bsnm{Pham}, \binits{H.}},
\bauthor{\bsnm{Dai}, \binits{Z.}},
\bauthor{\bsnm{Ghiasi}, \binits{G.}},
\bauthor{\bsnm{Kawaguchi}, \binits{K.}},
\bauthor{\bsnm{Liu}, \binits{H.}},
\bauthor{\bsnm{Yu}, \binits{A.W.}},
\bauthor{\bsnm{Yu}, \binits{J.}},
\bauthor{\bsnm{Chen}, \binits{Y.-T.}},
\bauthor{\bsnm{Luong}, \binits{M.-T.}},
\bauthor{\bsnm{Wu}, \binits{Y.}}, \betal:
\batitle{Combined scaling for zero-shot transfer learning}.
\bjtitle{Neurocomputing}
\bvolume{555},
\bfpage{126658}
(\byear{2023})
\end{barticle}
\endbibitem

\bibitem[\protect\citeauthoryear{Wang et~al.}{2021}]{wang2021simvlm}
\begin{botherref}
\oauthor{\bsnm{Wang}, \binits{Z.}},
\oauthor{\bsnm{Yu}, \binits{J.}},
\oauthor{\bsnm{Yu}, \binits{A.W.}},
\oauthor{\bsnm{Dai}, \binits{Z.}},
\oauthor{\bsnm{Tsvetkov}, \binits{Y.}},
\oauthor{\bsnm{Cao}, \binits{Y.}}:
Simvlm: Simple visual language model pretraining with weak supervision.
arXiv preprint arXiv:2108.10904
(2021)
\end{botherref}
\endbibitem

\bibitem[\protect\citeauthoryear{Wang et~al.}{2022}]{wang2022ofa}
\begin{bchapter}
\bauthor{\bsnm{Wang}, \binits{P.}},
\bauthor{\bsnm{Yang}, \binits{A.}},
\bauthor{\bsnm{Men}, \binits{R.}},
\bauthor{\bsnm{Lin}, \binits{J.}},
\bauthor{\bsnm{Bai}, \binits{S.}},
\bauthor{\bsnm{Li}, \binits{Z.}},
\bauthor{\bsnm{Ma}, \binits{J.}},
\bauthor{\bsnm{Zhou}, \binits{C.}},
\bauthor{\bsnm{Zhou}, \binits{J.}},
\bauthor{\bsnm{Yang}, \binits{H.}}:
\bctitle{Ofa: Unifying architectures, tasks, and modalities through a simple sequence-to-sequence learning framework}.
In: \bbtitle{International Conference on Machine Learning},
pp. \bfpage{23318}--\blpage{23340}
(\byear{2022}).
\bcomment{PMLR}
\end{bchapter}
\endbibitem

\bibitem[\protect\citeauthoryear{Li et~al.}{2022}]{li2022blip}
\begin{bchapter}
\bauthor{\bsnm{Li}, \binits{J.}},
\bauthor{\bsnm{Li}, \binits{D.}},
\bauthor{\bsnm{Xiong}, \binits{C.}},
\bauthor{\bsnm{Hoi}, \binits{S.}}:
\bctitle{Blip: Bootstrapping language-image pre-training for unified vision-language understanding and generation}.
In: \bbtitle{International Conference on Machine Learning},
pp. \bfpage{12888}--\blpage{12900}
(\byear{2022}).
\bcomment{PMLR}
\end{bchapter}
\endbibitem

\bibitem[\protect\citeauthoryear{Zhai et~al.}{2023}]{zhai2023sigmoid}
\begin{botherref}
\oauthor{\bsnm{Zhai}, \binits{X.}},
\oauthor{\bsnm{Mustafa}, \binits{B.}},
\oauthor{\bsnm{Kolesnikov}, \binits{A.}},
\oauthor{\bsnm{Beyer}, \binits{L.}}:
Sigmoid loss for language image pre-training.
arXiv preprint arXiv:2303.15343
(2023)
\end{botherref}
\endbibitem

\bibitem[\protect\citeauthoryear{Luo et~al.}{2023}]{luo2023lexlip}
\begin{bchapter}
\bauthor{\bsnm{Luo}, \binits{Z.}},
\bauthor{\bsnm{Zhao}, \binits{P.}},
\bauthor{\bsnm{Xu}, \binits{C.}},
\bauthor{\bsnm{Geng}, \binits{X.}},
\bauthor{\bsnm{Shen}, \binits{T.}},
\bauthor{\bsnm{Tao}, \binits{C.}},
\bauthor{\bsnm{Ma}, \binits{J.}},
\bauthor{\bsnm{Lin}, \binits{Q.}},
\bauthor{\bsnm{Jiang}, \binits{D.}}:
\bctitle{Lexlip: Lexicon-bottlenecked language-image pre-training for large-scale image-text sparse retrieval}.
In: \bbtitle{Proceedings of the IEEE/CVF International Conference on Computer Vision},
pp. \bfpage{11206}--\blpage{11217}
(\byear{2023})
\end{bchapter}
\endbibitem

\bibitem[\protect\citeauthoryear{Wang et~al.}{2023}]{wang2023equivariant}
\begin{botherref}
\oauthor{\bsnm{Wang}, \binits{T.}},
\oauthor{\bsnm{Lin}, \binits{K.}},
\oauthor{\bsnm{Li}, \binits{L.}},
\oauthor{\bsnm{Lin}, \binits{C.-C.}},
\oauthor{\bsnm{Yang}, \binits{Z.}},
\oauthor{\bsnm{Zhang}, \binits{H.}},
\oauthor{\bsnm{Liu}, \binits{Z.}},
\oauthor{\bsnm{Wang}, \binits{L.}}:
Equivariant similarity for vision-language foundation models.
arXiv preprint arXiv:2303.14465
(2023)
\end{botherref}
\endbibitem

\bibitem[\protect\citeauthoryear{Weng et~al.}{2024}]{weng2024longvlm}
\begin{botherref}
\oauthor{\bsnm{Weng}, \binits{Y.}},
\oauthor{\bsnm{Han}, \binits{M.}},
\oauthor{\bsnm{He}, \binits{H.}},
\oauthor{\bsnm{Chang}, \binits{X.}},
\oauthor{\bsnm{Zhuang}, \binits{B.}}:
Longvlm: Efficient long video understanding via large language models.
arXiv preprint arXiv:2404.03384
(2024)
\end{botherref}
\endbibitem

\bibitem[\protect\citeauthoryear{Zhao et~al.}{2023}]{zhao2023learning}
\begin{bchapter}
\bauthor{\bsnm{Zhao}, \binits{Y.}},
\bauthor{\bsnm{Misra}, \binits{I.}},
\bauthor{\bsnm{Kr{\"a}henb{\"u}hl}, \binits{P.}},
\bauthor{\bsnm{Girdhar}, \binits{R.}}:
\bctitle{Learning video representations from large language models}.
In: \bbtitle{Proceedings of the IEEE/CVF Conference on Computer Vision and Pattern Recognition},
pp. \bfpage{6586}--\blpage{6597}
(\byear{2023})
\end{bchapter}
\endbibitem

\bibitem[\protect\citeauthoryear{Hussain et~al.}{2023}]{hussain2023m}
\begin{botherref}
\oauthor{\bsnm{Hussain}, \binits{A.S.}},
\oauthor{\bsnm{Liu}, \binits{S.}},
\oauthor{\bsnm{Sun}, \binits{C.}},
\oauthor{\bsnm{Shan}, \binits{Y.}}:
M2ugen: Multi-modal music understanding and generation with the power of large language models.
arXiv preprint arXiv:2311.11255
(2023)
\end{botherref}
\endbibitem

\bibitem[\protect\citeauthoryear{Ghosal et~al.}{2023}]{ghosal2023text}
\begin{botherref}
\oauthor{\bsnm{Ghosal}, \binits{D.}},
\oauthor{\bsnm{Majumder}, \binits{N.}},
\oauthor{\bsnm{Mehrish}, \binits{A.}},
\oauthor{\bsnm{Poria}, \binits{S.}}:
Text-to-audio generation using instruction-tuned llm and latent diffusion model.
arXiv preprint arXiv:2304.13731
(2023)
\end{botherref}
\endbibitem

\bibitem[\protect\citeauthoryear{Chen and Zhang}{2024}]{chenfedmbridge}
\begin{bchapter}
\bauthor{\bsnm{Chen}, \binits{J.}},
\bauthor{\bsnm{Zhang}, \binits{A.}}:
\bctitle{Fed{MB}ridge: Bridgeable multimodal federated learning}.
In: \bbtitle{Forty-first International Conference on Machine Learning}
(\byear{2024}).
\burl{https://openreview.net/forum?id=jrHUbftLd6}
\end{bchapter}
\endbibitem

\bibitem[\protect\citeauthoryear{Ren et~al.}{2024}]{ren2024pixellm}
\begin{bchapter}
\bauthor{\bsnm{Ren}, \binits{Z.}},
\bauthor{\bsnm{Huang}, \binits{Z.}},
\bauthor{\bsnm{Wei}, \binits{Y.}},
\bauthor{\bsnm{Zhao}, \binits{Y.}},
\bauthor{\bsnm{Fu}, \binits{D.}},
\bauthor{\bsnm{Feng}, \binits{J.}},
\bauthor{\bsnm{Jin}, \binits{X.}}:
\bctitle{Pixellm: Pixel reasoning with large multimodal model}.
In: \bbtitle{Proceedings of the IEEE/CVF Conference on Computer Vision and Pattern Recognition},
pp. \bfpage{26374}--\blpage{26383}
(\byear{2024})
\end{bchapter}
\endbibitem

\bibitem[\protect\citeauthoryear{Yuan et~al.}{2024}]{yuan2024osprey}
\begin{bchapter}
\bauthor{\bsnm{Yuan}, \binits{Y.}},
\bauthor{\bsnm{Li}, \binits{W.}},
\bauthor{\bsnm{Liu}, \binits{J.}},
\bauthor{\bsnm{Tang}, \binits{D.}},
\bauthor{\bsnm{Luo}, \binits{X.}},
\bauthor{\bsnm{Qin}, \binits{C.}},
\bauthor{\bsnm{Zhang}, \binits{L.}},
\bauthor{\bsnm{Zhu}, \binits{J.}}:
\bctitle{Osprey: Pixel understanding with visual instruction tuning}.
In: \bbtitle{Proceedings of the IEEE/CVF Conference on Computer Vision and Pattern Recognition},
pp. \bfpage{28202}--\blpage{28211}
(\byear{2024})
\end{bchapter}
\endbibitem

\bibitem[\protect\citeauthoryear{Zhang et~al.}{2024}]{zhang2024omgllava}
\begin{bchapter}
\bauthor{\bsnm{Zhang}, \binits{T.}},
\bauthor{\bsnm{Li}, \binits{X.}},
\bauthor{\bsnm{Fei}, \binits{H.}},
\bauthor{\bsnm{Yuan}, \binits{H.}},
\bauthor{\bsnm{Wu}, \binits{S.}},
\bauthor{\bsnm{Ji}, \binits{S.}},
\bauthor{\bsnm{Loy}, \binits{C.C.}},
\bauthor{\bsnm{YAN}, \binits{S.}}:
\bctitle{{OMG}-{LL}a{VA}: Bridging image-level, object-level, pixel-level reasoning and understanding}.
In: \bbtitle{The Thirty-eighth Annual Conference on Neural Information Processing Systems}
(\byear{2024}).
\burl{https://openreview.net/forum?id=WeoNd6PRqS}
\end{bchapter}
\endbibitem

\bibitem[\protect\citeauthoryear{Fei et~al.}{2024}]{fei2024vitron}
\begin{botherref}
\oauthor{\bsnm{Fei}, \binits{H.}},
\oauthor{\bsnm{Wu}, \binits{S.}},
\oauthor{\bsnm{Zhang}, \binits{H.}},
\oauthor{\bsnm{Chua}, \binits{T.-S.}},
\oauthor{\bsnm{Yan}, \binits{S.}}:
VITRON: A Unified Pixel-level Vision LLM for Understanding, Generating, Segmenting, Editing.
CoRR
(2024)
\end{botherref}
\endbibitem

\bibitem[\protect\citeauthoryear{Wu et~al.}{2024}]{wu24next}
\begin{bchapter}
\bauthor{\bsnm{Wu}, \binits{S.}},
\bauthor{\bsnm{Fei}, \binits{H.}},
\bauthor{\bsnm{Qu}, \binits{L.}},
\bauthor{\bsnm{Ji}, \binits{W.}},
\bauthor{\bsnm{Chua}, \binits{T.-S.}}:
\bctitle{{NE}x{T}-{GPT}: Any-to-any multimodal {LLM}}.
In: \bbtitle{Proceedings of the International Conference on Machine Learning},
pp. \bfpage{53366}--\blpage{53397}
(\byear{2024})
\end{bchapter}
\endbibitem

\bibitem[\protect\citeauthoryear{Yue et~al.}{2024}]{yue2024mmmu}
\begin{bchapter}
\bauthor{\bsnm{Yue}, \binits{X.}},
\bauthor{\bsnm{Ni}, \binits{Y.}},
\bauthor{\bsnm{Zhang}, \binits{K.}},
\bauthor{\bsnm{Zheng}, \binits{T.}},
\bauthor{\bsnm{Liu}, \binits{R.}},
\bauthor{\bsnm{Zhang}, \binits{G.}},
\bauthor{\bsnm{Stevens}, \binits{S.}},
\bauthor{\bsnm{Jiang}, \binits{D.}},
\bauthor{\bsnm{Ren}, \binits{W.}},
\bauthor{\bsnm{Sun}, \binits{Y.}},
\bauthor{\bsnm{Wei}, \binits{C.}},
\bauthor{\bsnm{Yu}, \binits{B.}},
\bauthor{\bsnm{Yuan}, \binits{R.}},
\bauthor{\bsnm{Sun}, \binits{R.}},
\bauthor{\bsnm{Yin}, \binits{M.}},
\bauthor{\bsnm{Zheng}, \binits{B.}},
\bauthor{\bsnm{Yang}, \binits{Z.}},
\bauthor{\bsnm{Liu}, \binits{Y.}},
\bauthor{\bsnm{Huang}, \binits{W.}},
\bauthor{\bsnm{Sun}, \binits{H.}},
\bauthor{\bsnm{Su}, \binits{Y.}},
\bauthor{\bsnm{Chen}, \binits{W.}}:
\bctitle{Mmmu: A massive multi-discipline multimodal understanding and reasoning benchmark for expert agi}.
In: \bbtitle{Proceedings of the IEEE/CVF Conference on Computer Vision and Pattern Recognition},
pp. \bfpage{9556}--\blpage{9567}
(\byear{2024})
\end{bchapter}
\endbibitem

\bibitem[\protect\citeauthoryear{Ye et~al.}{2024}]{ye2024mm}
\begin{botherref}
\oauthor{\bsnm{Ye}, \binits{W.}},
\oauthor{\bsnm{Zheng}, \binits{G.}},
\oauthor{\bsnm{Ma}, \binits{Y.}},
\oauthor{\bsnm{Cao}, \binits{X.}},
\oauthor{\bsnm{Lai}, \binits{B.}},
\oauthor{\bsnm{Rehg}, \binits{J.M.}},
\oauthor{\bsnm{Zhang}, \binits{A.}}:
Mm-spubench: Towards better understanding of spurious biases in multimodal llms.
arXiv preprint arXiv:2406.17126
(2024)
\end{botherref}
\endbibitem

\bibitem[\protect\citeauthoryear{Achiam et~al.}{2023}]{achiam2023gpt}
\begin{botherref}
\oauthor{\bsnm{Achiam}, \binits{J.}},
\oauthor{\bsnm{Adler}, \binits{S.}},
\oauthor{\bsnm{Agarwal}, \binits{S.}},
\oauthor{\bsnm{Ahmad}, \binits{L.}},
\oauthor{\bsnm{Akkaya}, \binits{I.}},
\oauthor{\bsnm{Aleman}, \binits{F.L.}},
\oauthor{\bsnm{Almeida}, \binits{D.}},
\oauthor{\bsnm{Altenschmidt}, \binits{J.}},
\oauthor{\bsnm{Altman}, \binits{S.}},
\oauthor{\bsnm{Anadkat}, \binits{S.}}, et al.:
Gpt-4 technical report.
arXiv preprint arXiv:2303.08774
(2023)
\end{botherref}
\endbibitem

\bibitem[\protect\citeauthoryear{Dubey et~al.}{2024}]{dubey2024llama}
\begin{botherref}
\oauthor{\bsnm{Dubey}, \binits{A.}},
\oauthor{\bsnm{Jauhri}, \binits{A.}},
\oauthor{\bsnm{Pandey}, \binits{A.}},
\oauthor{\bsnm{Kadian}, \binits{A.}},
\oauthor{\bsnm{Al-Dahle}, \binits{A.}},
\oauthor{\bsnm{Letman}, \binits{A.}},
\oauthor{\bsnm{Mathur}, \binits{A.}},
\oauthor{\bsnm{Schelten}, \binits{A.}},
\oauthor{\bsnm{Yang}, \binits{A.}},
\oauthor{\bsnm{Fan}, \binits{A.}}, et al.:
The llama 3 herd of models.
arXiv preprint arXiv:2407.21783
(2024)
\end{botherref}
\endbibitem

\bibitem[\protect\citeauthoryear{Chiang et~al.}{2023}]{vicuna2023}
\begin{botherref}
\oauthor{\bsnm{Chiang}, \binits{W.-L.}},
\oauthor{\bsnm{Li}, \binits{Z.}},
\oauthor{\bsnm{Lin}, \binits{Z.}},
\oauthor{\bsnm{Sheng}, \binits{Y.}},
\oauthor{\bsnm{Wu}, \binits{Z.}},
\oauthor{\bsnm{Zhang}, \binits{H.}},
\oauthor{\bsnm{Zheng}, \binits{L.}},
\oauthor{\bsnm{Zhuang}, \binits{S.}},
\oauthor{\bsnm{Zhuang}, \binits{Y.}},
\oauthor{\bsnm{Gonzalez}, \binits{J.E.}},
\oauthor{\bsnm{Stoica}, \binits{I.}},
\oauthor{\bsnm{Xing}, \binits{E.P.}}:
Vicuna: An Open-Source Chatbot Impressing GPT-4 with 90\%* ChatGPT Quality
(2023).
\url{https://lmsys.org/blog/2023-03-30-vicuna/}
\end{botherref}
\endbibitem

\bibitem[\protect\citeauthoryear{Noroozi and Favaro}{2016}]{noroozi2016unsupervised}
\begin{bchapter}
\bauthor{\bsnm{Noroozi}, \binits{M.}},
\bauthor{\bsnm{Favaro}, \binits{P.}}:
\bctitle{Unsupervised learning of visual representations by solving jigsaw puzzles}.
In: \bbtitle{European Conference on Computer Vision},
pp. \bfpage{69}--\blpage{84}
(\byear{2016}).
\bcomment{Springer}
\end{bchapter}
\endbibitem

\bibitem[\protect\citeauthoryear{Devlin et~al.}{2018}]{devlin2018bert}
\begin{botherref}
\oauthor{\bsnm{Devlin}, \binits{J.}},
\oauthor{\bsnm{Chang}, \binits{M.-W.}},
\oauthor{\bsnm{Lee}, \binits{K.}},
\oauthor{\bsnm{Toutanova}, \binits{K.}}:
Bert: Pre-training of deep bidirectional transformers for language understanding.
arXiv preprint arXiv:1810.04805
(2018)
\end{botherref}
\endbibitem

\bibitem[\protect\citeauthoryear{Liu et~al.}{2019}]{liu2019roberta}
\begin{botherref}
\oauthor{\bsnm{Liu}, \binits{Y.}},
\oauthor{\bsnm{Ott}, \binits{M.}},
\oauthor{\bsnm{Goyal}, \binits{N.}},
\oauthor{\bsnm{Du}, \binits{J.}},
\oauthor{\bsnm{Joshi}, \binits{M.}},
\oauthor{\bsnm{Chen}, \binits{D.}},
\oauthor{\bsnm{Levy}, \binits{O.}},
\oauthor{\bsnm{Lewis}, \binits{M.}},
\oauthor{\bsnm{Zettlemoyer}, \binits{L.}},
\oauthor{\bsnm{Stoyanov}, \binits{V.}}:
Roberta: A robustly optimized bert pretraining approach.
arXiv preprint arXiv:1907.11692
(2019)
\end{botherref}
\endbibitem

\bibitem[\protect\citeauthoryear{Lewis et~al.}{2019}]{lewis2019bart}
\begin{botherref}
\oauthor{\bsnm{Lewis}, \binits{M.}},
\oauthor{\bsnm{Liu}, \binits{Y.}},
\oauthor{\bsnm{Goyal}, \binits{N.}},
\oauthor{\bsnm{Ghazvininejad}, \binits{M.}},
\oauthor{\bsnm{Mohamed}, \binits{A.}},
\oauthor{\bsnm{Levy}, \binits{O.}},
\oauthor{\bsnm{Stoyanov}, \binits{V.}},
\oauthor{\bsnm{Zettlemoyer}, \binits{L.}}:
Bart: Denoising sequence-to-sequence pre-training for natural language generation, translation, and comprehension.
arXiv preprint arXiv:1910.13461
(2019)
\end{botherref}
\endbibitem

\bibitem[\protect\citeauthoryear{Raffel et~al.}{2020}]{raffel2020exploring}
\begin{barticle}
\bauthor{\bsnm{Raffel}, \binits{C.}},
\bauthor{\bsnm{Shazeer}, \binits{N.}},
\bauthor{\bsnm{Roberts}, \binits{A.}},
\bauthor{\bsnm{Lee}, \binits{K.}},
\bauthor{\bsnm{Narang}, \binits{S.}},
\bauthor{\bsnm{Matena}, \binits{M.}},
\bauthor{\bsnm{Zhou}, \binits{Y.}},
\bauthor{\bsnm{Li}, \binits{W.}},
\bauthor{\bsnm{Liu}, \binits{P.J.}}:
\batitle{Exploring the limits of transfer learning with a unified text-to-text transformer}.
\bjtitle{The Journal of Machine Learning Research}
\bvolume{21}(\bissue{1}),
\bfpage{5485}--\blpage{5551}
(\byear{2020})
\end{barticle}
\endbibitem

\bibitem[\protect\citeauthoryear{Lu et~al.}{2022}]{lu2022learn}
\begin{bchapter}
\bauthor{\bsnm{Lu}, \binits{P.}},
\bauthor{\bsnm{Mishra}, \binits{S.}},
\bauthor{\bsnm{Xia}, \binits{T.}},
\bauthor{\bsnm{Qiu}, \binits{L.}},
\bauthor{\bsnm{Chang}, \binits{K.-W.}},
\bauthor{\bsnm{Zhu}, \binits{S.-C.}},
\bauthor{\bsnm{Tafjord}, \binits{O.}},
\bauthor{\bsnm{Clark}, \binits{P.}},
\bauthor{\bsnm{Kalyan}, \binits{A.}}:
\bctitle{Learn to explain: Multimodal reasoning via thought chains for science question answering}.
In: \bbtitle{The 36th Conference on Neural Information Processing Systems (NeurIPS)}
(\byear{2022})
\end{bchapter}
\endbibitem

\bibitem[\protect\citeauthoryear{Lin et~al.}{2014}]{lin2014microsoft}
\begin{bchapter}
\bauthor{\bsnm{Lin}, \binits{T.-Y.}},
\bauthor{\bsnm{Maire}, \binits{M.}},
\bauthor{\bsnm{Belongie}, \binits{S.}},
\bauthor{\bsnm{Hays}, \binits{J.}},
\bauthor{\bsnm{Perona}, \binits{P.}},
\bauthor{\bsnm{Ramanan}, \binits{D.}},
\bauthor{\bsnm{Doll{\'a}r}, \binits{P.}},
\bauthor{\bsnm{Zitnick}, \binits{C.L.}}:
\bctitle{Microsoft coco: Common objects in context}.
In: \bbtitle{Computer Vision--ECCV 2014: 13th European Conference, Zurich, Switzerland, September 6-12, 2014, Proceedings, Part V 13},
pp. \bfpage{740}--\blpage{755}
(\byear{2014}).
\bcomment{Springer}
\end{bchapter}
\endbibitem

\bibitem[\protect\citeauthoryear{Dosovitskiy et~al.}{2020}]{dosovitskiy2020image}
\begin{botherref}
\oauthor{\bsnm{Dosovitskiy}, \binits{A.}},
\oauthor{\bsnm{Beyer}, \binits{L.}},
\oauthor{\bsnm{Kolesnikov}, \binits{A.}},
\oauthor{\bsnm{Weissenborn}, \binits{D.}},
\oauthor{\bsnm{Zhai}, \binits{X.}},
\oauthor{\bsnm{Unterthiner}, \binits{T.}},
\oauthor{\bsnm{Dehghani}, \binits{M.}},
\oauthor{\bsnm{Minderer}, \binits{M.}},
\oauthor{\bsnm{Heigold}, \binits{G.}},
\oauthor{\bsnm{Gelly}, \binits{S.}}, et al.:
An image is worth 16x16 words: Transformers for image recognition at scale.
arXiv preprint arXiv:2010.11929
(2020)
\end{botherref}
\endbibitem

\bibitem[\protect\citeauthoryear{Paszke et~al.}{2019}]{paszke2019pytorch}
\begin{botherref}
\oauthor{\bsnm{Paszke}, \binits{A.}},
\oauthor{\bsnm{Gross}, \binits{S.}},
\oauthor{\bsnm{Massa}, \binits{F.}},
\oauthor{\bsnm{Lerer}, \binits{A.}},
\oauthor{\bsnm{Bradbury}, \binits{J.}},
\oauthor{\bsnm{Chanan}, \binits{G.}},
\oauthor{\bsnm{Killeen}, \binits{T.}},
\oauthor{\bsnm{Lin}, \binits{Z.}},
\oauthor{\bsnm{Gimelshein}, \binits{N.}},
\oauthor{\bsnm{Antiga}, \binits{L.}}, et al.:
Pytorch: An imperative style, high-performance deep learning library.
Advances in neural information processing systems
\textbf{32}
(2019)
\end{botherref}
\endbibitem

\bibitem[\protect\citeauthoryear{Loshchilov and Hutter}{2016}]{loshchilov2016sgdr}
\begin{botherref}
\oauthor{\bsnm{Loshchilov}, \binits{I.}},
\oauthor{\bsnm{Hutter}, \binits{F.}}:
Sgdr: Stochastic gradient descent with warm restarts.
arXiv preprint arXiv:1608.03983
(2016)
\end{botherref}
\endbibitem

\bibitem[\protect\citeauthoryear{Loshchilov and Hutter}{2017}]{loshchilov2017decoupled}
\begin{botherref}
\oauthor{\bsnm{Loshchilov}, \binits{I.}},
\oauthor{\bsnm{Hutter}, \binits{F.}}:
Decoupled weight decay regularization.
arXiv preprint arXiv:1711.05101
(2017)
\end{botherref}
\endbibitem

\bibitem[\protect\citeauthoryear{Nanni et~al.}{2020}]{nanni2020insect}
\begin{barticle}
\bauthor{\bsnm{Nanni}, \binits{L.}},
\bauthor{\bsnm{Maguolo}, \binits{G.}},
\bauthor{\bsnm{Pancino}, \binits{F.}}:
\batitle{Insect pest image detection and recognition based on bio-inspired methods}.
\bjtitle{Ecological Informatics}
\bvolume{57},
\bfpage{101089}
(\byear{2020})
\end{barticle}
\endbibitem

\bibitem[\protect\citeauthoryear{Ayan et~al.}{2020}]{ayan2020crop}
\begin{barticle}
\bauthor{\bsnm{Ayan}, \binits{E.}},
\bauthor{\bsnm{Erbay}, \binits{H.}},
\bauthor{\bsnm{Var{\c{c}}{\i}n}, \binits{F.}}:
\batitle{Crop pest classification with a genetic algorithm-based weighted ensemble of deep convolutional neural networks}.
\bjtitle{Computers and Electronics in Agriculture}
\bvolume{179},
\bfpage{105809}
(\byear{2020})
\end{barticle}
\endbibitem

\bibitem[\protect\citeauthoryear{He et~al.}{2024}]{he2024weakly}
\begin{botherref}
\oauthor{\bsnm{He}, \binits{C.}},
\oauthor{\bsnm{Li}, \binits{K.}},
\oauthor{\bsnm{Zhang}, \binits{Y.}},
\oauthor{\bsnm{Xu}, \binits{G.}},
\oauthor{\bsnm{Tang}, \binits{L.}},
\oauthor{\bsnm{Zhang}, \binits{Y.}},
\oauthor{\bsnm{Guo}, \binits{Z.}},
\oauthor{\bsnm{Li}, \binits{X.}}:
Weakly-supervised concealed object segmentation with sam-based pseudo labeling and multi-scale feature grouping.
Advances in Neural Information Processing Systems
\textbf{36}
(2024)
\end{botherref}
\endbibitem

\bibitem[\protect\citeauthoryear{Fan et~al.}{2020}]{fan2020camouflaged}
\begin{bchapter}
\bauthor{\bsnm{Fan}, \binits{D.-P.}},
\bauthor{\bsnm{Ji}, \binits{G.-P.}},
\bauthor{\bsnm{Sun}, \binits{G.}},
\bauthor{\bsnm{Cheng}, \binits{M.-M.}},
\bauthor{\bsnm{Shen}, \binits{J.}},
\bauthor{\bsnm{Shao}, \binits{L.}}:
\bctitle{Camouflaged object detection}.
In: \bbtitle{Proceedings of the IEEE/CVF Conference on Computer Vision and Pattern Recognition},
pp. \bfpage{2777}--\blpage{2787}
(\byear{2020})
\end{bchapter}
\endbibitem

\bibitem[\protect\citeauthoryear{He et~al.}{2023}]{he2023camouflaged}
\begin{bchapter}
\bauthor{\bsnm{He}, \binits{C.}},
\bauthor{\bsnm{Li}, \binits{K.}},
\bauthor{\bsnm{Zhang}, \binits{Y.}},
\bauthor{\bsnm{Tang}, \binits{L.}},
\bauthor{\bsnm{Zhang}, \binits{Y.}},
\bauthor{\bsnm{Guo}, \binits{Z.}},
\bauthor{\bsnm{Li}, \binits{X.}}:
\bctitle{Camouflaged object detection with feature decomposition and edge reconstruction}.
In: \bbtitle{Proceedings of the IEEE/CVF Conference on Computer Vision and Pattern Recognition},
pp. \bfpage{22046}--\blpage{22055}
(\byear{2023})
\end{bchapter}
\endbibitem

\bibitem[\protect\citeauthoryear{Krizhevsky et~al.}{2012}]{krizhevsky2012imagenet}
\begin{botherref}
\oauthor{\bsnm{Krizhevsky}, \binits{A.}},
\oauthor{\bsnm{Sutskever}, \binits{I.}},
\oauthor{\bsnm{Hinton}, \binits{G.E.}}:
Imagenet classification with deep convolutional neural networks.
Advances in neural information processing systems
\textbf{25}
(2012)
\end{botherref}
\endbibitem

\bibitem[\protect\citeauthoryear{Szegedy et~al.}{2015}]{szegedy2015going}
\begin{bchapter}
\bauthor{\bsnm{Szegedy}, \binits{C.}},
\bauthor{\bsnm{Liu}, \binits{W.}},
\bauthor{\bsnm{Jia}, \binits{Y.}},
\bauthor{\bsnm{Sermanet}, \binits{P.}},
\bauthor{\bsnm{Reed}, \binits{S.}},
\bauthor{\bsnm{Anguelov}, \binits{D.}},
\bauthor{\bsnm{Erhan}, \binits{D.}},
\bauthor{\bsnm{Vanhoucke}, \binits{V.}},
\bauthor{\bsnm{Rabinovich}, \binits{A.}}:
\bctitle{Going deeper with convolutions}.
In: \bbtitle{Proceedings of the IEEE Conference on Computer Vision and Pattern Recognition},
pp. \bfpage{1}--\blpage{9}
(\byear{2015})
\end{bchapter}
\endbibitem

\bibitem[\protect\citeauthoryear{Simonyan and Zisserman}{2014}]{simonyan2014very}
\begin{botherref}
\oauthor{\bsnm{Simonyan}, \binits{K.}},
\oauthor{\bsnm{Zisserman}, \binits{A.}}:
Very deep convolutional networks for large-scale image recognition.
arXiv preprint arXiv:1409.1556
(2014)
\end{botherref}
\endbibitem

\bibitem[\protect\citeauthoryear{He et~al.}{2016}]{he2016deep}
\begin{bchapter}
\bauthor{\bsnm{He}, \binits{K.}},
\bauthor{\bsnm{Zhang}, \binits{X.}},
\bauthor{\bsnm{Ren}, \binits{S.}},
\bauthor{\bsnm{Sun}, \binits{J.}}:
\bctitle{Deep residual learning for image recognition}.
In: \bbtitle{Proceedings of the IEEE Conference on Computer Vision and Pattern Recognition},
pp. \bfpage{770}--\blpage{778}
(\byear{2016})
\end{bchapter}
\endbibitem

\bibitem[\protect\citeauthoryear{Ren et~al.}{2015}]{ren2015faster}
\begin{botherref}
\oauthor{\bsnm{Ren}, \binits{S.}},
\oauthor{\bsnm{He}, \binits{K.}},
\oauthor{\bsnm{Girshick}, \binits{R.}},
\oauthor{\bsnm{Sun}, \binits{J.}}:
Faster r-cnn: Towards real-time object detection with region proposal networks.
Advances in neural information processing systems
\textbf{28}
(2015)
\end{botherref}
\endbibitem

\bibitem[\protect\citeauthoryear{Lin et~al.}{2017}]{lin2017feature}
\begin{bchapter}
\bauthor{\bsnm{Lin}, \binits{T.-Y.}},
\bauthor{\bsnm{Doll{\'a}r}, \binits{P.}},
\bauthor{\bsnm{Girshick}, \binits{R.}},
\bauthor{\bsnm{He}, \binits{K.}},
\bauthor{\bsnm{Hariharan}, \binits{B.}},
\bauthor{\bsnm{Belongie}, \binits{S.}}:
\bctitle{Feature pyramid networks for object detection}.
In: \bbtitle{Proceedings of the IEEE Conference on Computer Vision and Pattern Recognition},
pp. \bfpage{2117}--\blpage{2125}
(\byear{2017})
\end{bchapter}
\endbibitem

\bibitem[\protect\citeauthoryear{Liu et~al.}{2016}]{liu2016ssd}
\begin{bchapter}
\bauthor{\bsnm{Liu}, \binits{W.}},
\bauthor{\bsnm{Anguelov}, \binits{D.}},
\bauthor{\bsnm{Erhan}, \binits{D.}},
\bauthor{\bsnm{Szegedy}, \binits{C.}},
\bauthor{\bsnm{Reed}, \binits{S.}},
\bauthor{\bsnm{Fu}, \binits{C.-Y.}},
\bauthor{\bsnm{Berg}, \binits{A.C.}}:
\bctitle{Ssd: Single shot multibox detector}.
In: \bbtitle{Computer Vision--ECCV 2016: 14th European Conference, Amsterdam, The Netherlands, October 11--14, 2016, Proceedings, Part I 14},
pp. \bfpage{21}--\blpage{37}
(\byear{2016}).
\bcomment{Springer}
\end{bchapter}
\endbibitem

\bibitem[\protect\citeauthoryear{Zhang et~al.}{2018}]{zhang2018single}
\begin{bchapter}
\bauthor{\bsnm{Zhang}, \binits{S.}},
\bauthor{\bsnm{Wen}, \binits{L.}},
\bauthor{\bsnm{Bian}, \binits{X.}},
\bauthor{\bsnm{Lei}, \binits{Z.}},
\bauthor{\bsnm{Li}, \binits{S.Z.}}:
\bctitle{Single-shot refinement neural network for object detection}.
In: \bbtitle{Proceedings of the IEEE Conference on Computer Vision and Pattern Recognition},
pp. \bfpage{4203}--\blpage{4212}
(\byear{2018})
\end{bchapter}
\endbibitem

\bibitem[\protect\citeauthoryear{Redmon and Farhadi}{2018}]{redmon2018yolov3}
\begin{botherref}
\oauthor{\bsnm{Redmon}, \binits{J.}},
\oauthor{\bsnm{Farhadi}, \binits{A.}}:
Yolov3: An incremental improvement.
arXiv preprint arXiv:1804.02767
(2018)
\end{botherref}
\endbibitem

\end{thebibliography}

\end{document}